%% file: iclr2027_conference.tex
\documentclass{article} 
\PassOptionsToPackage{table}{xcolor}
\usepackage{welm}
\renewcommand*\ttdefault{cmtt}
\input{math_commands.tex}

\usepackage{hyperref}
\usepackage{url}

\usepackage{booktabs}
\usepackage{tabularx}
\usepackage{array}
\usepackage{amssymb}
\usepackage{graphicx}
\usepackage{multirow}
\usepackage{pifont}

\usepackage[table]{xcolor}
\usepackage{colortbl}
\usepackage{tikz}

\usepackage{subcaption}
\usepackage{wrapfig}
\definecolor{aegisgreen}{HTML}{2E8B57}
\definecolor{aegislight}{HTML}{E8F5E9}
\definecolor{aegishighlight}{HTML}{F1F8E9}
\definecolor{aegiszebra}{HTML}{F5FAF5}
\definecolor{aegisbar}{HTML}{66BB6A}
\definecolor{aegisbartrack}{HTML}{E0E0E0}

\usepackage{booktabs}
\usepackage[table]{xcolor}
\usepackage{pifont}
\usepackage{graphicx}
\usepackage{array}

\definecolor{TableRule}{HTML}{737373}
\definecolor{HeaderBg}{HTML}{E9EDF2}

\definecolor{CodeHeader}{HTML}{E5E7EB}
\definecolor{CodeRowA}{HTML}{F4F5F6}
\definecolor{CodeRowB}{HTML}{F8F8F8}
\definecolor{CodeText}{HTML}{777777}

\definecolor{DomainHeader}{HTML}{DDEAF4}
\definecolor{DomainRowA}{HTML}{F2F7FB}
\definecolor{DomainRowB}{HTML}{F8FBFD}
\definecolor{DomainText}{HTML}{294A63}

\definecolor{GeneralHeader}{HTML}{F2E7CF}
\definecolor{GeneralRowA}{HTML}{FBF7EF}
\definecolor{GeneralRowB}{HTML}{FEFCF8}
\definecolor{GeneralText}{HTML}{59472C}

\definecolor{GRASPRow}{HTML}{DDEFE5}
\definecolor{GRASPAccent}{HTML}{176B4D}
\definecolor{GRASPText}{HTML}{104E39}

\definecolor{MarkFull}{HTML}{21835A}
\definecolor{MarkPartial}{HTML}{D58924}
\definecolor{MarkNone}{HTML}{9B9B9B}

\newcommand{\cmark}{%
  \textcolor{MarkFull}{\ding{51}}%
}
\newcommand{\pmark}{%
  \textcolor{MarkPartial}{\raisebox{0.05ex}{\large$\bullet$}}%
}
\newcommand{\xmark}{%
  \textcolor{MarkNone}{\ding{55}}%
}

\DeclareRobustCommand{\skillforgebench}{\textsc{SkillForgeBench}}

\newcolumntype{Y}{>{\raggedright\arraybackslash}X}

\title{\texorpdfstring{\textcolor{WELMGreen}{SkillForgeBench}}{SkillForgeBench}: Evaluating Agent Reliability Grounded in Agent Skill Specifications}

\author{%
Zhengxi Li\textsuperscript{1,*}, Junyi Zhou\textsuperscript{2,3,*}, Qiyuan Zhang\textsuperscript{2,3}, Chuhan Wu\textsuperscript{3},\\
Fuyuan Lyu\textsuperscript{4}, Le Tian\textsuperscript{3}, Xiao Zhou\textsuperscript{3}, Xue Liu\textsuperscript{1,4}, Chen Ma\textsuperscript{2}, Yufei Wang\textsuperscript{3,\dag}\\[0.7em]
{\normalfont\WELMDisplayFont\fontsize{10}{12}\selectfont \textsuperscript{1}MBZUAI \qquad \textsuperscript{2}City University of Hong Kong}\\[0.2em]
{\normalfont\WELMDisplayFont\fontsize{10}{12}\selectfont \textsuperscript{3}WeChat AI, Tencent \qquad \textsuperscript{4}McGill \textendash{} Mila \& Quebec AI Institute}\\[0.3em]
{\normalfont\rmfamily\fontsize{10}{12}\selectfont \textsuperscript{*}Equal contribution \qquad \textsuperscript{\dag}Corresponding author}
}

\hypersetup{
  pdftitle={SkillForgeBench: Evaluating Agent Reliability Grounded in Agent Skill Specifications},
  pdfauthor={Zhengxi Li, Junyi Zhou, Qiyuan Zhang, Chuhan Wu, Fuyuan Lyu, Le Tian, Xiao Zhou, Xue Liu, Chen Ma, Yufei Wang}
}

\usepackage[T1]{fontenc}
\input{sections/appendix_sections/prompts/appendix_prompt_style}

\begin{document}

\maketitle
\begingroup
\makeatletter
\let\@footnotetext\H@@footnotetext
\long\def\@makefntext#1{\noindent#1}
\makeatother
\renewcommand{\thefootnote}{}
\footnotetext{\fontsize{8}{10}\selectfont\raggedright\hypersetup{urlcolor=black}%
\mbox{Zhengxi Li \href{mailto:zhengxi.li@mbzuai.ac.ae}{\texttt{<zhengxi.li@mbzuai.ac.ae>}}},
\mbox{Junyi Zhou \href{mailto:junyizhou8-c@my.cityu.edu.hk}{\texttt{<junyizhou8-c@my.cityu.edu.hk>}}},
\mbox{Yufei Wang \href{mailto:garyyfwang@tencent.com}{\texttt{<garyyfwang@tencent.com>}}}}
\endgroup

\begin{abstract} 
Evaluating an agent's ability to strictly satisfy diverse and procedurally complex requirements is essential to determining its practical reliability.
Conducting such an evaluation demands agentic tasks that reconcile broad domain coverage with deep procedural complexity.
Existing paradigms face a fundamental trade-off: expert-curated benchmarks achieve procedural depth but suffer from domain diversity bottlenecks, whereas crowdsourced benchmarks span diverse scenarios but lack execution rigor.
To resolve this tension, we present \textbf{\skillforgebench{}}, which leverages agent skills as unifying execution blueprints across three co-designed stages: 
(1) \textbf{Resource Preparation} uses domain-balanced skills to guide artifact retrieval, establishing broad workflow coverage;
(2) \textbf{Skill-driven Task Synthesis} employs skills as procedural anchors to partition artifact roles, synthesize naturalistic task instructions, and package sandboxed scenarios for deep procedural execution; and
(3) \textbf{Holistic Evaluation Design} translates skill-embedded requirements into fine-grained completion and process rubrics to enable strict requirement-level scoring.
Comprising 842 sandboxed tasks across five major professional domains, \textbf{\skillforgebench{}} reveals a steep capability hierarchy among eight mainstream agents, led by \textsc{Opus 4.8} ($40.9\%$ strict completion).
Requiring all constrains satisfaction uncovers severe capability bottlenecks across all model tiers that remain masked under individual rubric pass rates ($70.0\%$ to $85.7\%$ vs.\ $12.6\%$ to $40.9\%$).
Crucially, under the external guidance from skills,  top-tier agents obtain larger gains than low-tier agents (i.e., $\Delta 19.2\%$ for \textsc{Opus 4.8} vs.\ $\Delta 5.8\%$ for \textsc{Haiku 4.5}).
Finally, agents satisfy process rubrics far less frequently than completion rubrics (under $3\%$ for \textsc{Haiku 4.5}), proving that completion-only checks overestimate real-world agent reliability.
\end{abstract}

\input{sections/introduction}

\input{sections/related}

\input{sections/method}

\input{sections/experiment}

\input{sections/results}
\input{sections/conclusion}

\input{sections/limitations}

\bibliography{iclr2027_conference}
\bibliographystyle{iclr2027_conference}

\newpage
\appendix

\input{sections/appendix}

\end{document}

%% file: math_commands.tex
\usepackage{amsmath,amsfonts,bm}

\def\eqref#1{equation~\ref{#1}}

\def\1{\bm{1}}

\DeclareMathAlphabet{\mathsfit}{\encodingdefault}{\sfdefault}{m}{sl}
\SetMathAlphabet{\mathsfit}{bold}{\encodingdefault}{\sfdefault}{bx}{n}



%% file: sections/appendix_sections/prompts/appendix_prompt_style.tex
\usepackage[T1]{fontenc}
\usepackage{xcolor}
\usepackage{listings}
\usepackage{upquote}
\usepackage{booktabs}
\usepackage{array}
\usepackage{url}
\definecolor{SFPromptBorder}{gray}{0.68}
\definecolor{SFPromptBackground}{gray}{0.975}
\definecolor{SFPromptMuted}{gray}{0.40}
\newcommand{\SFPromptFont}{\fontencoding{T1}\fontfamily{lmtt}\selectfont}
\newcommand{\SFNeedspace}[1]{\par\begingroup
  \dimen0=\pagegoal\advance\dimen0 by -\pagetotal
  \ifdim\dimen0<#1\relax\newpage\fi\endgroup}
\lstdefinestyle{sfprompt}{
  language={},
  basicstyle=\SFPromptFont\fontsize{9}{11}\selectfont,
  columns=fullflexible,keepspaces=true,
  showstringspaces=false,showspaces=false,showtabs=false,
  breaklines=true,breakatwhitespace=false,breakautoindent=true,
  breakindent=1em,
  postbreak=\mbox{\textcolor{SFPromptMuted}{\(\hookrightarrow\)}\space},
  frame=single,framerule=0.3pt,framesep=6pt,
  rulecolor=\color{SFPromptBorder},backgroundcolor=\color{SFPromptBackground},
  xleftmargin=7pt,xrightmargin=7pt,
  aboveskip=0.7em,belowskip=1em,
  tabsize=2,upquote=true,
  literate=
    {·}{{\ensuremath{\cdot}}}1
    {×}{{\ensuremath{\times}}}1
    {–}{{\textendash{}}}1
    {—}{{\textemdash{}}}1
    {…}{{\ldots{}}}1
    {→}{{\ensuremath{\rightarrow}}}1
    {↔}{{\ensuremath{\leftrightarrow}}}1
    {⇒}{{\ensuremath{\Rightarrow}}}1
    {∈}{{\ensuremath{\in}}}1
    {≤}{{\ensuremath{\leq}}}1
    {≥}{{\ensuremath{\geq}}}1
    {⚠}{{\ensuremath{\triangle}}}1
    {️}{{}}1
}

%% file: sections/introduction.tex
\section{Introduction}

As large language model (LLM) agents take on increasingly complex professional work across diverse domains~\citep{merrill2026terminalbench,ye2026claweval}, establishing rigorous benchmarks to evaluate their reliability across varied workflows has become essential.
In practical deployments, a reliable agent must strictly satisfy all operational constraints (e.g., procedural steps and formatting specifications) across diverse scenarios, rather than produce superficially plausible deliverables.
Every unfulfilled constraint forces human intervention, invalidating the practical utility of autonomous agents.
However, existing agentic benchmarks struggle to evaluate reliability across diverse workflows due to a fundamental dilemma in task curation.
Expert-curated benchmarks pair complex tasks with strict verification scripts, offering high task fidelity, yet they remain strictly bottlenecked by manual authoring costs and confined to narrow domains~\citep{yang2026make,Wu2026LargeScaleTA}.
Conversely, crowdsourced benchmarks achieve domain diversity, but their evaluation signals are frequently decoupled from workspace contexts, ultimately yielding shallow tasks that fail to probe multi-dimensional reliability~\citep{Hua2026CLIUniverseTV,Han2026RideWayBE}.

To bridge this gap, we leverage \textbf{agent skills}~\citep{li2026dynamic}, the native structural primitives of agent ecosystems, as execution-grounded blueprints for benchmark design.
Authored by human experts, skills encode rich procedural workflows, domain-specific knowledge, and professional conventions that naturally reflect user demands across diverse domains.
Building on this insight, we present \textbf{\skillforgebench{}}, an agentic evaluation benchmark built through a three-stage skill-centric design architecture that directly fulfills these essential properties:
First, during \textbf{Resource Preparation} (\S\ref{sec:preparation}), we curate a domain-balanced skill set and deploy web-enhanced retrieval to harvest artifacts and user queries, establishing broad workflow coverage.
Second, during \textbf{Skill-driven Task Synthesis} (\S\ref{sec:task_synthesis}), skills guide the functional partitioning of artifacts into editable targets, reference evidence, and read-only contexts to construct concrete task scenarios. We then leverage retrieved user queries to compose naturalistic task instructions and package each scenario into a sandboxed environment, achieving deep procedural instantiation.
Third, during \textbf{Holistic Evaluation Design} (\S\ref{sec:rubric_construction}), we evaluate requirement satisfaction across both final deliverables via \textit{completion rubrics} and execution trajectories via \textit{process rubrics}. To establish truly rigorous reliability, we prioritize \textit{strict scoring} that requires full compliance with constraints, complemented by \textit{loose scoring} for fine-grained capability diagnosis, while conducting all evaluations in both skill-conditioned and unconditioned settings.
In total, \skillforgebench{} currently comprises 842 rigorously verified tasks across diverse domains, uniquely combining real-world demand, authentic environments, fine-grained rubrics, execution evaluation, and explicit skill use (Table~\ref{tab:benchmark_comparison}), and establishing a challenging testbed for evaluating agent capabilities under realistic constraint adherence. 

Evaluating eight mainstream agents on \skillforgebench{} reveals a steep capability hierarchy across leading proprietary and open-weight models.
Frontier models, led by \textsc{Opus 4.8} ($40.9\%$ strict completion), substantially outpace low-tier models such as \textsc{Qwen3.5-397B}
, which drops to near $12.6\%$.
Crucially, while loose scoring yields deceptively high rubric pass rates ($70.0\%$ to $85.7\%$), strict scoring requiring total constraint compliance exposes severe capability bottlenecks across all model tiers.
Beyond leaderboard rankings, our diagnostic analyses yield three core insights:
First, strict evaluation defines the true boundary of agent reliability, while loose scoring mask hidden process failures and create an illusion of competence.
Second, skill conditioning improves individual rubric pass rates similarly across models ($\Delta 5.1\%$ to $7.4\%$), and frontier models convert skills into far larger strict completion gains ($\Delta 19.2\%$ for \textsc{Opus 4.8} vs.\ $\Delta 5.8\%$ for \textsc{Haiku 4.5}).
Third, performance bottlenecks stem from process discipline rather than deliverable production: agents satisfy process rubrics far less frequently than completion rubrics (i.e., \textsc{Haiku 4.5} meeting all process constraints on fewer than $3\%$ of tasks).
In summary, these findings confirm that real-world agent reliability hinges on strict constraint adherence across the entire workflow execution. \textbf{\skillforgebench{}} thus provides a rigorous measurement standard to evaluate this essential capability.

%% file: sections/related.tex
\section{Related Work}
\subsection{Agentic Benchmarks}

Agent evaluation has evolved alongside expanding LLM agents' capabilities. Early benchmarks target specialized, single-domain workflows, such as code repair~\citep{jimenez2024swebench}, web navigation~\citep{zhou2024webarena},
multi-app API coordination~\citep{trivedi2024appworld},
and policy-adherent customer service~\citep{yao2025taubench}. 
Recent benchmarks target autonomous agents operating in general-purpose environments. OSWorld~\citep{xie2024osworld} provisions full desktop VMs; Terminal-Bench~\citep{merrill2026terminalbench} evaluates system administration in Docker containers; ClawBench~\citep{zhang2026clawbench} and Claw-Eval~\citep{ye2026claweval} define rubric-scored tasks across productivity domains for Claw-style agents. PinchBench~\citep{pinchbench2026} combines automated checks with LLM-judge rubrics. These benchmarks span more diverse scenarios but remain manually curated and static in scale.

To bypass manual authoring constraints, recent works explore automated task harvesting and synthesis. RealClawBench~\citep{lv2026realclawbench} reconstructs environments from production agent sessions; WildClawBench \citep{ding2026wildclawbench} runs tasks inside real CLI harnesses; LiveClawBench \citep{long2026liveclawbench} continuously refreshes tasks against live services to prevent contamination. 
Automated synthesis pipelines address scale and cost simultaneously. ClawEnvKit~\citep{li2026clawenvkit} generates environments from natural-language descriptions at orders-of-magnitude lower cost. ClawForge~\citep{lai2026clawforge} templates state-conflict scenarios with grounded slots. However, current synthesis pipelines lack a principled framework to reconcile broad domain diversity with deep procedural complexity, often generating tasks that lack structured requirement grounding and rigorous evaluation rubrics. Scaling evaluation reliability remains their central limitation. Table~\ref{tab:benchmark_comparison} summarizes these comparisons.

\subsection{Synthesis from Skills}


Synthetic data generation has emerged as a key paradigm for scaling agent training and evaluation. Early approaches synthesize function-calling instructions, and interaction traces over real or mocked API endpoints, validating trajectory executability programmatically~\citep{liu2024apigen, wang2026synthagent}. Subsequent methods ground trajectory synthesis in live web or desktop environments to produce realistic execution traces~\citep{xu2025agenttrek, wang2024executable}, while environment synthesis techniques automatically generate or expand sandbox setups to diversify execution contexts~\citep{zhu2026termigen, dong2026agentworld}. Notably, Terminal-World~\citep{cheng2026terminalworld} also uses skills for data synthesis, but does not use them to introduce process constraints into tasks: the skill's procedure serves only as a guideline for the teacher model, so in the task itself, skills define only the required final outcome. In contrast, \skillforgebench{} leverages the rich procedural steps and execution constraints of skills to derive tasks that specify both final outcomes and process constraints, guaranteeing task realism.

\begin{table*}[t]
\centering
\small
\setlength{\tabcolsep}{4pt}
\renewcommand{\arraystretch}{1.15}

\caption{
Comparison of \skillforgebench{} with representative agent benchmarks.
\textit{Real Demand} and \textit{Authentic Env.} measure task and
environment realism;
\textit{Fine Rubrics} and \textit{Process Eval.} characterize evaluation granularity;
and \textit{Skill Use} indicates whether skills are explicitly evaluated.
A green check, orange dot, and gray cross denote full, partial, and no support.
}
\label{tab:benchmark_comparison}

\resizebox{\textwidth}{!}{%
\begin{tabular}{l l r c c c c c}
\toprule

\textbf{Benchmark}
& \textbf{Task Source}
& \textbf{Scale}
& \textbf{\shortstack{Real\\[-1pt]Demand}}
& \textbf{\shortstack{Authentic\\[-1pt]Env.}}
& \textbf{\shortstack{Fine\\[-1pt]Rubrics}}
& \textbf{\shortstack{Process\\[-1pt]Eval.}}
& \textbf{\shortstack{Skill\\[-1pt]Use}}
\\

\midrule

WebArena~\citep{zhou2024webarena}
& Authored web intents
& 812
& \pmark
& \pmark
& \xmark
& \xmark
& \xmark
\\

AppWorld~\citep{trivedi2024appworld}
& Authored app workflows
& 750
& \pmark
& \pmark
& \xmark
& \xmark
& \xmark
\\

Terminal-Bench 2.0~\citep{merrill2026terminalbench}
& Expert workflows
& 89
& \pmark
& \cmark
& \xmark
& \xmark
& \xmark
\\

ClawBench~\citep{zhang2026clawbench}
& Everyday assistant tasks
& 153
& \pmark
& \cmark
& \pmark
& \cmark
& \xmark
\\

LiveClawBench~\citep{long2026liveclawbench}
& Real usage cases
& 134
& \pmark
& \pmark
& \cmark
& \xmark
& \xmark
\\

RealClawBench~\citep{lv2026realclawbench}
& Production sessions
& 281
& \cmark
& \cmark
& \xmark
& \xmark
& \xmark
\\

Claw-Eval~\citep{ye2026claweval}
& Human-curated tasks
& 300
& \pmark
& \pmark
& \cmark
& \cmark
& \xmark
\\

SkillsBench~\citep{li2026skillsbenchbenchmarkingagentskills}
& Curated skill tasks
& 86
& \xmark
& \pmark
& \xmark
& \xmark
& \cmark
\\

Agentic Skills at Scale~\citep{shaposhnikov2026frameworkevaluatingagenticskills}
& LLM-synthesized tasks
& 1,000
& \xmark
& \xmark
& \cmark
& \cmark
& \cmark
\\

\midrule

\skillforgebench{}
& \textbf{Skills + real artifacts}
& \textbf{842}
& \cmark
& \cmark
& \cmark
& \cmark
& \cmark
\\

\bottomrule
\end{tabular}%
}

\end{table*}

%% file: sections/method.tex
\section{SkillForgeBench Construction}
\label{sec:construction}

\begin{figure*}[t]
    \centering
    \includegraphics[width=0.98\textwidth]{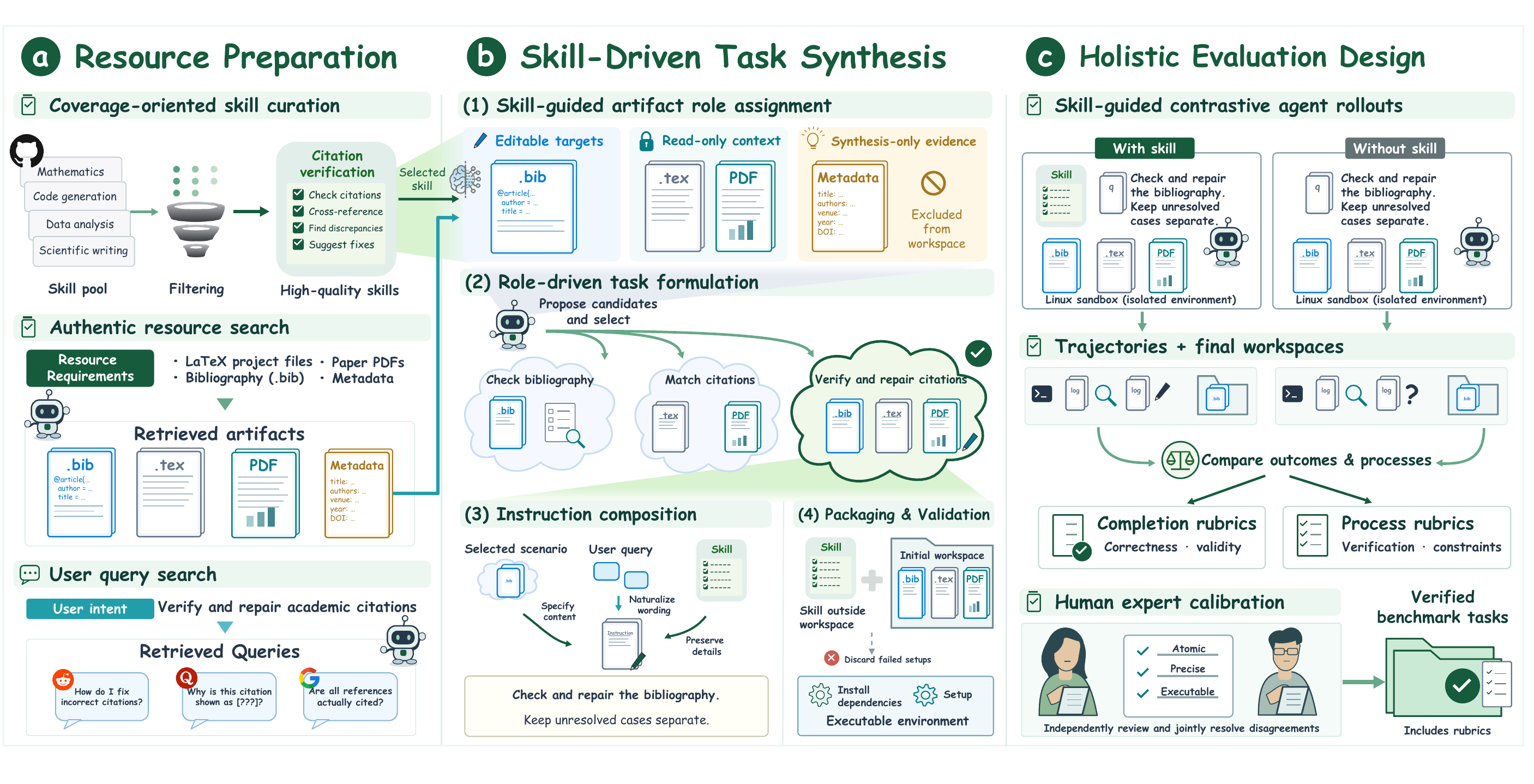}
\caption{
Overview of the \skillforgebench{} construction pipeline, demonstrated using a citation verification skill as a running example:
(a) Skills guide retrieval of real artifacts and user queries;
(b) These inputs support task formulation, instruction composition, and environment packaging; and
(c) Contrastive rollouts with and without skill access inform completion and process rubrics, followed by human task validation.
}
    \label{fig:pipeline}
\end{figure*}

\skillforgebench{} is designed to evaluate an agent's ability to holistically satisfy diverse and procedurally complex constraints in long-horizon workflows. This requires benchmark to fulfill three properties simultaneously: broad workflow coverage, deep procedural instantiation, and fine-grained yet holistic evaluation. We achieve these through three components (Figure~\ref{fig:pipeline}): 
1. \textbf{Resource Preparation} (\S\ref{sec:preparation}): We curate a domain-balanced, procedurally rich skill set and retrieve authentic open-web materials to establish broad workflow coverage;
2. \textbf{Skill-driven Task Synthesis} (\S\ref{sec:task_synthesis}): We leverage skills to partition artifact roles, formulate scenario tasks, and compose naturalistic instructions, deeply instantiating procedural logic into executable environments;
3. \textbf{Holistic Evaluation Design} (\S\ref{sec:rubric_construction}): We derive calibrated completion and process rubrics through contrastive rollouts and human expert audits, evaluating agents across fine-grained constraints and dual scoring regimes. 

\subsection{Resource Preparation}
\label{sec:preparation}

In this stage, our primary objective is to establish \textbf{broad workflow coverage}. 
Since skills originate from diverse real-world domains and developer communities, they collectively provide an expansive pool of realistic workflows. 
Accordingly, we design a two-fold resource preparation strategy: 
first, we implement a rigorous skill curation pipeline to select a domain-balanced, procedurally complex skill set $\mathcal{S}=\{s_1,\ldots,s_N\}$; 
second, we deploy a web-enhanced retrieval mechanism to harvest real-world artifacts and user queries that ground each skill workflow with authentic materials. 


\paragraph{Coverage-oriented skill curation.}
To establish diverse and complex workflows, we curate skills through a three-stage process. 
\textbf{(1) Broad skill collection and filtering.}
We collect over 50{,}000 candidate skills from major public Agent Skill ecosystems and repositories, 
including \texttt{SkillsMP}, \texttt{SkillHub}, 
and public skill repositories maintained by Anthropic, OpenAI, and other developer communities.
A rule-based filter removes malformed entries, duplicates, and skills that rely on GUI interaction, private services, or non-local execution. 
\textbf{(2) Procedural quality assessment and domain tagging.}
We employ a frontier LLM to evaluate each remaining skill across \textit{workflow complexity}, \textit{required domain knowledge}, and \textit{resource richness}, averaging these dimension scores into a unified quality score $Q(s)$. Concurrently, the model assigns a domain tag to each skill to facilitate balance. 
\textbf{(3) Domain-balanced selection.}
We rank skills by $Q(s)$ within each domain and select them uniformly across domains until reaching the total budget $N$, yielding the final skill set $\mathcal{S}$. 
The skills behind the final 842 tasks span five domains---Technology, Business, Creative, Harness Behavior, and Knowledge.
Table~\ref{tab:skill_domains} reports the full distribution.

\paragraph{Authentic resource and demand search.}
Beyond skill curation, we further leverage each skill $s \in \mathcal{S}$ as a focal seed to retrieve diverse, authentic open-web resources across two parallel search streams. 
Specifically, we analyze each skill specification to extract its core \textbf{resource requirements} $R_s=\{r_1,\ldots,r_m\}$ and \textbf{user intent summary} $G_s$, where each $r_i$ defines the expected content of one required material, and $G_s$ is a concise sentence describing what a user aims to accomplish. 
To harvest realistic grounding materials, we query the open web using $R_s$ to collect diverse \textbf{retrieved artifacts} $\mathcal{A}_s=\Psi_A(R_s)$ matching each structural requirement. 
Concurrently, we search 20 predefined public forums\footnote{The list of 20 forums is shown in Table~\ref{tab:forums}.} using $G_s$ to gather real-world community discussions, producing matching \textbf{user queries} $\mathcal{D}_s=\Psi_D(G_s)$. 
These queries serve as naturalistic demonstrations for instruction composition, providing realistic vocabulary and authentic background knowledge. 
By grounding each skill in authentic web artifacts and user expressions, this retrieval process significantly expands the diversity and realistic depth of our benchmark input resources. 


\subsection{Skill-driven Task Synthesis} 
\label{sec:task_synthesis} 

With curated skills and authentic materials in hand, our next step is to achieve \textbf{deep procedural instantiation}---faithfully injecting skills' complex procedural logic into executable benchmark tasks. To accomplish this, we design a skill-driven synthesis framework that transforms raw materials into skill-grounded scenarios, fuses procedural constraints into naturalistic user instructions, and packages them into executable environments. 


\paragraph{Skill-guided task formulation.}
While retrieval yields relevant artifacts, directly formulating tasks without constraints risks shallow edits or evidence leakage.
Since skills operate on specific input resources and produce outcomes, we partition the retrieved artifacts $\mathcal{A}_s$ into three functional roles to constrain task formulation:
(1) \textbf{editable targets} $\mathcal{A}_s^{\mathrm{edit}}$ corresponding to the required inputs;
(2) \textbf{synthesis-only evidence} $\mathcal{A}_s^{\mathrm{syn}}$ providing ground-truth reference for expected outcomes; and
(3) \textbf{read-only context} $\mathcal{A}_s^{\mathrm{ctx}}$ embedding realistic background constraints. 
Using these roles, an agent groups related artifacts into candidate scenarios $\mathcal{C}_s=\{c_1,\ldots,c_K\}$, where each $c_i$ defines a coherent artifact collection by associating editable targets with evidence for expected outcomes and relevant contextual resources.
The agent then evaluates these candidates for skill alignment, procedural depth, and executability to select the optimal task scenario $T_s$. 
For instance, in a citation verification skill, a flawed BibTeX file serves as $\mathcal{A}_s^{\mathrm{edit}}$, a metadata dump of verified citations as $\mathcal{A}_s^{\mathrm{syn}}$, and paper PDFs as $\mathcal{A}_s^{\mathrm{ctx}}$—together instantiating a task scenario $T_s$ where the agent audits and repairs the bibliography against ground-truth metadata. 

\paragraph{Naturalistic instruction composition.}
Given the task scenario $T_s$ and skill $s$, we leverage retrieved user queries $\mathcal{D}_s$ as in-context demonstrations to compose a naturalistic \textbf{task instruction} $q_s$. Notably, this retrieval-guided grounding prevents $q_s$ from collapsing into rigid synthetic templates inspired by Natural Questions~\citep{kwiatkowski-etal-2019-natural}.
Specifically, $T_s$ defines what should be accomplished, $\mathcal{D}_s$ shapes the natural phrasing and vocabulary, 
and $s$ ensures the essential details from the skill to produce $q_s$:
\begin{equation}
q_s = \Phi(T_s, \mathcal{D}_s, s).
\label{eq:instruction_composition}
\end{equation}

where the agent $\Phi$ rewrites the structured task $T_s$ into a natural user query conditioned on $\mathcal{D}_s$.

\paragraph{Packaging and execution validation.}
To ensure reproducible execution, we construct an initial \textbf{workspace} $\mathcal{W}_s^0$ containing $\mathcal{A}_s^{\mathrm{edit}}$ and $\mathcal{A}_s^{\mathrm{ctx}}$, while strictly withholding ground-truth $\mathcal{A}_s^{\mathrm{syn}}$ to prevent answer leakage.
We then package $\mathcal{W}_s^0$ into a sandboxed Linux container and automatically resolve its dependencies to form the \textbf{executable environment} $\mathcal{E}_s$.
Any task requiring persistent external services or failing sandbox build validation is discarded.

\subsection{Holistic Evaluation Design}
\label{sec:rubric_construction}

With executable tasks in place, our final goal here is to establish a \textbf{fine-grained yet holistic evaluation} that assesses an agent's capability to satisfy diverse requirements and procedural constraints in long-horizon workflows. 
In real-world environments, an agent may produce a superficially correct output while violating critical process rules and skipping mandatory safety verifications. 
Thus, \skillforgebench{} evaluates individual requirements in detail while verifying holistic compliance across the entire workflow. 

\paragraph{Completion and process rubric construction.}
To cover both outcome and execution constraints, we structure evaluation into two fine-grained rubric dimensions: \textbf{Completion Rubrics ($\mathcal{R}^{\mathrm{comp}}$)} evaluate outcome constraints on final workspace, focusing on the structural correctness and content validity of skill-specified deliverables; \textbf{Process Rubrics ($\mathcal{R}^{\mathrm{proc}}$)} evaluate execution constraints along agent trajectories, covering adherence to procedural steps, intermediate verification checks, and safety bounds in the skill workflow. 
However, directly prompting language models to synthesize rubrics from task descriptions often yields abstract and superficial rules~\citep{zhou-etal-2026-rubricbench}, missing critical execution details and specific environment changes. 
To derive precise and executable rubrics, we combine skill-guided \textbf{contrastive agent rollouts} with \textbf{human expert calibration}. 
Specifically, a reference agent $\pi_{\mathrm{ref}}$ executes task instruction $q_s$ twice on identical environments $\mathcal{E}_s$: 
\begin{equation}
(\tau_s^{+}, \mathcal{W}_s^{+}) = \operatorname{Rollout}(\pi_{\mathrm{ref}}, q_s, \mathcal{E}_s, s), 
\qquad 
(\tau_s^{-}, \mathcal{W}_s^{-}) = \operatorname{Rollout}(\pi_{\mathrm{ref}}, q_s, \mathcal{E}_s, \varnothing),
\label{eq:contrastive_rollouts}
\end{equation}
where $(\tau_s^+, \mathcal{W}_s^+)$ and $(\tau_s^-, \mathcal{W}_s^-)$ denote the resulting trajectories and workspace states generated with skill ($s$) and without skill ($\varnothing$), respectively. 
An evaluation agent then contrasts the trajectory deltas ($\tau_s^+ \text{ vs. } \tau_s^-$) and state deltas ($\mathcal{W}_s^+ \text{ vs. } \mathcal{W}_s^-$) to isolate the procedural constraints introduced by skill $s$, synthesizing draft rubrics for $\mathcal{R}_s^{\mathrm{comp}}$ and $\mathcal{R}_s^{\mathrm{proc}}$. 
Finally, two human experts calibrate these draft rubrics through three explicit steps: \textbf{atomic decomposition} splits complex rules into single atomic rubrics, \textbf{filtering} removes superficial formatting noise while preserving core domain requirements, and \textbf{executable grounding} refines vague descriptions into clear, unambiguous natural language rubrics with explicit pass/fail conditions. In practice, the two experts independently review draft rubrics, resolving disagreements jointly.

\paragraph{Evaluation Metrics Design.} 
To assess both outcome correctness and execution constraints, \skillforgebench{} evaluates completion and process rubrics ($\mathcal{R}^{\mathrm{comp}}, \mathcal{R}^{\mathrm{proc}}$) under two distinct execution settings: \textbf{with skill ($w/\text{skill}$)}, which tests an agent's capacity to adhere to explicit procedural instructions, and \textbf{without skill ($w/o\text{ skill}$)}, which probes whether the agent can complete complex workflows without external guidance. This design allows us to distinguish whether successful task execution stems from the model's internal knowledge or from external procedural guidance. 
For a given task $T$ evaluated against a calibrated rubric set $\mathcal{R} = \{r_1, r_2, \ldots, r_k\}$, an evaluation agent inspecting the execution trajectory $\tau$ and final workspace state $\mathcal{W}$ to evaluate each atomic check $r_i \in \{0, 1\}$. 
We then compute the Completion Score ($S_{\text{comp}}$) and Process Score ($S_{\text{proc}}$) using two scoring regimes: 
\begin{equation}
S^{\text{loose}}(T) = \frac{1}{k} \sum_{i=1}^{k} r_i,
\qquad
S^{\text{strict}}(T) = \prod_{i=1}^{k} r_i.
\label{eq:scoring_schemes}
\end{equation}
\textit{Loose scoring} ($S^{\text{loose}}$) measures the mean pass rate across rubric items to provide fine-grained diagnostic progress, whereas \textit{strict scoring} ($S^{\text{strict}}$) enforces an all-or-nothing logical AND to evaluate overall system reliability under real-world conditions. These two regimes capture complementary demands: fine-grained feedback on incremental progress and strict satisfaction of all constraints for reliable end-to-end execution.

%% file: sections/experiment.tex
\section{Experiments}

In this section, we use LLM-as-a-Judge to execute rubrics and conduct a systematic empirical evaluation of how well mainstream LLM-based agents satisfy the constraints of \skillforgebench{} tasks. To demonstrate the evaluation utility and diagnostic power of our benchmark, our experiments are organized into 3 core parts:
(1) \textbf{Main Leaderboard}, measuring how well agents satisfy individual constraints and all constraints jointly, together with the impact of skill grounding and execution efficiency;
(2) \textbf{Reliability of Task and Rubric Construction}, independently auditing the quality of generated tasks and rubrics and measuring agreement across expert judgments; and
(3) \textbf{Diagnostic Analysis}, dissecting results across release dates, model capability tiers, and rubric categories to identify which constraints agents fail to satisfy.

\paragraph{Experimental Setup.}
We evaluate eight mainstream LLMs, including both proprietary and open-weight models, using the same OpenHands agent harness. The evaluated backbone models are Claude Opus 4.8, Claude Sonnet 4.6, Claude Haiku 4.5, Gemini 3.6 Flash, GLM-5.2, Kimi K2.6, Qwen3.5-397B, and Tencent Hy3. Using a common harness controls for performance differences attributable to the agent framework, enabling a fairer comparison of the underlying models.

\begin{table*}[t]
\centering
\small
\setlength{\tabcolsep}{3.5pt}
\renewcommand{\arraystretch}{1.2}
\caption{
\textbf{Main leaderboard of model performance and execution efficiency.}
Task-level mean scores (\%) are reported under loose and strict scoring for Completion ($S_{\text{comp}}$), evaluated with and without skill, and Process ($S_{\text{proc}}$), evaluated with skill. 
The efficiency columns report the mean total number of interaction steps (\# Assistant) and tool calls (\# Tool-call) for runs with and without skill.
\textbf{Bold} indicates the highest score and \underline{underlining} indicates the second-highest score in each performance column. 
}
\resizebox{.95\textwidth}{!}{%
\begin{tabular}{@{}l ccc p{6pt} ccc p{6pt} cc cc@{}}
\toprule
& \multicolumn{3}{c}{\cellcolor{aegisgreen}\textcolor{white}{\textbf{Loose}}}
& {}
& \multicolumn{3}{c}{\cellcolor{aegisgreen!80!black}\textcolor{white}{\textbf{Strict}}}
& {}
& \multicolumn{4}{c}{\cellcolor{aegisgreen!60!black}\textcolor{white}{\textbf{Efficiency}}} \\
\cmidrule(lr){2-4} \cmidrule(lr){6-8} \cmidrule(lr){10-13}
\textbf{Model}
& \cellcolor{aegislight}\footnotesize\textbf{$S_{\text{comp}}$}
& \cellcolor{aegislight}\footnotesize\textbf{$S_{\text{proc}}$}
& \cellcolor{aegislight}\footnotesize\textbf{$S_{\text{comp}}$}
& {}
& \cellcolor{aegislight}\footnotesize\textbf{$S_{\text{comp}}$}
& \cellcolor{aegislight}\footnotesize\textbf{$S_{\text{proc}}$}
& \cellcolor{aegislight}\footnotesize\textbf{$S_{\text{comp}}$}
& {}
& \multicolumn{2}{c}{\cellcolor{aegislight}\footnotesize\textbf{\# Assistant}}
& \multicolumn{2}{c}{\cellcolor{aegislight}\footnotesize\textbf{\# Tool-call}} \\
\cmidrule(lr){10-11} \cmidrule(lr){12-13}
& \cellcolor{aegislight}\footnotesize w/ skill
& \cellcolor{aegislight}\footnotesize w/ skill
& \cellcolor{aegislight}\footnotesize w/o skill
& {}
& \cellcolor{aegislight}\footnotesize w/ skill
& \cellcolor{aegislight}\footnotesize w/ skill
& \cellcolor{aegislight}\footnotesize w/o skill
& {}
& \cellcolor{aegislight}\footnotesize w/ skill
& \cellcolor{aegislight}\footnotesize w/o skill
& \cellcolor{aegislight}\footnotesize w/ skill
& \cellcolor{aegislight}\footnotesize w/o skill \\
\midrule
\rowcolor{aegiszebra}
Opus 4.8
& \textbf{85.7}
& \textbf{87.9}
& \textbf{79.8}
& {}
& \textbf{40.9}
& \underline{27.7}
& \textbf{21.7}
& {}
& 28.6 & 24.6 & 41.0 & 38.5 \\
GLM-5.2
& \underline{84.1}
& \underline{86.7}
& \underline{76.9}
& {}
& \underline{36.8}
& \textbf{28.7}
& \underline{20.0}
& {}
& 42.6 & 39.6 & 58.0 & 54.8 \\
\rowcolor{aegiszebra}
Sonnet 4.6
& 81.5 & 76.0 & 75.3
& {}
& 28.5 & 10.2 & 15.0
& {}
& 15.5 & 11.7 & 38.2 & 29.6 \\
Hy3
& 80.9 & 84.2 & 73.9
& {}
& 27.3 & 21.6 & 14.0
& {}
& 34.7 & 31.8 & 44.3 & 41.0 \\
\rowcolor{aegiszebra}
Gemini 3.6 Flash
& 79.2 & 70.9 & 71.8
& {}
& 18.4 & 5.11 & 8.91
& {}
& 47.0 & 44.5 & 47.5 & 44.6 \\
Kimi K2.6
& 78.1 & 73.0 & 72.2
& {}
& 20.1 & 6.06 & 11.2
& {}
& 35.6 & 32.6 & 44.6 & 39.1 \\
\rowcolor{aegiszebra}
Qwen3.5-397B
& 71.7 & 63.9 & 66.6
& {}
& 13.3 & 2.61 & 5.34
& {}
& 15.7 & 14.9 & 34.9 & 30.4 \\
Haiku 4.5
& 70.0 & 63.1 & 64.3
& {}
& 12.6 & 2.38 & 6.77
& {}
& 28.2 & 23.3 & 32.9 & 29.0 \\
\bottomrule
\end{tabular}%
}
\label{main_leaderboard_table}
\end{table*}

\subsection{Main Leaderboard \& Benchmark Utility}
\label{sec:main_leaderboard}

Table~\ref{main_leaderboard_table} presents performance and trajectory efficiency across 8 mainstream models evaluated with and without agent skill. \skillforgebench{} induces substantially long execution horizons: across models, with-skill runs require 15.5--47.0 total steps and 32.9--58.0 tool calls per task on average. Under the primary metric of \textit{strict-scoring completion with skill}, \textbf{Opus 4.8} leads the panel at $40.9\%$, followed by \textbf{GLM-5.2} at $36.8\%$, with \textbf{Sonnet 4.6} and \textbf{Hy3} tied at roughly $28\%$. Open-weight and proprietary models interleave throughout the ranking rather than forming disjoint capability tiers.

\paragraph{Holistic Constraint Satisfaction Reveals Hidden Capability Gaps.} 
While loose scoring evaluation suggests a deceptively narrow capability gap (Haiku 4.5-$70.0\%$ to Opus 4.8-$85.7\%$), strict evaluation isolates holistic constraint satisfaction as the primary factor dividing model tiers, expanding the performance spread dramatically (Haiku 4.5-$12.6\%$ to Opus 4.8-$40.9\%$). Non-frontier agents routinely succeed on isolated rubrics but suffer compound execution failures when all environmental constraints must hold simultaneously. Thus, the true gap between model tiers lies in satisfying all constraints simultaneously, not in isolation. This requires shifting agent optimization from individual rubric fixes to holistic constraint alignment.

\begin{wrapfigure}{R}{0.43\linewidth}
    \centering
    \includegraphics[width=.93\linewidth]{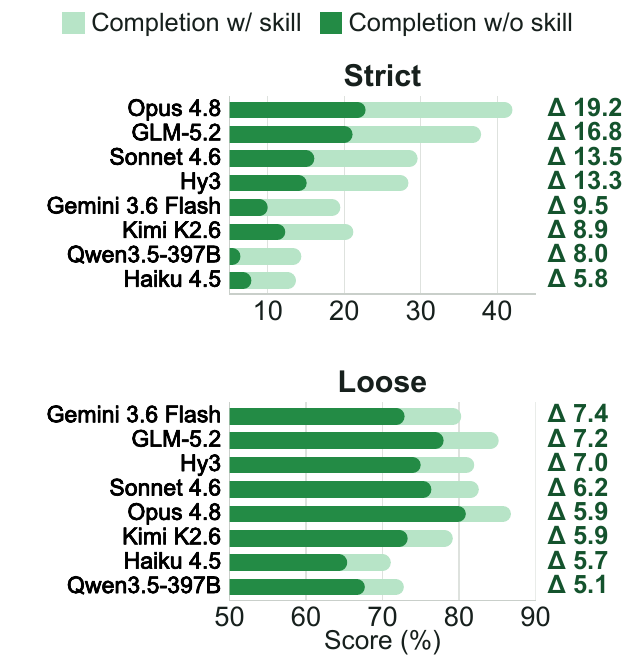}
    \caption{\textbf{Completion gains from skill.} Bars compare Completion scores with and without skill under strict and loose scoring. $\Delta = S_{\mathrm{comp}}^{\mathrm{WS}}-S_{\mathrm{comp}}^{\mathrm{NS}}$, shown to the right of each bar, denotes the Completion-score gain from skill access.}
\label{fig:completion_score_bars}
\end{wrapfigure}
\paragraph{Skill Grounding Widens Rather Than Narrows the Gap in Satisfying All Constraints.}
Comparing performance with and without skill disentangles the contributions of an agent's internal knowledge from external procedural guidance, testing whether skills act as compensatory scaffolding that uplifts weaker models.
While external skills might intuitively seem like compensatory scaffolding, \textbf{frontier models convert skill guidance into complete constraint satisfaction far more effectively than lower-tier models}. As shown in Figure~\ref{fig:completion_score_bars}, loose scoring reveals uniform gains in satisfying individual constraints across models ($\Delta = 5.1\%$ to $7.4\%$), whereas strict scoring exposes sharply divergent gains.
Skill guidance yields dramatically larger improvements for top-tier agents ($\Delta = 19.2\%$ for \textsc{Opus 4.8}; $16.8\%$ for \textsc{GLM-5.2}) than for lower-tier agents ($\Delta = 8.0\%$ for \textsc{Qwen 3.5 397B}; $5.8\%$ for \textsc{Haiku 4.5}).
This implies that external procedural guidance cannot serve as a standalone substitute for internal knowledge.
Instead, effectively parsing and executing external guidance itself demands strong internal knowledge, meaning that advancing lower-tier agents requires improving how procedural skills are grounded into execution steps (see Appendix~\ref{sec:skill-gains-vs-baseline}).

\paragraph{Divergence Between Process and Completion Constraints.}
Comparing $S_{\mathrm{comp}}$ against $S_{\mathrm{proc}}$ reveals a fundamental disparity in how models satisfy the two kinds of constraints. Under $S^{\text{loose}}$, top-performing agents achieve $S_{\mathrm{proc}}$ that systematically outpace $S_{\mathrm{comp}}$ (e.g., Opus 4.8: $87.9\%$ vs.\ $85.7\%$), whereas non-frontier models exhibit the inverse pattern (e.g., Gemini 3.6 Flash: $70.9\%$ vs.\ $79.2\%$; Haiku 4.5: $63.1\%$ vs.\ $70.0\%$). This divergence indicates that these agents frequently secure deliverables while leaving execution constraints unsatisfied, so completion-only checks overestimate how well they satisfy all constraints. Under $S^{\text{strict}}$, satisfying all execution constraints further decouples from completion rank: GLM-5.2 leads $S_{\mathrm{proc}}$ ($28.7\%$) despite ranking second in $S_{\mathrm{comp}}$ ($36.8\%$). Ultimately, this establishes that satisfying outcome constraints and satisfying execution constraints are distinct dimensions of agent capability.

\noindent\textbf{Non-Monotonic Efficiency in Long-Horizon Execution.}
Interactive agent execution exhibits a non-monotonic relationship between trajectory length and task success.
For instance, Opus 4.8 reaches top strict completion ($40.9\%$) in $28.6$ steps, whereas GLM-5.2 consumes $49\%$ more steps ($42.6$) for lower performance ($36.8\%$). Similarly, Gemini 3.6 Flash uses $3.03\times$ the step budget of Sonnet 4.6 ($47.0$ vs.\ $15.5$) yet trails by $10.1\%$, resulting in a $3.7\times$ spread in strict completion per step. This demonstrates that extra steps do not reliably translate into satisfying more constraints. Consequently, efficiency is not merely a matter of resource cost; advancing agent capabilities requires executing each step in compliance with the constraints rather than extending trajectory length. 

\subsection{Reliability of Task and Rubric Construction}
\label{sec:human_validation}

\begin{table*}[t]
\centering
\small
\setlength{\tabcolsep}{7pt}
\renewcommand{\arraystretch}{1.15}
\caption{\textbf{Independent audit of \skillforgebench{} task and rubric quality.} Values are mean ratings on a 1--5 scale. Human agreement is the percentage of paired ratings that differ by at most one point. Task quality covers 100 tasks; rubric quality covers 771 completion and 840 process criteria. The overall row aggregates all 5,133 dimension-level ratings.}
\label{tab:human_validation}
\begin{tabular}{lrrrrr}
\toprule
\textbf{Audit Target} & \textbf{\# Units} & \textbf{Expert 1} & \textbf{Expert 2} & \textbf{GPT-5.6 Sol} & \textbf{Human Agr. (\%)} \\
\midrule
Task Quality & 100 & 4.56 & 4.71 & 4.28 & 98.0 \\
Completion Rubrics & 771 & 4.47 & 4.56 & 4.43 & 96.1 \\
Process Rubrics & 840 & 4.44 & 4.49 & 4.13 & 95.4 \\
\midrule
\textbf{Overall} & 1,711 & \textbf{4.46} & \textbf{4.53} & \textbf{4.27} & \textbf{95.9} \\
\bottomrule
\end{tabular}
\end{table*}

To assess the reliability of our automated construction, two expert annotators independently review the same stratified sample, evaluating task quality in terms of realism, task--environment alignment, and solvability; completion rubrics in terms of relevance, observability, and non-redundancy; and process rubrics in terms of workflow coverage, observability, and implementation independence. As an external robustness check, GPT-5.6 Sol independently applies the same evaluation protocol without replacing or adjudicating the expert judgments. \textbf{Independent audits confirm construction reliability.} As shown in Table~\ref{tab:human_validation}, both experts consistently rate the generated tasks and rubrics highly, with overall mean ratings of 4.46 and 4.53 and strong agreement: 95.9\% of paired ratings differ by no more than one point. GPT-5.6 Sol provides a consistent external assessment, with an overall mean of 4.27 and scores above 4.0 for all three benchmark components. These results support the reliability of the generated tasks and the completion and process rubrics that check their constraints.

\subsection{Analysis}

Beyond the aggregate leaderboard, we analyze how constraint satisfaction varies with model release date and across fine-grained rubric categories. 


\begin{figure}[h]
    \centering
    \includegraphics[width=0.95\textwidth]{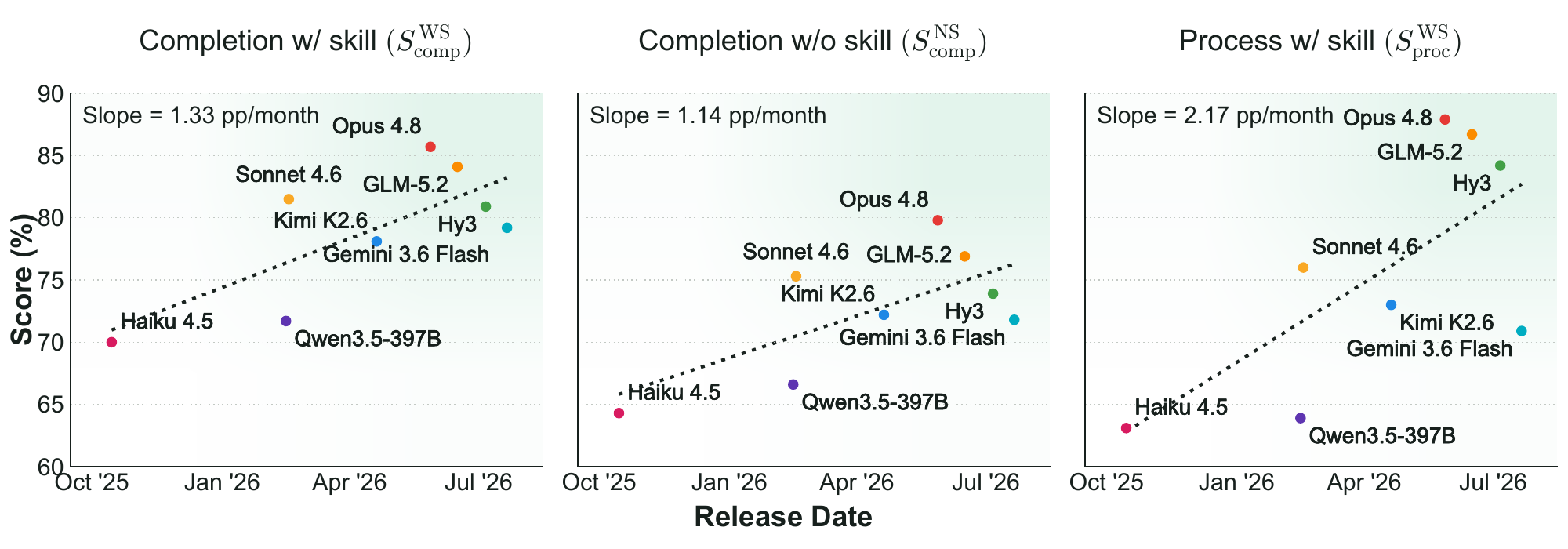}
    \caption{\textbf{Model performance versus release date under loose scoring.} Each panel plots $S_{\mathrm{comp}}^{\mathrm{WS}}$, $S_{\mathrm{comp}}^{\mathrm{NS}}$, and $S_{\mathrm{proc}}^{\mathrm{WS}}$ against official release date, with dotted lines denoting linear least-squares fits. All three scores are positively associated with release date ($r>0.68$), with fitted slopes of 1.32, 1.14, and 2.17 pp/month, respectively. Corresponding strict scoring results are reported in Figure~\ref{fig:appendix_score_release_time_strict}.}
    \label{fig:score_release_time_loose}
\end{figure}

\paragraph{Model performance over release time.}
Figure~\ref{fig:score_release_time_loose} reveals a consistent trend: newer models usually perform better under both evaluation metrics. In particular, the Completion w/ skill ($S_{\mathrm{comp}}^{\mathrm{WS}}$) and Completion w/o skill ($S_{\mathrm{comp}}^{\mathrm{NS}}$) results show medium rates of improvement over time, with slopes of $1.32$ and $1.14$ percentage points per month, respectively. In contrast, Process w/ skill ($S_{\mathrm{proc}}^{\mathrm{WS}}$) exhibits a substantially steeper upward trend, with a slope of $2.17$ percentage points per month. All the Pearson $r$ values are above $0.68$, and scores under the strict scoring exhibit a similar pattern (see Figure \ref{fig:appendix_score_release_time_strict}).
These results suggest that, for LLM agents, \textbf{execution-constraint satisfaction is more strongly associated with model release date than outcome-constraint satisfaction}, indicating that it more clearly differentiates earlier- and later-released models. 

\begin{figure}[htbp]
    \centering
    \includegraphics[width=\linewidth]{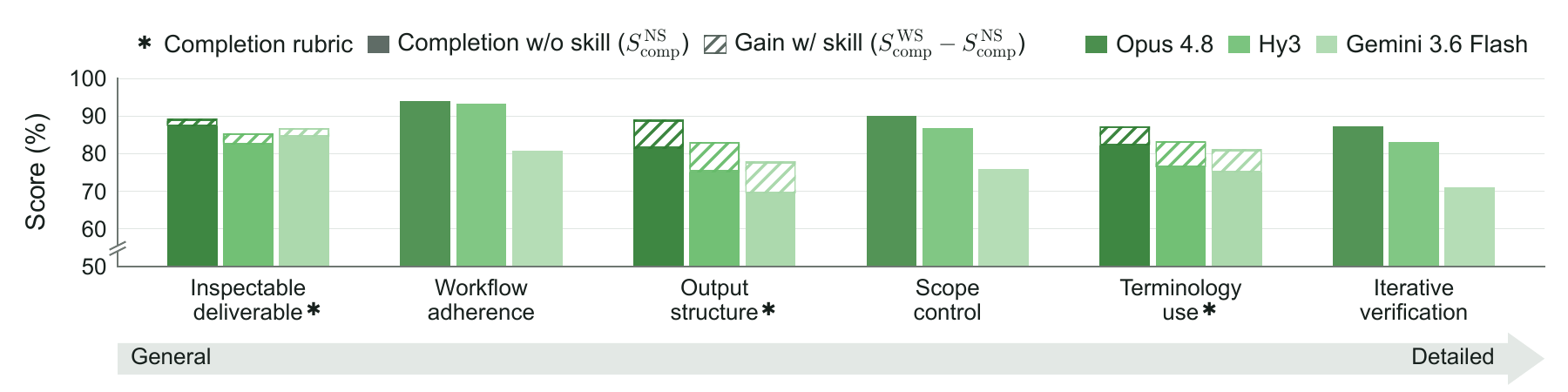}

    \caption{
    \textbf{Loose rubric-category scores for models at different leaderboard positions.}
    Opus 4.8, Hy3, and Gemini 3.6 Flash are selected as representatives of different positions in the main leaderboard.
    For the Completion categories, WS and NS denote evaluation with and without the skill, respectively;
    the Process categories are evaluated only with the skill.
    The three models achieve relatively similar Inspectable Deliverable scores but separate more clearly in the remaining categories, particularly the three Process categories.
}
    \label{loose_score_vs_rubric_cat}
\end{figure}

\paragraph{Failure modes across fine-grained rubric categories.}
\label{sec:failure_by_tier}

To analyze performance gaps across model tiers, we decompose agent pass rates into six functional rubric categories (Appendix~\ref{sec:appendix-six-categories}), separating the coarse constraint of producing the deliverable (\emph{Inspectable Deliverable}) from five finer constraints on output format, domain content, and execution. We compare three representative models spanning distinct leaderboard tiers: Opus 4.8, Hy3, and Gemini 3.6 Flash. Figure~\ref{loose_score_vs_rubric_cat} yields two key findings. First, \textbf{producing final task deliverables does not differentiate model tiers.} All three models achieve closely comparable \emph{Inspectable Deliverable} pass rates, indicating that even lower-tier agents can generate plausible task outcomes. Second, \textbf{model capabilities diverge primarily on the finer constraints.} Gemini 3.6 Flash trails significantly in procedural step adherence and empirical verification loops, whereas Hy3 maintains overall workflow structure but struggles with strict output formatting schemas and domain-specific logic. These results suggest that advancing agents requires shifting optimization from producing plausible deliverables toward satisfying fine-grained operational constraints and validation requirements.

%% file: sections/conclusion.tex
\section{Conclusion}

In this work, we present \skillforgebench{}, an agentic benchmark that leverages agent skills to bridge broad domain diversity with authentic procedural depth.
By anchoring the task lifecycle in agent skills, \textsc{SkillForgeBench} automatically synthesizes 842 executable environments, naturalistic instructions, and fine-grained rubrics that check constraints on both the final outcome and the execution.
Our empirical evaluations yield three core findings:
(1) strict, holistic constraint satisfaction defines the true capability divide between model tiers;
(2) explicit skill guidance primarily amplifies frontier model performance rather than compensating for weaker baselines; and
(3) non-frontier agents routinely sacrifice procedural discipline to force final deliverables.
Ultimately, \textsc{SkillForgeBench} demonstrates the necessity of moving beyon completion-only checks, providing a scalable foundation for trajectory-level verification and trustworthy agent development. 

%% file: sections/limitations.tex



\section*{AI Use Statement}
AI tools were used to assist with and polish manuscript writing, and to support the generation of synthetic benchmark data, including task instructions, environments, and evaluation rubrics. While AI models were used throughout the benchmark synthesis pipeline, every benchmark task was reviewed by human experts before inclusion. The benchmark construction, validation, and evaluation procedures involving AI models are described in detail in the paper. All final content, analyses, and scientific conclusions were reviewed by the authors. 





%% file: sections/appendix.tex
\input{sections/appendix_sections/skill_delta}

\newpage

\input{sections/appendix_sections/score_vs_steps}

\newpage

\input{sections/appendix_sections/strict_vs_time}

\newpage

\input{sections/appendix_sections/criteria_category}

\clearpage

\input{sections/appendix_sections/judge_sensitivity}

\clearpage

\input{sections/appendix_sections/length_bin_scores}

\clearpage

\input{sections/appendix_sections/domain_scores}

\clearpage

\input{sections/appendix_sections/example_task_comparison}

\clearpage

\input{sections/appendix_sections/prompts/appendix_prompts}

%% file: sections/appendix_sections/skill_delta.tex
\section{Skill Gains as a Function of Baseline Performance}
\label{sec:skill-gains-vs-baseline}
\begin{figure}[htbp]
    \centering
    \includegraphics[width=\linewidth]{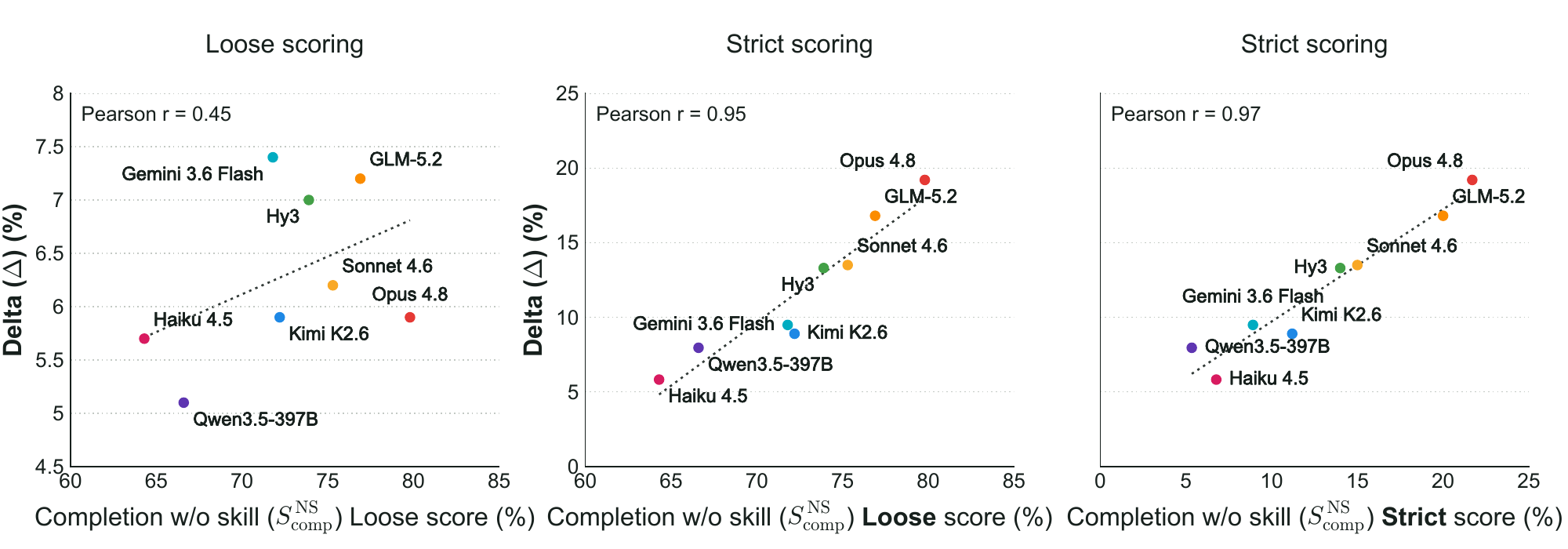}

    \caption{
    \textbf{Gains from skill access versus baseline completion performance.}
    We define the skill gain, $\Delta$, as the difference between the completion scores obtained with and without skill access:
    $\Delta = S_{\mathrm{comp}}^{\mathrm{WS}} - S_{\mathrm{comp}}^{\mathrm{NS}}$,
    where WS and NS denote the with-skill and without-skill conditions, respectively.
    The scatter plots relate the loose-score gain to the loose $S_{\mathrm{comp}}^{\mathrm{NS}}$ baseline and the strict-score gain to both loose and strict $S_{\mathrm{comp}}^{\mathrm{NS}}$ baselines.
    Each point represents one model, dotted lines show least-squares fits, and annotations report Pearson's $r$.
    Loose-score gains are comparatively uniform ($5.1 \%$--$7.4 \%$; $r=0.45$), whereas strict-score gains increase strongly with baseline performance ($r=0.95$ and $r=0.97$).
}
    \label{fig:skill-gain-vs-baseline}
\end{figure}

Figure~\ref{fig:skill-gain-vs-baseline} examines how gains $\Delta = S_{\mathrm{comp}}^{\mathrm{WS}} -
S_{\mathrm{comp}}^{\mathrm{NS}}$ from skill access vary with baseline completion no skill score $S_{\mathrm{comp}}^{\mathrm{NS}}$. Under loose scoring, $\Delta$ are comparatively uniform across models, ranging from $5.1\%$ to $7.4\%$, and are only moderately correlated with the no-skill baseline ($r=0.45$). Under strict scoring, however, $\Delta$ vary substantially, from $5.8 \%$ to $19.2 \%$, and are strongly correlated with $S_{\mathrm{comp}}^{\mathrm{NS}}$ measured under both loose ($r=0.95$) and strict ($r=0.97$) scoring. This pattern supports the main finding: skill grounding provides broadly similar partial-credit improvements across models, but models with stronger baseline capabilities are considerably more likely to use the guidance to satisfy the full set of strict requirements. Thus, skill grounding appears to help models capitalize on existing operational competence rather than uniformly compensate for weak baseline capability.

%% file: sections/appendix_sections/score_vs_steps.tex
\section{Model Performance Over Steps}

\begin{figure}[h]
    \centering
    \includegraphics[width=\textwidth]{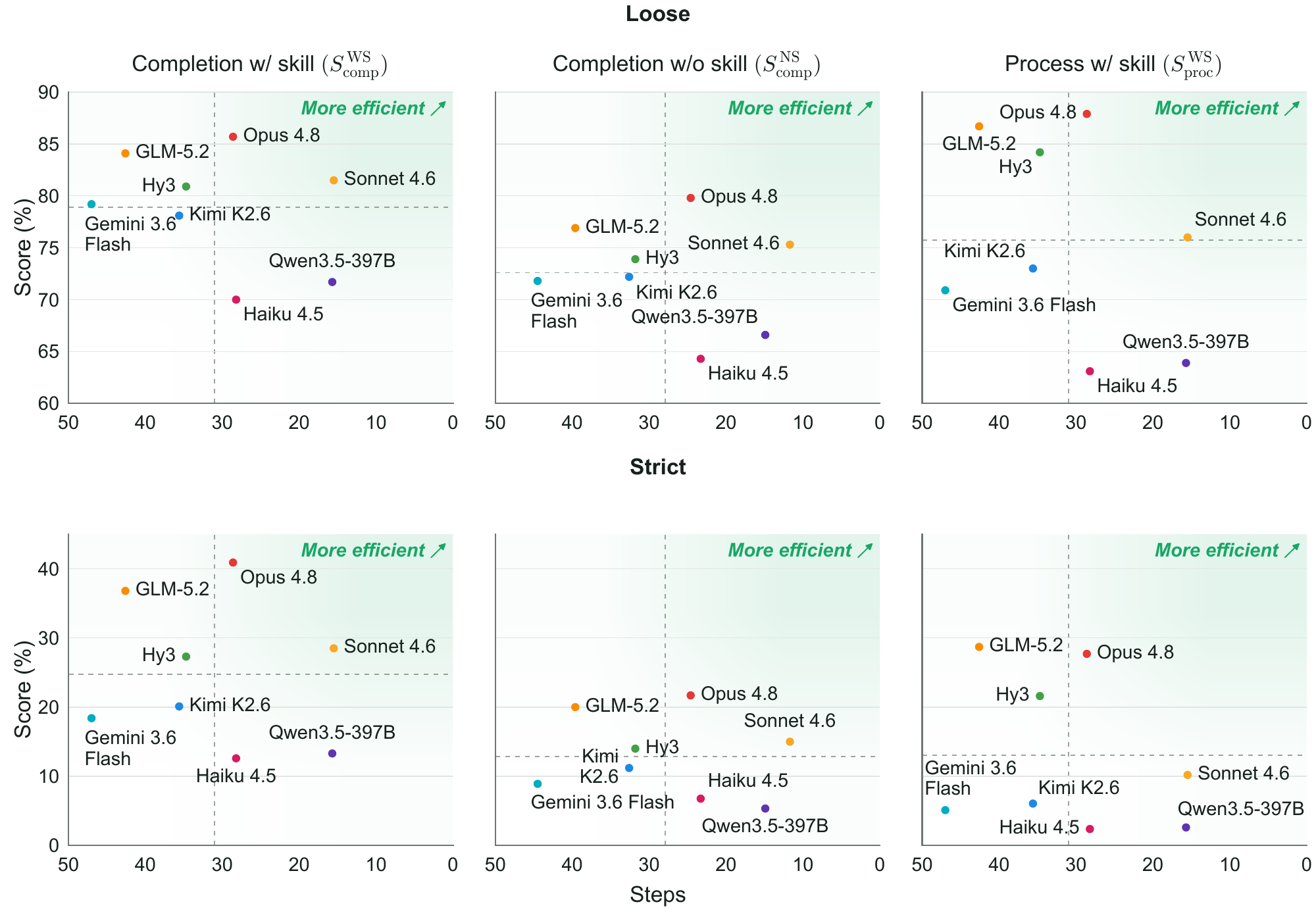}
    \caption{
    \textbf{Model performance versus trajectory length.}
    Each panel plots model performance against mean trajectory length, measured in interaction steps. Rows correspond to loose and strict scoring, while columns report Completion w/ skill ($S_{\mathrm{comp}}^{\mathrm{WS}}$), Completion w/o skill ($S_{\mathrm{comp}}^{\mathrm{NS}}$), and Process w/ skill ($S_{\mathrm{proc}}^{\mathrm{WS}}$). 
    The horizontal axis is reversed, such that shorter trajectories appear farther to the right. Dashed lines indicate the cross-model mean score and mean trajectory length; consequently, the shaded upper-right region identifies models that achieve above-average performance with fewer-than-average steps. Across metrics and scoring schemes, longer trajectories do not consistently correspond to higher performance.
}
\end{figure}

%% file: sections/appendix_sections/strict_vs_time.tex
\section{Model Performance in Strict Scheme Over Release Time}

\begin{figure}[h]
    \centering
    \includegraphics[width=\textwidth]{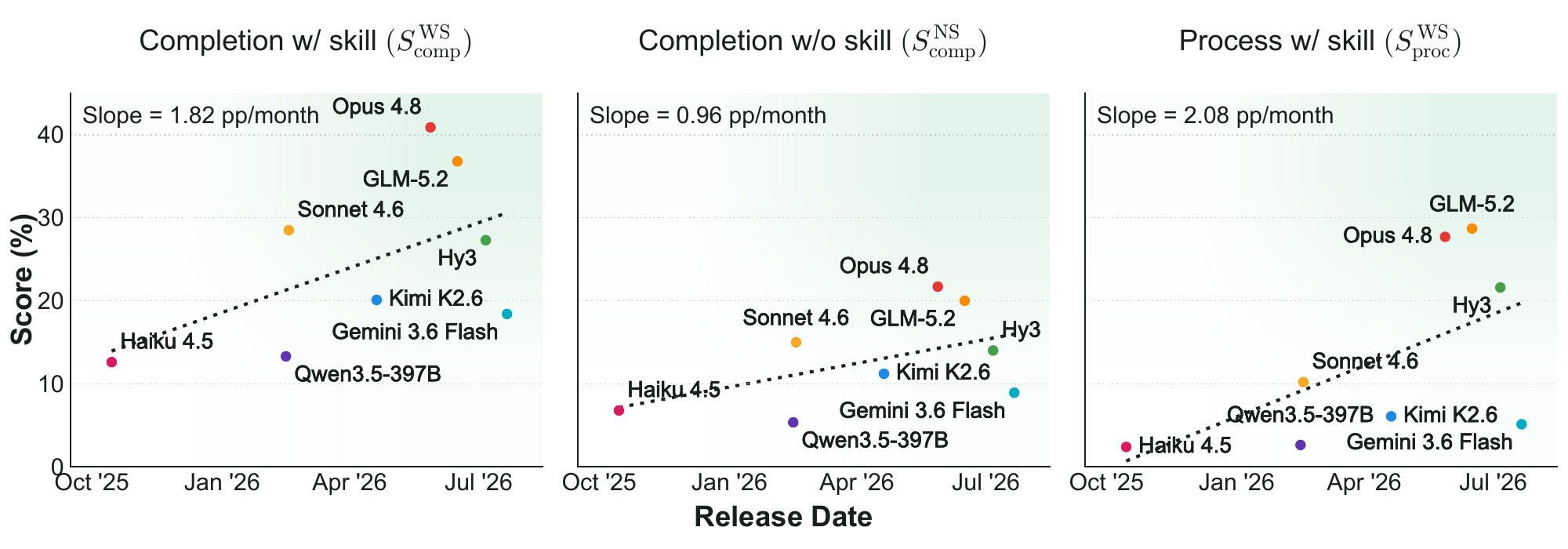}
    \caption{
    \textbf{Model performance versus release date under strict scoring.}
    Each panel plots model performance against official release date for Completion w/ skill ($S_{\mathrm{comp}}^{\mathrm{WS}}$), Completion w/o skill ($S_{\mathrm{comp}}^{\mathrm{NS}}$), and Process w/ skill ($S_{\mathrm{proc}}^{\mathrm{WS}}$). Points represent individual models, dotted lines show linear least-squares fits, and the annotated slopes are measured in percentage points (pp) per month. Release date is positively correlated with all three metrics, with Pearson's $r$ values of $0.536$, $0.502$, and $0.574$, respectively. As under loose scoring, $S_{\mathrm{proc}}^{\mathrm{WS}}$ exhibits the steepest fitted trend ($2.08$ pp/month), compared with $S_{\mathrm{comp}}^{\mathrm{WS}}$ ($1.82$ pp/month) and $S_{\mathrm{comp}}^{\mathrm{NS}}$ ($0.96$ pp/month).
    }
\label{fig:appendix_score_release_time_strict}
\end{figure}

%% file: sections/appendix_sections/criteria_category.tex
\section{Task Synthesis Details}

\subsection{Synthesis Model and GPT-Series Exclusion}

At every stage of our data-synthesis pipeline, we use GPT-5.5 in Codex with xhigh reasoning effort as the executor agent. Consequently, the synthesized tasks and rubrics may be comparatively in-distribution for GPT-series models, potentially giving them a generator-family advantage. We therefore exclude GPT-series models from evaluation.

\subsection{The List of 20 Predefined Forums}

\begin{table*}[h]
\centering
\small
\setlength{\tabcolsep}{7pt}
\renewcommand{\arraystretch}{1.15}
\caption{\textbf{Forums and their primary discussion topics.}}
\label{tab:forums}

\begin{tabularx}{\textwidth}{@{}rlX@{}}
\toprule
\textbf{ID} & \textbf{Forum name} & \textbf{Description} \\
\midrule
1 & Reddit
  & Topic-based communities called subreddits. \\
2 & Quora
  & Questions, answers, and opinion discussions. \\
3 & Stack Exchange
  & Expert Q\&A network, including Stack Overflow. \\
4 & Stack Overflow
  & Programming questions and answers. \\
5 & Tripadvisor Forums
  & Travel discussions. \\
6 & XDA Developers
  & Android, mobile devices, ROMs, and device modification. \\
7 & Hacker News
  & Technology, startups, and programming discussions. \\
8 & Digital Point
  & Search engine optimization, web marketing, and webmaster discussions. \\
9 & Warrior Forum
  & Internet marketing and online business. \\
10 & MacRumors Forums
   & Apple products and rumors. \\
11 & Mumsnet Talk
   & Parenting, family life, relationships, and lifestyle discussions. \\
12 & Linus Tech Tips Forum
   & PC building, hardware, and technology. \\
13 & Ubuntu Forums
   & Ubuntu and Linux support. \\
14 & BleepingComputer Forums
   & Malware removal and computer support. \\
15 & AVS Forum
   & Home theater, televisions, and audio/video equipment. \\
16 & Bodybuilding.com Forums
   & Fitness and bodybuilding. \\
17 & The Student Room
   & Student life, universities, and examinations. \\
18 & Wrong Planet
   & Autism, neurodiversity, and life advice. \\
19 & ResetEra
   & Gaming, entertainment, and culture. \\
20 & Mumsnet Talk
   & Parenting and lifestyle discussions. \\
\bottomrule
\end{tabularx}
\end{table*}

\subsection{Skill Domain Distribution}
\label{sec:appendix-skill-domains}

\begin{figure}[t]
    \centering
    \includegraphics[width=\linewidth]{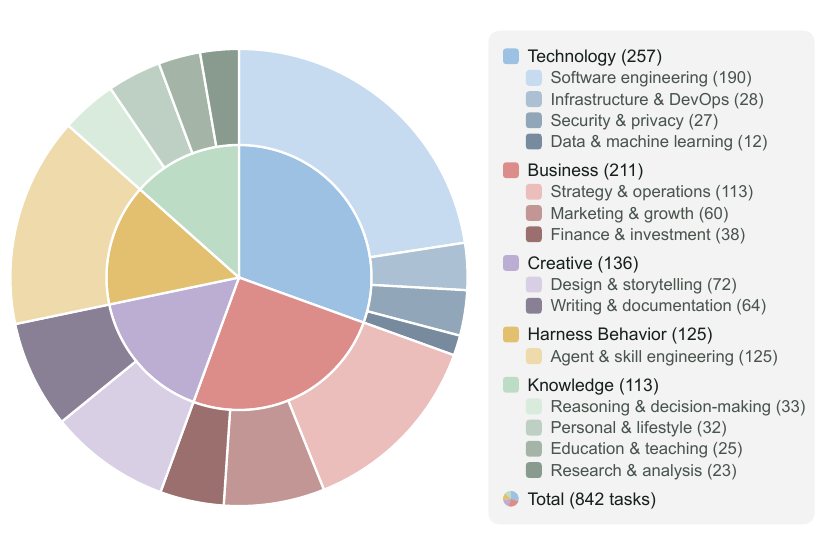}
    \caption{\textbf{Domain and sub-field composition of the 842 \skillforgebench{} tasks.} The inner ring shows the five domains and the outer ring their 14 sub-fields, each sized by its number of tasks; since each task comes from a distinct skill, the same numbers hold for skills.}
    \label{fig:task_domain_composition}
\end{figure}

Figure~\ref{fig:task_domain_composition} shows how the 842 tasks divide into domains and sub-fields, and Table~\ref{tab:skill_domains} gives the exact shares and the fine-grained topics within each sub-field.

\input{sections/appendix_sections/skill_domains}

\subsection{Criteria Category}

\label{sec:appendix-six-categories}
During contrastive rubric construction, each criterion is assigned to a category. There are three categories in completion rubrics and three in process rubrics, specifically defined as follows.

\begin{enumerate}
    \item \textbf{Inspectable Deliverable:} Assesses whether the agent produces a persistent and inspectable deliverable, such as an executable proof of concept, test, before-and-after demonstration, or report, in the form prescribed by the skill.
    \item \textbf{Output Structure:} Assesses whether the agent follows the required output structure and reporting conventions, including mandated summaries, scores, counts, or verdict formats.
    \item \textbf{Terminology Use:} Assesses whether the agent correctly applies the skill’s domain-specific concepts in practice, rather than merely repeating relevant terminology.
\end{enumerate}

\begin{enumerate}
    \item \textbf{Workflow Adherence:} Assesses whether the agent visibly follows the skill’s named workflow phases in the prescribed order.
    \item \textbf{Iterative Verification:} Assesses whether the agent probes the issue before acting, verifies the result afterward, and uses fail–diagnose–fix cycles when necessary.
    \item \textbf{Scope Control:} Assesses whether the agent checks the skill’s applicability and task readiness, remains within the appropriate scope, and resists overreach or unsupported rationalization.
\end{enumerate}

\input{sections/appendix_sections/rubric_category_table}

\input{sections/appendix_sections/strict_rubric_category_figure}

\input{sections/appendix_sections/task_stats}

%% file: sections/appendix_sections/skill_domains.tex
\begin{table*}[t]
\centering
\footnotesize
\setlength{\tabcolsep}{3.5pt}
\renewcommand{\arraystretch}{1.12}
\caption{
\textbf{Domain distribution of the skills behind the 842 \skillforgebench{} tasks.}
Each task is synthesized from a distinct skill, so skill counts equal task counts.
The \% column gives the share of the 842 skills in each row.
\textit{Topics} is the number of distinct fine-grained topics a sub-field covers (83 in total).
The last column lists the three most frequent topics with their skill counts; ties are broken by taxonomy order.
}
\label{tab:skill_domains}
\begin{tabularx}{\textwidth}{l r r r Y}
\toprule
\textbf{Domain / sub-field} & \textbf{Skills} & \textbf{\%} & \textbf{Topics} & \textbf{Most frequent topics (skills)} \\
\midrule
\rowcolor{aegislight}
\textbf{Technology} & \textbf{257} & \textbf{30.5} & \textbf{27} & \\
\hspace{0.8em}Software engineering & 190 & 22.6 & 10 & Code implementation \& refactoring (88); architecture \& design decisions (31); debugging \& root-cause analysis (19) \\
\hspace{0.8em}Infrastructure \& DevOps & 28 & 3.3 & 6 & Cloud platforms \& IaC (8); observability \& alerting (7); CI/CD \& release management (6) \\
\hspace{0.8em}Security \& privacy & 27 & 3.2 & 6 & Security operations \& incident response (10); threat modeling \& security architecture (7); privacy \& compliance auditing (4) \\
\hspace{0.8em}Data \& machine learning & 12 & 1.4 & 5 & Data processing \& pipelines (4); querying \& SQL (3); machine \& deep learning (3) \\
\addlinespace[2pt]
\rowcolor{aegislight}
\textbf{Business} & \textbf{211} & \textbf{25.1} & \textbf{17} & \\
\hspace{0.8em}Strategy \& operations & 113 & 13.4 & 6 & Strategy \& competitive analysis (43); product requirements \& planning (22); organization, talent \& leadership (18) \\
\hspace{0.8em}Marketing \& growth & 60 & 7.1 & 5 & Social-media operations (25); advertising \& marketing copy (16); SEO, GEO \& ASO (10) \\
\hspace{0.8em}Finance \& investment & 38 & 4.5 & 6 & Macro \& commodity research (13); equity analysis (9); market-data APIs \& tools (5) \\
\addlinespace[2pt]
\rowcolor{aegislight}
\textbf{Creative} & \textbf{136} & \textbf{16.2} & \textbf{12} & \\
\hspace{0.8em}Design \& storytelling & 72 & 8.6 & 6 & UI/UX \& visual design (21); fiction, scripts \& stories (16); infographics \& diagrams (11) \\
\hspace{0.8em}Writing \& documentation & 64 & 7.6 & 6 & Technical \& product documentation (38); office documents \& presentations (12); translation \& language editing (5) \\
\addlinespace[2pt]
\rowcolor{aegislight}
\textbf{Harness Behavior} & \textbf{125} & \textbf{14.8} & \textbf{6} & \\
\hspace{0.8em}Agent \& skill engineering & 125 & 14.8 & 6 & Skill authoring \& governance (68); prompt engineering (19); tool \& MCP building (12) \\
\addlinespace[2pt]
\rowcolor{aegislight}
\textbf{Knowledge} & \textbf{113} & \textbf{13.4} & \textbf{21} & \\
\hspace{0.8em}Reasoning \& decision-making & 33 & 3.9 & 6 & Decision frameworks \& scenario analysis (12); first-principles decomposition (7); debate \& red-teaming (6) \\
\hspace{0.8em}Personal \& lifestyle & 32 & 3.8 & 6 & Personal growth \& self-management (10); metaphysics, divination \& spirituality (6); healthcare \& life sciences (5) \\
\hspace{0.8em}Education \& teaching & 25 & 3.0 & 4 & Course \& teaching materials (15); tutoring \& problem solving (6); pedagogy \& learning strategies (2) \\
\hspace{0.8em}Research \& analysis & 23 & 2.7 & 5 & Academic writing \& submission (11); deep research \& retrieval (4); fact-checking \& argument auditing (4) \\
\midrule
\textbf{Total} & \textbf{842} & \textbf{100.0} & \textbf{83} & 5 domains, 14 sub-fields \\
\bottomrule
\end{tabularx}
\end{table*}

%% file: sections/appendix_sections/rubric_category_table.tex
\begin{table*}[t]
\centering
\small
\setlength{\tabcolsep}{3.5pt}
\renewcommand{\arraystretch}{1.2}
\caption{
\textbf{Rubric-category scores under loose and strict scoring.}
Category-level averages are reported over the common set of 842 evaluation tasks.
The completion metric ($S_{\mathrm{comp}}$) comprises Inspectable Deliverable,
Output Structure, and Terminology Use, evaluated both with and
without the skill. The process metric ($S_{\mathrm{proc}}$) comprises
Workflow Adherence, Iterative Verification, and Scope Control, evaluated
with the skill. Under loose scoring, each task--category instance receives the
weighted mean of its applicable criteria. Under strict scoring, it receives
full credit only when every applicable criterion attains its maximum score,
and zero otherwise. All reported values are percentages.
\textbf{Bold} marks the highest score, and \underline{underlining} marks the
second-highest score in each column.
}
\resizebox{\textwidth}{!}{%
\begin{tabular}{@{}l ccccccccc p{6pt} ccccccccc@{}}
\toprule
& \multicolumn{9}{c}{\cellcolor{aegisgreen}\textcolor{white}{\textbf{Loose}}}
& {}
& \multicolumn{9}{c}{\cellcolor{aegisgreen!80!black}\textcolor{white}{\textbf{Strict}}} \\
\cmidrule(lr){2-10} \cmidrule(lr){12-20}
\textbf{Model}
& \multicolumn{6}{c}{\cellcolor{aegislight}\footnotesize\textbf{$S_{\text{comp}}$}}
& \multicolumn{3}{c}{\cellcolor{aegislight}\footnotesize\textbf{$S_{\text{proc}}$}}
& {}
& \multicolumn{6}{c}{\cellcolor{aegislight}\footnotesize\textbf{$S_{\text{comp}}$}}
& \multicolumn{3}{c}{\cellcolor{aegislight}\footnotesize\textbf{$S_{\text{proc}}$}} \\
\cmidrule(lr){2-7} \cmidrule(lr){8-10} \cmidrule(lr){12-17} \cmidrule(lr){18-20}
& \multicolumn{2}{c}{\cellcolor{aegislight}\scriptsize\textbf{\shortstack{Inspectable\\Deliverable}}}
& \multicolumn{2}{c}{\cellcolor{aegislight}\scriptsize\textbf{\shortstack{Output\\Structure}}}
& \multicolumn{2}{c}{\cellcolor{aegislight}\scriptsize\textbf{\shortstack{Terminology\\Use}}}
& \cellcolor{aegislight}\scriptsize\textbf{\shortstack{Workflow\\Adherence}}
& \cellcolor{aegislight}\scriptsize\textbf{\shortstack{Iterative\\Verification}}
& \cellcolor{aegislight}\scriptsize\textbf{\shortstack{Scope\\Control}}
& {}
& \multicolumn{2}{c}{\cellcolor{aegislight}\scriptsize\textbf{\shortstack{Inspectable\\Deliverable}}}
& \multicolumn{2}{c}{\cellcolor{aegislight}\scriptsize\textbf{\shortstack{Output\\Structure}}}
& \multicolumn{2}{c}{\cellcolor{aegislight}\scriptsize\textbf{\shortstack{Terminology\\Use}}}
& \cellcolor{aegislight}\scriptsize\textbf{\shortstack{Workflow\\Adherence}}
& \cellcolor{aegislight}\scriptsize\textbf{\shortstack{Iterative\\Verification}}
& \cellcolor{aegislight}\scriptsize\textbf{\shortstack{Scope\\Control}} \\
\cmidrule(lr){2-3} \cmidrule(lr){4-5} \cmidrule(lr){6-7}
\cmidrule(lr){8-8} \cmidrule(lr){9-9} \cmidrule(lr){10-10}
\cmidrule(lr){12-13} \cmidrule(lr){14-15} \cmidrule(lr){16-17}
\cmidrule(lr){18-18} \cmidrule(lr){19-19} \cmidrule(lr){20-20}
& \cellcolor{aegislight}\scriptsize w/ skill
& \cellcolor{aegislight}\scriptsize wo/ skill
& \cellcolor{aegislight}\scriptsize w/ skill
& \cellcolor{aegislight}\scriptsize wo/ skill
& \cellcolor{aegislight}\scriptsize w/ skill
& \cellcolor{aegislight}\scriptsize wo/ skill
& \cellcolor{aegislight}\scriptsize w/ skill
& \cellcolor{aegislight}\scriptsize w/ skill
& \cellcolor{aegislight}\scriptsize w/ skill
& {}
& \cellcolor{aegislight}\scriptsize w/ skill
& \cellcolor{aegislight}\scriptsize wo/ skill
& \cellcolor{aegislight}\scriptsize w/ skill
& \cellcolor{aegislight}\scriptsize wo/ skill
& \cellcolor{aegislight}\scriptsize w/ skill
& \cellcolor{aegislight}\scriptsize wo/ skill
& \cellcolor{aegislight}\scriptsize w/ skill
& \cellcolor{aegislight}\scriptsize w/ skill
& \cellcolor{aegislight}\scriptsize w/ skill \\
\midrule
\rowcolor{aegiszebra}
Opus 4.8
& \textbf{89.2} & \textbf{87.5} & \textbf{88.9} & \textbf{81.5} & \textbf{87.1} & \textbf{82.2} & \underline{94.1} & \underline{87.3} & \textbf{90.1}
& {}
& \textbf{72.2} & \textbf{66.0} & \textbf{69.4} & \textbf{52.6} & \textbf{57.7} & \textbf{45.5} & \underline{75.1} & \underline{50.1} & \textbf{69.8} \\
GLM-5.2
& \underline{88.3} & \underline{85.9} & \underline{85.5} & \underline{76.8} & \underline{85.6} & \underline{80.3} & \textbf{96.3} & \textbf{90.1} & 80.7
& {}
& \underline{69.3} & \underline{62.0} & \underline{64.6} & \underline{46.6} & \underline{55.7} & \underline{44.7} & \textbf{86.2} & \textbf{63.2} & 48.7 \\
\rowcolor{aegiszebra}
Hy3
& 85.3 & 82.6 & 83.0 & 75.2 & 83.2 & 76.5 & 93.3 & 83.1 & \underline{86.9}
& {}
& 63.5 & 57.0 & 58.9 & 42.3 & 50.1 & 34.4 & 74.2 & 41.5 & \underline{62.8} \\
Sonnet 4.6
& 86.9 & 84.8 & 82.9 & 75.5 & 84.0 & 78.7 & 85.6 & 74.8 & 82.9
& {}
& 66.3 & 59.7 & 55.8 & 43.2 & 49.8 & 38.7 & 52.0 & 28.5 & 53.4 \\
\rowcolor{aegiszebra}
Haiku 4.5
& 78.3 & 77.4 & 73.4 & 66.3 & 71.6 & 67.3 & 84.6 & 60.8 & 67.6
& {}
& 52.9 & 48.6 & 40.8 & 27.8 & 25.3 & 19.5 & 46.5 & 11.5 & 26.9 \\
Kimi K2.6
& 83.7 & 81.7 & 80.4 & 72.9 & 79.8 & 75.5 & 83.8 & 69.7 & 83.4
& {}
& 58.8 & 53.8 & 51.0 & 37.8 & 39.2 & 32.0 & 49.5 & 19.9 & 54.2 \\
\rowcolor{aegiszebra}
Qwen3.5-397B
& 79.9 & 79.0 & 75.4 & 68.3 & 73.6 & 69.1 & 83.2 & 56.7 & 77.0
& {}
& 53.9 & 49.3 & 40.8 & 29.3 & 27.1 & 21.0 & 44.4 & 7.02 & 42.3 \\
Gemini 3.6 Flash
& 86.6 & 84.6 & 77.8 & 69.5 & 81.0 & 75.1 & 80.8 & 71.1 & 76.0
& {}
& 63.2 & 58.0 & 46.8 & 32.1 & 40.2 & 27.8 & 43.3 & 18.8 & 42.5 \\
\bottomrule
\end{tabular}%
}
\label{tab:rubric-category-scores}
\end{table*}

%% file: sections/appendix_sections/strict_rubric_category_figure.tex
\subsection{Strict rubric-category scores across capability tiers}

\begin{figure}[htbp]
    \centering
    \includegraphics[width=\linewidth]{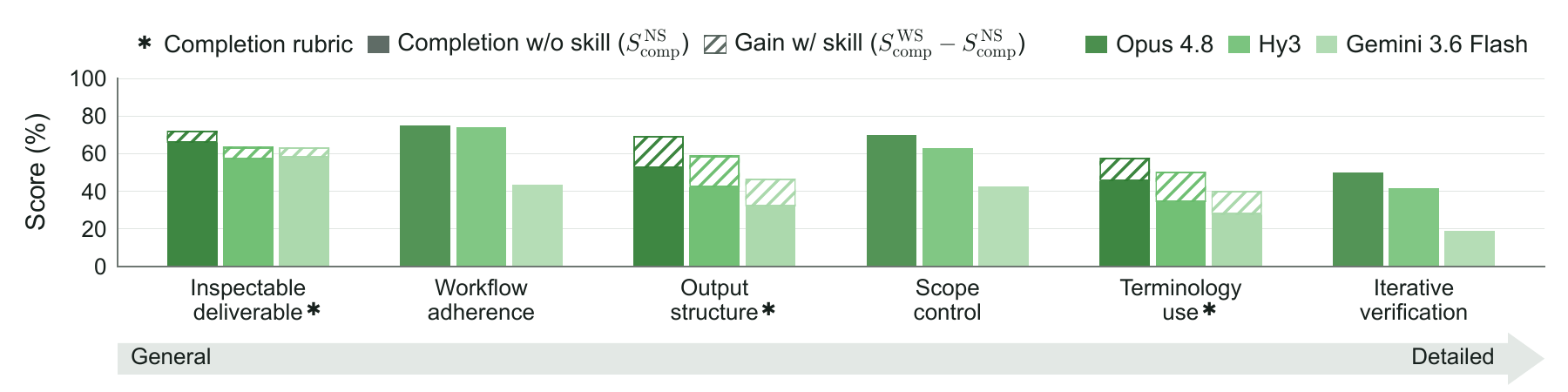}
    \caption{
\textbf{Strict rubric-category scores for models at different leaderboard positions.}
Opus 4.8, Hy3, and Gemini 3.6 Flash are selected as representatives of different positions in the main leaderboard. For the Completion categories, WS and NS denote evaluation with and without the skill, respectively; the Process categories are evaluated only with the skill. Under strict scoring, credit is awarded only when all applicable requirements within a category are fully satisfied. Gemini 3.6 Flash remains comparable to Hy3 in Inspectable Deliverable but trails both Hy3 and Opus 4.8 in Output Structure, Terminology Use, and all three Process categories. Full numerical results are reported in
Table~\ref{tab:rubric-category-scores}.
}
\label{fig:strict-rubric-category-scores}
\end{figure}

%% file: sections/appendix_sections/task_stats.tex
\begin{table*}[t]
\centering
\caption{\textbf{Descriptive statistics for the 842-task cohort.}}
\label{tab:task-stats-842}

{\small
\setlength{\tabcolsep}{7pt}
\renewcommand{\arraystretch}{1.15}

\begin{tabular}{@{}lrrr@{}}
\toprule
\textbf{Measure}
& \textbf{Mean}
& \textbf{Population SD}
& \textbf{Median} \\
\addlinespace[3pt]

\toprule

\rowcolor{aegiszebra}
Skill files per task
& 11.32
& 28.90
& 6 \\

Distinct file types per task
& 9.46
& 11.17
& 4 \\

\rowcolor{aegiszebra}
Logical task size (MiB)
& 13.67
& 27.32
& 3.37 \\

Files per task
& 172.89
& 434.80
& 10 \\

\rowcolor{aegiszebra}
Instruction length (characters)
& 1,333.05
& 311.37
& 1,306.5 \\

\bottomrule
\end{tabular}
}
\end{table*}

%% file: sections/appendix_sections/judge_sensitivity.tex

\section{Reliability of the LLM Judge}
\label{app:judge_sensitivity}

Our judge for the main evaluation is GPT-5.5 with xhigh reasoning effort running in Codex.
We test this judge on a sample of 50 tasks in two ways.
First, we compare the judge with human judgments (Section~\ref{app:judge_human}).
Second, we judge the same executions again with a judge model from a different model family (Section~\ref{app:judge_model}).

\subsection{Agreement with Human Judgments}
\label{app:judge_human}

\textbf{The judge agrees with human judgments on 96\% of decisions (Cohen's $\kappa = 0.90$--$0.91$).}
We compared two independent runs of the judge (judge-1 and judge-2) with human judgments on the executions of Opus 4.8 for the 50 tasks.
In total, the comparison covers 941 decisions.
Table~\ref{tab:judge_human_agreement} shows that each judge run disagrees with the human on about as many decisions as the two runs disagree with each other.

\begin{table}[h!]
\centering
\small
\setlength{\tabcolsep}{3.5pt}
\renewcommand{\arraystretch}{1.2}
\caption{
\textbf{Agreement between the judge and human judgments.}
Disagreement is the share of the 941 decisions on the executions of Opus 4.8 where the two sides give different values.
All 95\% confidence intervals are bootstrapped over tasks (2,000 resamples, seed 0).
}
\begin{tabularx}{0.75\textwidth}{@{}l >{\centering\arraybackslash}X >{\centering\arraybackslash}X@{}}
\toprule
\textbf{Pair}
& \cellcolor{aegislight}\footnotesize\textbf{Disagreement (\%) [95\% CI]}
& \cellcolor{aegislight}\footnotesize\textbf{Cohen's $\kappa$ [95\% CI]} \\
\midrule
\rowcolor{aegiszebra}
judge-1 vs judge-2 & 3.29 [2.06, 4.76] & 0.907 [0.872, 0.939] \\
judge-1 vs human   & 3.61 [2.22, 5.39] & 0.901 [0.862, 0.935] \\
\rowcolor{aegiszebra}
judge-2 vs human   & 3.40 [2.19, 4.85] & 0.905 [0.871, 0.935] \\
\bottomrule
\end{tabularx}
\label{tab:judge_human_agreement}
\end{table}

\textbf{Across process and completion rubrics, strict scores from the judge closely match human judgments.}
Table~\ref{tab:judge_human_strict} shows that, in every setting, the human strict score is close to at least one judge run.
For $S_{\text{comp}}$ without skill, human judgments match judge-1 exactly (28.0\%) and differ from judge-2 by only 2.0 points (30.0\%).
For $S_{\text{comp}}$ with skill, human judgments stay within 2.0 points of both runs (54.0\% vs.\ 52.0\% and 56.0\%).
For $S_{\text{proc}}$ with skill, the difference from judge-1 is larger than for completion (6.0 points, 26.0\% vs.\ 20.0\%), but judge-2 matches human judgments exactly (20.0\%).

\begin{table}[h!]
\centering
\small
\setlength{\tabcolsep}{3.5pt}
\renewcommand{\arraystretch}{1.2}
\caption{
\textbf{Strict scores of Opus 4.8 under the two judge runs and human judgments.}
Strict scores (\%) of Completion ($S_{\text{comp}}$), evaluated with and without skill, and Process ($S_{\text{proc}}$), evaluated with skill.
All scores are computed over the 50 tasks.
}
\begin{tabularx}{0.6\textwidth}{@{}l >{\centering\arraybackslash}X >{\centering\arraybackslash}X >{\centering\arraybackslash}X@{}}
\toprule
\textbf{Setting}
& \cellcolor{aegislight}\footnotesize\textbf{judge-1}
& \cellcolor{aegislight}\footnotesize\textbf{judge-2}
& \cellcolor{aegislight}\footnotesize\textbf{Human} \\
\midrule
\rowcolor{aegiszebra}
$S_{\text{comp}}$ w/o skill & 28.0 & 30.0 & 28.0 \\
$S_{\text{comp}}$ w/ skill  & 52.0 & 56.0 & 54.0 \\
\rowcolor{aegiszebra}
$S_{\text{proc}}$ w/ skill  & 26.0 & 20.0 & 20.0 \\
\bottomrule
\end{tabularx}
\label{tab:judge_human_strict}
\end{table}

\subsection{Sensitivity to the Judge Model}
\label{app:judge_model}

To test sensitivity to the judge model, we independently judge the executions of Opus 4.8, GLM-5.2, and Gemini 3.6 Flash on the same 50 tasks with Sonnet 5.5 in Claude Code, which comes from a completely different model family.
Table~\ref{tab:judge_model_scores} shows that the model ranking (Opus 4.8 $>$ GLM-5.2 $>$ Gemini 3.6 Flash) does not change between the two judge models for any of the six scores, although absolute scores judged by Sonnet 5.5 are systematically lower.
These results suggest that our rubrics are robust under completely different judge models.

\begin{table}[h!]
\centering
\small
\setlength{\tabcolsep}{3.5pt}
\renewcommand{\arraystretch}{1.2}
\caption{
\textbf{Scores under GPT-5.5 and Sonnet 5.5 as the judge.}
Task-level mean scores (\%) are reported under loose and strict scoring for Completion ($S_{\text{comp}}$), evaluated with and without skill, and Process ($S_{\text{proc}}$), evaluated with skill, on the same 50 tasks.
GPT-5.5 scores are from judge-1 for Opus 4.8 and GLM-5.2 and from the main-evaluation run for Gemini 3.6 Flash.
\textbf{Bold} indicates the highest score and \underline{underlining} indicates the second-highest score in each row under each judge.
}
\begin{tabularx}{\textwidth}{@{}l l >{\centering\arraybackslash}X >{\centering\arraybackslash}X >{\centering\arraybackslash}X p{6pt} >{\centering\arraybackslash}X >{\centering\arraybackslash}X >{\centering\arraybackslash}X@{}}
\toprule
& & \multicolumn{3}{c}{\cellcolor{aegisgreen}\textcolor{white}{\textbf{GPT-5.5 as judge}}}
& {}
& \multicolumn{3}{c}{\cellcolor{aegisgreen!80!black}\textcolor{white}{\textbf{Sonnet 5.5 as judge}}} \\
\cmidrule(lr){3-5} \cmidrule(lr){7-9}
\textbf{Scoring} & \textbf{Setting}
& \cellcolor{aegislight}\footnotesize\textbf{Opus 4.8}
& \cellcolor{aegislight}\footnotesize\textbf{GLM-5.2}
& \cellcolor{aegislight}\footnotesize\textbf{Gemini 3.6 Flash}
& {}
& \cellcolor{aegislight}\footnotesize\textbf{Opus 4.8}
& \cellcolor{aegislight}\footnotesize\textbf{GLM-5.2}
& \cellcolor{aegislight}\footnotesize\textbf{Gemini 3.6 Flash} \\
\midrule
\rowcolor{aegiszebra}
& $S_{\text{comp}}$ w/o skill & \textbf{78.4} & \underline{73.1} & 71.5 & {} & \textbf{67.9} & \underline{65.8} & 60.7 \\
& $S_{\text{comp}}$ w/ skill  & \textbf{82.2} & \underline{79.0} & 74.4 & {} & \textbf{78.0} & \underline{73.4} & 66.2 \\
\rowcolor{aegiszebra}
\multirow{-3}{*}{Loose}
& $S_{\text{proc}}$ w/ skill  & \textbf{86.3} & \underline{80.9} & 73.6 & {} & \textbf{80.1} & \underline{79.6} & 67.4 \\
\midrule
& $S_{\text{comp}}$ w/o skill & \textbf{28.0} & \underline{26.0} & 16.0 & {} & \textbf{12.0} & \underline{10.0} & 2.0 \\
\rowcolor{aegiszebra}
& $S_{\text{comp}}$ w/ skill  & \textbf{52.0} & \underline{42.0} & 20.0 & {} & \textbf{30.0} & \underline{28.0} & 8.0 \\
\multirow{-3}{*}{Strict}
& $S_{\text{proc}}$ w/ skill  & \textbf{26.0} & \underline{24.0} & 10.0 & {} & \textbf{8.0}  & \underline{6.0}  & 4.0 \\
\bottomrule
\end{tabularx}
\label{tab:judge_model_scores}
\end{table}

%% file: sections/appendix_sections/length_bin_scores.tex
\section{Scores within Checklist-Length Bins}
\label{app:length_bins}


\begin{table*}[h!]
\centering
\small
\setlength{\tabcolsep}{3.5pt}
\renewcommand{\arraystretch}{1.2}
\caption{
\textbf{Strict $S_{\text{comp}}$ w/ skill within checklist-length bins.}
Strict scores (\%) of Completion ($S_{\text{comp}}$), evaluated with skill, are reported over all tasks and over the tasks whose completion rubric set has $n$ criteria; the number of tasks is given under each column.
Kendall $\tau$ is the rank correlation between the model ranking in each bin and the ranking over all tasks.
\textbf{Bold} indicates the highest score and \underline{underlining} indicates the second-highest score in each column.
}
\begin{tabularx}{\textwidth}{@{}l >{\centering\arraybackslash}X p{6pt} >{\centering\arraybackslash}X >{\centering\arraybackslash}X >{\centering\arraybackslash}X >{\centering\arraybackslash}X >{\centering\arraybackslash}X@{}}
\toprule
& \cellcolor{aegisgreen}\textcolor{white}{\textbf{All}}
& {}
& \multicolumn{5}{c}{\cellcolor{aegisgreen!80!black}\textcolor{white}{\textbf{Checklist length $n$}}} \\
\cmidrule(lr){2-2} \cmidrule(lr){4-8}
\textbf{Model}
& \cellcolor{aegislight}\footnotesize\textbf{$S_{\text{comp}}$}
& {}
& \cellcolor{aegislight}\footnotesize\textbf{$1$--$5$}
& \cellcolor{aegislight}\footnotesize\textbf{$6$--$7$}
& \cellcolor{aegislight}\footnotesize\textbf{$8$}
& \cellcolor{aegislight}\footnotesize\textbf{$9$}
& \cellcolor{aegislight}\footnotesize\textbf{$\geq 10$} \\
& \cellcolor{aegislight}\footnotesize 842 tasks
& {}
& \cellcolor{aegislight}\footnotesize 186 tasks
& \cellcolor{aegislight}\footnotesize 159 tasks
& \cellcolor{aegislight}\footnotesize 191 tasks
& \cellcolor{aegislight}\footnotesize 198 tasks
& \cellcolor{aegislight}\footnotesize 108 tasks \\
\midrule
\rowcolor{aegiszebra}
Opus 4.8
& \textbf{40.9}
& {}
& \textbf{48.4} & \underline{32.1} & \textbf{38.2} & \textbf{43.4} & \textbf{40.7} \\
GLM-5.2
& \underline{36.8}
& {}
& \underline{33.9} & \textbf{33.3} & \underline{37.7} & \underline{40.4} & \underline{38.9} \\
\rowcolor{aegiszebra}
Sonnet 4.6
& 28.5
& {}
& 31.2 & 22.0 & 30.4 & 30.3 & 26.9 \\
Hy3
& 27.3
& {}
& 31.7 & 22.0 & 33.0 & 23.2 & 25.0 \\
\rowcolor{aegiszebra}
Gemini 3.6 Flash
& 18.4
& {}
& 25.8 & 15.1 & 18.8 & 16.2 & 13.9 \\
Kimi K2.6
& 20.1
& {}
& 29.6 & 16.4 & 18.3 & 16.2 & 19.4 \\
\rowcolor{aegiszebra}
Qwen3.5-397B
& 13.3
& {}
& 21.5 & 10.7 & 14.1 & 10.6 & 6.5 \\
Haiku 4.5
& 12.6
& {}
& 26.3 & 9.4 & 11.5 & 6.6 & 6.5 \\
\midrule
Kendall $\tau$
& --
& {}
& 0.79 & 0.93 & 0.86 & 0.93 & 1.00 \\
\bottomrule
\end{tabularx}
\label{tab:length_bin_comp_ws}
\end{table*}

\begin{table*}[h!]
\centering
\small
\setlength{\tabcolsep}{3.5pt}
\renewcommand{\arraystretch}{1.2}
\caption{
\textbf{Strict $S_{\text{comp}}$ w/o skill within checklist-length bins.}
Strict scores (\%) of Completion ($S_{\text{comp}}$), evaluated without skill, are reported over all tasks and over the tasks whose completion rubric set has $n$ criteria; the number of tasks is given under each column.
Kendall $\tau$ is the rank correlation between the model ranking in each bin and the ranking over all tasks.
\textbf{Bold} indicates the highest score and \underline{underlining} indicates the second-highest score in each column.
}
\begin{tabularx}{\textwidth}{@{}l >{\centering\arraybackslash}X p{6pt} >{\centering\arraybackslash}X >{\centering\arraybackslash}X >{\centering\arraybackslash}X >{\centering\arraybackslash}X >{\centering\arraybackslash}X@{}}
\toprule
& \cellcolor{aegisgreen}\textcolor{white}{\textbf{All}}
& {}
& \multicolumn{5}{c}{\cellcolor{aegisgreen!80!black}\textcolor{white}{\textbf{Checklist length $n$}}} \\
\cmidrule(lr){2-2} \cmidrule(lr){4-8}
\textbf{Model}
& \cellcolor{aegislight}\footnotesize\textbf{$S_{\text{comp}}$}
& {}
& \cellcolor{aegislight}\footnotesize\textbf{$1$--$5$}
& \cellcolor{aegislight}\footnotesize\textbf{$6$--$7$}
& \cellcolor{aegislight}\footnotesize\textbf{$8$}
& \cellcolor{aegislight}\footnotesize\textbf{$9$}
& \cellcolor{aegislight}\footnotesize\textbf{$\geq 10$} \\
& \cellcolor{aegislight}\footnotesize 842 tasks
& {}
& \cellcolor{aegislight}\footnotesize 186 tasks
& \cellcolor{aegislight}\footnotesize 159 tasks
& \cellcolor{aegislight}\footnotesize 191 tasks
& \cellcolor{aegislight}\footnotesize 198 tasks
& \cellcolor{aegislight}\footnotesize 108 tasks \\
\midrule
\rowcolor{aegiszebra}
Opus 4.8
& \textbf{21.7}
& {}
& \textbf{25.8} & \underline{18.2} & \textbf{24.6} & \underline{16.7} & \textbf{24.1} \\
GLM-5.2
& \underline{20.0}
& {}
& \underline{21.0} & \textbf{18.9} & \underline{22.0} & \textbf{18.2} & \underline{19.4} \\
\rowcolor{aegiszebra}
Sonnet 4.6
& 15.0
& {}
& 17.7 & 8.8 & 17.8 & 15.2 & 13.9 \\
Hy3
& 14.0
& {}
& 18.3 & 13.2 & 16.2 & 11.1 & 9.3 \\
\rowcolor{aegiszebra}
Gemini 3.6 Flash
& 8.9
& {}
& 14.5 & 6.3 & 9.9 & 7.1 & 4.6 \\
Kimi K2.6
& 11.2
& {}
& 14.0 & 5.7 & 16.2 & 9.1 & 9.3 \\
\rowcolor{aegiszebra}
Qwen3.5-397B
& 5.3
& {}
& 9.7 & 1.3 & 6.8 & 4.5 & 2.8 \\
Haiku 4.5
& 6.8
& {}
& 15.1 & 4.4 & 6.8 & 3.0 & 2.8 \\
\midrule
Kendall $\tau$
& --
& {}
& 0.71 & 0.79 & 0.93 & 0.86 & 0.93 \\
\bottomrule
\end{tabularx}
\label{tab:length_bin_comp_ns}
\end{table*}

\begin{table*}[h!]
\centering
\small
\setlength{\tabcolsep}{3.5pt}
\renewcommand{\arraystretch}{1.2}
\caption{
\textbf{Strict $S_{\text{proc}}$ w/ skill within checklist-length bins.}
Strict scores (\%) of Process ($S_{\text{proc}}$), evaluated with skill, are reported over all tasks and over the tasks whose process rubric set has $n$ criteria; the number of tasks is given under each column.
Kendall $\tau$ is the rank correlation between the model ranking in each bin and the ranking over all tasks.
\textbf{Bold} indicates the highest score and \underline{underlining} indicates the second-highest score in each column.
}
\begin{tabularx}{\textwidth}{@{}l >{\centering\arraybackslash}X p{6pt} >{\centering\arraybackslash}X >{\centering\arraybackslash}X >{\centering\arraybackslash}X >{\centering\arraybackslash}X >{\centering\arraybackslash}X@{}}
\toprule
& \cellcolor{aegisgreen}\textcolor{white}{\textbf{All}}
& {}
& \multicolumn{5}{c}{\cellcolor{aegisgreen!80!black}\textcolor{white}{\textbf{Checklist length $n$}}} \\
\cmidrule(lr){2-2} \cmidrule(lr){4-8}
\textbf{Model}
& \cellcolor{aegislight}\footnotesize\textbf{$S_{\text{proc}}$}
& {}
& \cellcolor{aegislight}\footnotesize\textbf{$1$--$5$}
& \cellcolor{aegislight}\footnotesize\textbf{$6$--$7$}
& \cellcolor{aegislight}\footnotesize\textbf{$8$}
& \cellcolor{aegislight}\footnotesize\textbf{$9$}
& \cellcolor{aegislight}\footnotesize\textbf{$\geq 10$} \\
& \cellcolor{aegislight}\footnotesize 842 tasks
& {}
& \cellcolor{aegislight}\footnotesize 64 tasks
& \cellcolor{aegislight}\footnotesize 155 tasks
& \cellcolor{aegislight}\footnotesize 304 tasks
& \cellcolor{aegislight}\footnotesize 230 tasks
& \cellcolor{aegislight}\footnotesize 89 tasks \\
\midrule
\rowcolor{aegiszebra}
Opus 4.8
& \underline{27.7}
& {}
& \underline{34.4} & \textbf{35.5} & \underline{25.0} & \underline{23.9} & \underline{28.1} \\
GLM-5.2
& \textbf{28.7}
& {}
& 29.7 & \underline{27.1} & \textbf{30.3} & \textbf{25.2} & \textbf{34.8} \\
\rowcolor{aegiszebra}
Sonnet 4.6
& 10.2
& {}
& 25.0 & 11.6 & 8.6 & 8.7 & 6.7 \\
Hy3
& 21.6
& {}
& \textbf{40.6} & 23.9 & 19.4 & 20.4 & 14.6 \\
\rowcolor{aegiszebra}
Gemini 3.6 Flash
& 5.1
& {}
& 14.1 & 7.1 & 3.9 & 3.0 & 4.5 \\
Kimi K2.6
& 6.1
& {}
& 15.6 & 5.8 & 7.6 & 3.0 & 2.2 \\
\rowcolor{aegiszebra}
Qwen3.5-397B
& 2.6
& {}
& 12.5 & 1.9 & 2.6 & 0.9 & 1.1 \\
Haiku 4.5
& 2.4
& {}
& 20.3 & 0.6 & 1.6 & 0.0 & 1.1 \\
\midrule
Kendall $\tau$
& --
& {}
& 0.57 & 0.86 & 1.00 & 0.93 & 0.93 \\
\bottomrule
\end{tabularx}
\label{tab:length_bin_proc_ws}
\end{table*}

%% file: sections/appendix_sections/domain_scores.tex
\section{Scores by Domain}
\label{app:domain_scores}

\begin{table*}[h!]
\centering
\small
\setlength{\tabcolsep}{5pt}
\renewcommand{\arraystretch}{1.2}
\caption{
\textbf{Strict completion scores and skill gains by domain.}
Each column reports strict $S_{\mathrm{comp}}$ with skill (\%) over the tasks of one domain in Table~\ref{tab:skill_domains}; the number of tasks is given under each domain name.
Parentheses give the skill gain $\Delta = S_{\mathrm{comp}}^{\mathrm{WS}} - S_{\mathrm{comp}}^{\mathrm{NS}}$.
\textbf{Bold} marks the highest score and \underline{underlining} the second-highest score in each column.
}
\label{tab:domain_strict_scores}
\resizebox{\textwidth}{!}{%
\begin{tabular}{@{}l cccccc@{}}
\toprule
\textbf{Model}
& \cellcolor{aegislight}\footnotesize\textbf{Technology}
& \cellcolor{aegislight}\footnotesize\textbf{Business}
& \cellcolor{aegislight}\footnotesize\textbf{Creative}
& \cellcolor{aegislight}\footnotesize\textbf{Harness Behavior}
& \cellcolor{aegislight}\footnotesize\textbf{Knowledge}
& \cellcolor{aegislight}\footnotesize\textbf{All} \\
& \cellcolor{aegislight}\scriptsize 257
& \cellcolor{aegislight}\scriptsize 211
& \cellcolor{aegislight}\scriptsize 136
& \cellcolor{aegislight}\scriptsize 125
& \cellcolor{aegislight}\scriptsize 113
& \cellcolor{aegislight}\scriptsize 842 \\
\midrule
\rowcolor{aegiszebra}
Opus 4.8         & \textbf{24.9} (+10.1) & \underline{51.2} (+23.2) & \textbf{44.1} (+19.1) & \textbf{32.8} (+20.8) & \textbf{62.8} (+30.1) & \textbf{40.9} (+19.2) \\
GLM-5.2          & \underline{19.5} (+4.3) & \textbf{55.0} (+29.4) & \underline{42.6} (+21.3) & \underline{25.6} (+12.0) & \underline{47.8} (+22.1) & \underline{36.8} (+16.8) \\
\rowcolor{aegiszebra}
Sonnet 4.6       & 14.0 (+2.7) & 45.0 (+24.6) & 30.1 (+14.7) & 16.0 (+11.2) & 42.5 (+18.6) & 28.5 (+13.5) \\
Hy3              & 14.4 (+1.6) & 35.5 (+23.2) & 30.9 (+15.5) & 20.0 (+8.0)  & 45.1 (+24.7) & 27.3 (+13.3) \\
\rowcolor{aegiszebra}
Gemini 3.6 Flash & 10.1 (+2.3) & 22.3 (+13.8) & 16.9 (+8.1)  & 18.4 (+16.0) & 31.9 (+12.4) & 18.4 (+9.5)  \\
Kimi K2.6        & 12.1 (+3.2) & 28.9 (+14.7) & 21.3 (+10.3) & 10.4 (+4.8)  & 31.0 (+14.2) & 20.1 (+8.9)  \\
\rowcolor{aegiszebra}
Qwen3.5-397B     & 7.0 (+0.4)  & 21.3 (+16.6) & 11.8 (+5.2)  & 9.6 (+7.2)   & 18.6 (+13.3) & 13.3 (+8.0)  \\
Haiku 4.5        & 9.7 (+4.3)  & 15.2 (+6.2)  & 15.4 (+8.8)  & 8.0 (+5.6)   & 15.9 (+5.3)  & 12.6 (+5.8)  \\
\bottomrule
\end{tabular}%
}
\end{table*}

\begin{table*}[h!]
\centering
\small
\setlength{\tabcolsep}{5pt}
\renewcommand{\arraystretch}{1.2}
\caption{
\textbf{Loose completion scores and skill gains by domain.}
Each column reports loose $S_{\mathrm{comp}}$ with skill (\%) over the tasks of one domain in Table~\ref{tab:skill_domains}; the number of tasks is given under each domain name.
Parentheses give the skill gain $\Delta = S_{\mathrm{comp}}^{\mathrm{WS}} - S_{\mathrm{comp}}^{\mathrm{NS}}$.
\textbf{Bold} marks the highest score and \underline{underlining} the second-highest score in each column.
}
\label{tab:domain_loose_scores}
\resizebox{\textwidth}{!}{%
\begin{tabular}{@{}l cccccc@{}}
\toprule
\textbf{Model}
& \cellcolor{aegislight}\footnotesize\textbf{Technology}
& \cellcolor{aegislight}\footnotesize\textbf{Business}
& \cellcolor{aegislight}\footnotesize\textbf{Creative}
& \cellcolor{aegislight}\footnotesize\textbf{Harness Behavior}
& \cellcolor{aegislight}\footnotesize\textbf{Knowledge}
& \cellcolor{aegislight}\footnotesize\textbf{All} \\
& \cellcolor{aegislight}\scriptsize 257
& \cellcolor{aegislight}\scriptsize 211
& \cellcolor{aegislight}\scriptsize 136
& \cellcolor{aegislight}\scriptsize 125
& \cellcolor{aegislight}\scriptsize 113
& \cellcolor{aegislight}\scriptsize 842 \\
\midrule
\rowcolor{aegiszebra}
Opus 4.8         & \textbf{77.4} (+1.8) & \textbf{92.7} (+9.3) & \textbf{86.3} (+5.7) & \textbf{83.4} (+8.4) & \textbf{93.7} (+7.2) & \textbf{85.7} (+5.9) \\
GLM-5.2          & \underline{76.4} (+4.9) & \underline{91.7} (+10.2) & \underline{84.9} (+6.9) & \underline{81.6} (+8.3) & 89.5 (+6.5) & \underline{84.1} (+7.2) \\
\rowcolor{aegiszebra}
Sonnet 4.6       & 72.5 (+1.1) & 89.1 (+8.9)  & 82.3 (+7.4) & 78.4 (+9.7)  & \underline{90.5} (+7.5) & 81.5 (+6.2) \\
Hy3              & 73.0 (+4.0) & 87.8 (+11.9) & 81.1 (+7.6) & 79.1 (+4.8)  & 87.7 (+6.8) & 80.9 (+7.0) \\
\rowcolor{aegiszebra}
Gemini 3.6 Flash & 73.5 (+5.5) & 82.0 (+8.0)  & 79.8 (+6.8) & 79.4 (+12.1) & 86.0 (+6.3) & 79.2 (+7.4) \\
Kimi K2.6        & 70.9 (+3.3) & 85.7 (+10.7) & 77.7 (+5.8) & 73.7 (+4.5)  & 85.2 (+4.3) & 78.1 (+5.9) \\
\rowcolor{aegiszebra}
Qwen3.5-397B     & 63.4 (+2.6) & 80.2 (+9.2)  & 71.6 (+6.2) & 68.2 (+2.9)  & 79.0 (+4.6) & 71.7 (+5.1) \\
Haiku 4.5        & 61.2 (+3.0) & 79.0 (+8.7)  & 70.3 (+6.8) & 64.8 (+5.9)  & 79.0 (+5.5) & 70.0 (+5.7) \\
\bottomrule
\end{tabular}%
}
\end{table*}

\begin{table*}[h!]
\centering
\small
\setlength{\tabcolsep}{5pt}
\renewcommand{\arraystretch}{1.2}
\caption{
\textbf{Process scores by domain.}
Each column reports loose or strict $S_{\mathrm{proc}}$ with skill (\%) over the tasks of one domain in Table~\ref{tab:skill_domains}; the number of tasks is given under each domain name.
\textbf{Bold} marks the highest score and \underline{underlining} the second-highest score in each column.
}
\label{tab:domain_process_scores}
\resizebox{\textwidth}{!}{%
\begin{tabular}{@{}l cccccccccccc@{}}
\toprule
\textbf{Model}
& \multicolumn{2}{c}{\cellcolor{aegislight}\footnotesize\textbf{Technology}}
& \multicolumn{2}{c}{\cellcolor{aegislight}\footnotesize\textbf{Business}}
& \multicolumn{2}{c}{\cellcolor{aegislight}\footnotesize\textbf{Creative}}
& \multicolumn{2}{c}{\cellcolor{aegislight}\footnotesize\textbf{Harness Behavior}}
& \multicolumn{2}{c}{\cellcolor{aegislight}\footnotesize\textbf{Knowledge}}
& \multicolumn{2}{c}{\cellcolor{aegislight}\footnotesize\textbf{All}} \\
& \multicolumn{2}{c}{\cellcolor{aegislight}\scriptsize 257}
& \multicolumn{2}{c}{\cellcolor{aegislight}\scriptsize 211}
& \multicolumn{2}{c}{\cellcolor{aegislight}\scriptsize 136}
& \multicolumn{2}{c}{\cellcolor{aegislight}\scriptsize 125}
& \multicolumn{2}{c}{\cellcolor{aegislight}\scriptsize 113}
& \multicolumn{2}{c}{\cellcolor{aegislight}\scriptsize 842} \\
\cmidrule(lr){2-3} \cmidrule(lr){4-5} \cmidrule(lr){6-7} \cmidrule(lr){8-9} \cmidrule(lr){10-11} \cmidrule(lr){12-13}
& \cellcolor{aegislight}\footnotesize Loose
& \cellcolor{aegislight}\footnotesize Strict
& \cellcolor{aegislight}\footnotesize Loose
& \cellcolor{aegislight}\footnotesize Strict
& \cellcolor{aegislight}\footnotesize Loose
& \cellcolor{aegislight}\footnotesize Strict
& \cellcolor{aegislight}\footnotesize Loose
& \cellcolor{aegislight}\footnotesize Strict
& \cellcolor{aegislight}\footnotesize Loose
& \cellcolor{aegislight}\footnotesize Strict
& \cellcolor{aegislight}\footnotesize Loose
& \cellcolor{aegislight}\footnotesize Strict \\
\midrule
\rowcolor{aegiszebra}
Opus 4.8         & \textbf{84.9} & \textbf{19.5} & \textbf{91.7} & \underline{37.9} & \textbf{86.3} & \textbf{27.9} & \textbf{87.6} & \underline{23.2} & \underline{89.8} & \underline{31.9} & \textbf{87.9} & \underline{27.7} \\
GLM-5.2          & \underline{84.1} & \underline{16.0} & \underline{91.6} & \textbf{39.8} & \underline{84.3} & \underline{27.2} & 83.1 & \textbf{25.6} & \textbf{90.7} & \textbf{42.5} & \underline{86.7} & \textbf{28.7} \\
\rowcolor{aegiszebra}
Sonnet 4.6       & 68.0 & 7.4 & 83.0 & 14.2 & 76.9 & 9.6 & 76.7 & 9.6 & 79.6 & 10.6 & 76.0 & 10.2 \\
Hy3              & 80.0 & \underline{16.0} & 90.1 & 30.3 & 82.7 & 17.6 & \underline{84.2} & 20.0 & 84.7 & 24.8 & 84.2 & 21.6 \\
\rowcolor{aegiszebra}
Gemini 3.6 Flash & 66.8 & 3.5 & 74.8 & 5.2 & 70.5 & 2.9 & 70.8 & 11.2 & 73.7 & 4.4 & 70.9 & 5.1 \\
Kimi K2.6        & 68.2 & 4.7 & 78.8 & 9.0 & 70.4 & 2.9 & 74.5 & 7.2 & 74.6 & 6.2 & 73.0 & 6.1 \\
\rowcolor{aegiszebra}
Qwen3.5-397B     & 58.5 & 1.6 & 69.0 & 3.8 & 60.4 & 1.5 & 66.4 & 4.0 & 68.5 & 2.7 & 63.9 & 2.6 \\
Haiku 4.5        & 57.6 & 2.7 & 66.6 & 2.4 & 63.4 & 3.7 & 65.1 & 1.6 & 66.2 & 0.9 & 63.1 & 2.4 \\
\bottomrule
\end{tabular}%
}
\end{table*}

%% file: sections/appendix_sections/example_task_comparison.tex

\section{Example Task and the Closest Tasks in Other Benchmarks}
\label{app:example_task_comparison}

We first introduce one of our tasks on commodity supply risk. We then compare it with the closest task that we could find in an adjacent domain in each of Claw-Eval, Agentic Skills at Scale (ASS), and SkillsBench.

\subsection{Our Example Task}
\label{app:example_task_ours}

Figure~\ref{fig:example_task_nickel} shows our task \texttt{nickel-\allowbreak concentration-\allowbreak risk-\allowbreak analyzer}. The instruction follows the workflow of a commodities analyst: the agent computes market concentration indices like HHI, simulates multi-layer policy shocks, and verifies an industry supply claim. The instruction also requires the agent to cross-check sources whose reporting periods do not match and to flag the inconsistencies with explicit warnings. The notes beside the instruction explain why the task is realistic. The figure continues on the next page with the rubrics of the task: completion rubrics check the final workspace, and process rubrics check the agent's trajectory.

\begin{figure}[p]
\centering
\includegraphics[width=0.84\textwidth]{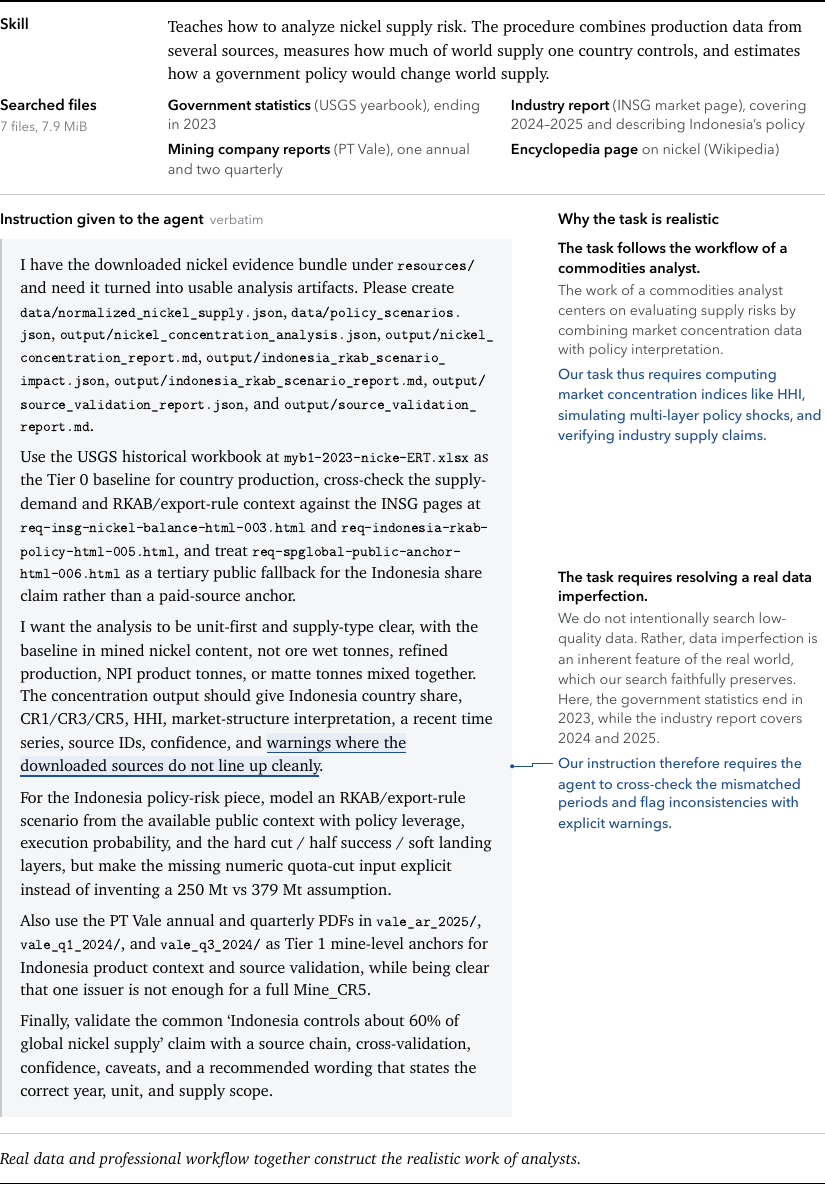}
\caption{
\textbf{Our example task \texttt{nickel-\allowbreak concentration-\allowbreak risk-\allowbreak analyzer}.}
Top: the skill that seeds the task and the searched files in the workspace.
Left: the full instruction given to the agent, with its text unchanged.
Right: why the task is realistic. The first note explains how the task follows the workflow of a commodities analyst. The second note explains the highlighted requirement.
The rubrics of the task continue on the next page.
}
\label{fig:example_task_nickel}
\end{figure}

\begin{figure}[p]
\ContinuedFloat
\centering
\includegraphics[width=0.84\textwidth]{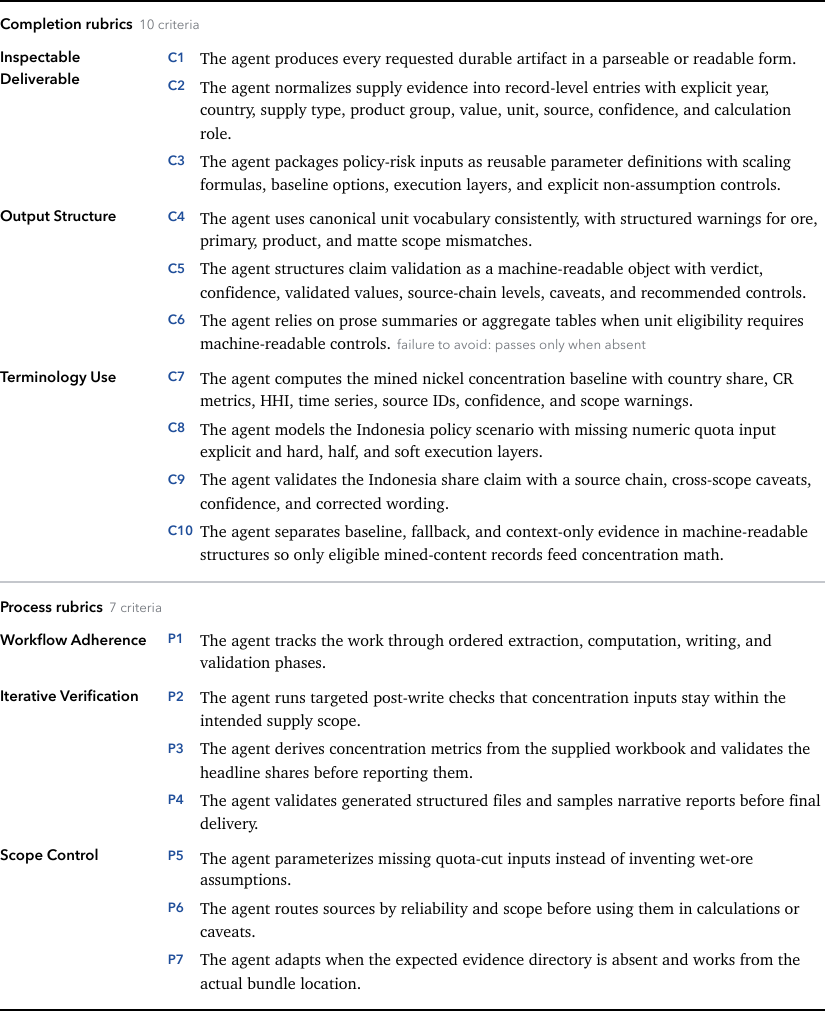}
\caption[]{
(Continued) \textbf{Rubrics of our example task \texttt{nickel-\allowbreak concentration-\allowbreak risk-\allowbreak analyzer}.}
Each criterion is shown with its text unchanged, grouped by the categories in Appendix~\ref{sec:appendix-six-categories}.
C6 describes a failure, so it passes only when the failure is absent.
}
\end{figure}

\clearpage
\subsection{Closest Tasks in Other Benchmarks}
\label{app:example_task_others}

For each other benchmark, we summarize the closest task. Table~\ref{tab:example_task_comparison} compares the four tasks.

\paragraph{Claw-Eval.}
The task \path{T067zh_synopsys_china_revenue} is in Chinese only. From two years of Synopsys 10-Ks, the agent measures the change in China revenue share and confirms or refutes the claim that this share is in structural decline under export controls.
\emph{Environment}: no files. The agent finds the 10-Ks with live web search.
Task page: \href{https://huggingface.co/datasets/claw-eval/Claw-Eval/blob/main/data/general-00000-of-00001.parquet?sql_row=63}{Hugging Face}.

\paragraph{ASS.}
In the task \texttt{tessl-\allowbreak multi-\allowbreak anthropics-\allowbreak financial-\allowbreak services-\allowbreak plugins\_3}, the agent reviews a bond-and-option portfolio: bond pricing and rate-shock sensitivity, a macro and rates snapshot, and option volatility and greeks, then an overall risk view.
\emph{Environment}: one 1\,KB holdings CSV. The agent gets market data from the web.
Task page: \href{https://huggingface.co/datasets/tesslio/task-evals-for-skills/viewer/default/train?row=856}{Hugging Face}. The prompt is in the \path{task_description} column. The full task, with input files and rubrics, is under \path{tasks/<task-name>/} in \texttt{tasks.zip} in the same repository.

\paragraph{SkillsBench.}
The task \path{reserves-at-risk-calc} has medium difficulty and is graded by a test script. In Excel only, the agent computes gold price volatility from IMF data, then countries' gold exposure and reserves-at-risk. The prompt names the rows to fill at each step.
\emph{Environment}: one Excel template (about 105\,KB).
Task page: \href{https://huggingface.co/datasets/benchflow/skillsbench/blob/main/reserves-at-risk-calc/task.md}{Hugging Face}.

\begin{table*}[h!]
\centering
\small
\renewcommand{\arraystretch}{1.25}
\caption{
\textbf{Our example task and the closest task in each other benchmark.}
Each column describes one task from Appendix~\ref{app:example_task_comparison}.
}
\label{tab:example_task_comparison}
\begin{tabularx}{\textwidth}{@{}>{\raggedright\arraybackslash}p{2.2cm} *{4}{>{\raggedright\arraybackslash}X}@{}}
\toprule
\rowcolor{aegislight}
& \textbf{Ours (nickel)} & \textbf{Claw-Eval T067zh} & \textbf{ASS portfolio} & \textbf{SkillsBench RaR} \\
\midrule
\textbf{Evidence}
& 4 kinds of official and company sources to rank and to align in units and scope
& One source (two 10-Ks)
& One holdings CSV, plus web data
& One IMF database, plus the template's sheets \\
\rowcolor{aegiszebra}
\textbf{Deliverables}
& 8 files (data, scenarios, reports, source validation)
& One analysis
& One combined review
& A filled-in workbook \\
\textbf{Prompt style}
& A user request in one paragraph that mixes what and how
& Numbered analysis steps
& Three sections, each listing the metrics to compute
& Each step names the rows to fill \\
\rowcolor{aegiszebra}
\textbf{Grading}
& LLM judge
& 70\% LLM judge, 30\% number of searches
& LLM judge
& Test script with exact answers \\
\textbf{Closest point to our task}
& ---
& Fixing the reporting basis, then confirming or refuting a market claim
& Combined macro and risk judgment
& Measuring commodity price risk \\
\bottomrule
\end{tabularx}
\end{table*}

%% file: sections/appendix_sections/prompts/appendix_prompts.tex
\begingroup
\renewcommand{\ttdefault}{lmtt}
\raggedbottom

\section{Prompt Templates for Benchmark Construction}
\label{app:synthesis-prompts}

\SFNeedspace{12\baselineskip}

\subsection{Resource preparation}

\SFNeedspace{7\baselineskip}

\subsubsection{Resource requirements planning}

\label{app:prompt-DS01}


\begin{lstlisting}[style=sfprompt]
# Requirements Planner Agent Prompt v0.3

You are the Codex Requirements Planner Agent in a data synthesis pipeline.

Your job is to read the complete supplied source skill, which is composed of the `SKILL.md` and its companions (`reference/`, `scripts/`, `agents/`, etc.). Finally, produce `resource_requirements.v0.3` JSON, which describes the concrete input resources that the skill needs in order to perform its tasks.

Hard rules:

- **Output JSON only**, matching `resource_requirements.schema.json`.
- Do not download external resource bytes. Do not run package managers, `git clone`, `curl`, `wget`, `requests`, or browser-save operations.
- Use only resource types listed in the `## Resource type compatibility (authoritative)` table later in this prompt. That table is generated from the same source the deterministic validator uses; do not invent types or subtypes.

Use the accompanying contract, schema, taxonomy, and complete skill context as the source of truth.
\end{lstlisting}

\SFNeedspace{7\baselineskip}

\subsubsection{Requirements extraction contract}

\label{app:prompt-DS10}


\begin{lstlisting}[style=sfprompt]
# Requirements Extraction Contract v0.3

## Core Task

In this pipeline, we need to extract from the complete source skill the **actual input resources** that the skill would require when performing tasks that make use of it.

**The skill itself implicitly describes what it operates on**. For example, a code review skill needs changed code, diffs, source context, tests, and review guidelines; a document summarization skill needs complete documents; an SQL generation skill needs database schemas, example data, and query/task context.

Your core task is: **read the complete source skill (`SKILL.md` and its companions), understand what the skill is supposed to do, then deduce the input files/resources that must be provided for the skill to work properly, and output these requirements in a structured format**.

You **must avoid** extracting resources that merely look relevant by format but are unused by the skill. Do not request an irrelevant demo SQLite database, a generic DOCX template, a random Markdown syntax guide, or a package lockfile unless the skill actually operates on that object. What you need are the **object resources that can be directly manipulated by the skill to accomplish its described task**.

The output resource requirements will be passed to a _Resource Research Agent_, which will search the web for real, public, stable resources. Later deterministic stages will validate, download, and safely extract only resources listed by that research agent.

## Required behavior

- Keep the requirements compact, normally 4-10 items.
- Extract intent, domain, primary operations, expected inputs/outputs, forbidden scope, and domain-specific terminology from the **full source skill**, not only name or description.
- Determine resource types based on skill operational semantics, not fixed format coverage. Avoid format-only requirements. If a file format is requested, explain why that format is naturally operated on by the skill.
- `resource_family` is fully determined by `resource_type`: set each requirement's `resource_family` to exactly the Family shown for its type in that table, never choose it independently. Likewise pick `acquisition` from that type's listed acquisition modes, and (for Extractable=yes types) set `archive_constraints.extract` to true.
- Set `planner_assertions.read_full_skill_md`, `planner_assertions.did_not_download`, and `planner_assertions.used_schema` to `true`.

### Required Specifics Preferences

- Include linkage hints when multiple resources are semantically coupled, such as source + manifest + config + tests from one repository snapshot.
- Include search queries that help the next agent find stable public resources.
- Include enough constraints for the next agent and deterministic downloader: HTTPS, no auth, expected MIME/extensions, acquisition modes, archive extraction expectations, GitHub pinning policy, and linkage hints.
- Prefer requirements that can be satisfied by public, stable, no-auth resources.

### Rules For Specific Tpyes

- For HTML/documentation, request complete pages or page-set bundles if one page is incomplete.
- For GitHub-related selected resources, require full 40-character commit pinning.
- For code-review or diff/PR-based skills, remember a review spans two commit states: a base tree plus a proposed change on top. When the requirement set pairs a diff/PR (the change under review) with its repository snapshot, the repository requirement's `pinning_policy`/`linkage_hints` must state that the repository is pinned to the change's **base** commit (pre-change state), not the merged/head state, and should carry the base/head relationship (the research agent records these as `repository.base_sha`/`head_sha` on the diff/PR resource). This does not apply when the diff/PR is only background context for editing the current code.

## Archive rule

Archive requirements must describe expected members and extraction limits. The deterministic downloader performs safe extraction. Agents never download or extract resource bytes.
\end{lstlisting}

\SFNeedspace{7\baselineskip}

\subsubsection{Resource taxonomy}

\label{app:prompt-DS11}


\begin{lstlisting}[style=sfprompt]
# Resource Taxonomy v0.3

The v0.3 taxonomy describes input resources that a skill may genuinely operate on. The Requirements Planner Agent chooses from these types after reading the complete source skill; deterministic code validates that choices stay within this taxonomy.

## Families

- `document`: `pdf`, `text`, `markdown`, `docx`, `pptx`
- `web`: `html`
- `structured_data`: `csv`, `tsv`, `json`, `jsonl`, `yaml`, `toml`, `xml`, `parquet`, `xlsx`
- `code`: `github_repository`, `source_code`, `diff_patch`, `github_issue`, `github_pull_request`
- `manifest`: `package_manifest`, `lockfile`, `openapi_spec`
- `config`: `config`
- `archive`: `archive`, `zip_archive`, `tar_archive`, `tar_gz_archive`, `package_archive`
- `database`: `sqlite`
- `geospatial`: `geojson`, `shapefile_archive`
- `notebook`: `notebook`
- `media`: `image`, `audio`, `video`

## Acquisition modes

- `direct_file` — URL points directly to downloadable bytes.
- `api_endpoint` — URL points to a public no-auth endpoint returning resource bytes or JSON/HTML representation.
- `landing_page` — URL points to a public page intended as the resource itself, usually HTML.
- `github_archive` — URL points to a GitHub archive pinned to a 40-character commit SHA.
- `raw_file` — URL points to a raw source file, preferably pinned when GitHub-hosted.
- `archive_extract` — URL points to an archive that must be safely extracted before benchmark use.

Resource choices must use an acquisition mode compatible with the chosen resource type.

## Validation depth

Deep-ish stdlib validation:

- PDF magic/header checks
- CSV/TSV parsing
- JSON/JSONL parsing
- XML parsing
- TOML parsing when `tomllib` is available
- SQLite header checks
- notebook JSON shape checks
- ZIP/TAR structural checks
- DOCX/PPTX/XLSX container path checks
- source code/text heuristics
- diff marker checks
- package manifest and lockfile structure where feasible

Shallow validation:

- YAML
- Parquet
- image/audio/video magic bytes and MIME/extension checks
- some package archive formats

## GitHub resources

Selected GitHub resources must be pinned for repeatability:

- `github_repository` uses a `github.com/<owner>/<repo>/archive/<40-char-sha>.zip` or equivalent pinned archive URL.
- raw file resources use `raw.githubusercontent.com/<owner>/<repo>/<40-char-sha>/<path>`.
- repository metadata records owner, repo, requested ref when known, and commit SHA.
- repo homepages are allowed as `landing_url`, not as the selected fetch URL for `github_repository`.

## Archive resources

Archive-like resource types are extractable and must declare archive handling:

- `github_repository`
- `archive`
- `zip_archive`
- `tar_archive`
- `tar_gz_archive`
- `package_archive`
- `shapefile_archive`

The deterministic downloader extracts these with path-safe member handling and budget limits. Requirements and manifests should describe expected top-level entries, required member paths or extensions when known, maximum members, maximum extracted bytes, and whether nested archives are allowed.

## Bundles

Some skills need linked files rather than a single URL. The manifest schema supports bundles for cases such as:

- multiple raw files from the same repository commit
- documentation page sets
- dataset releases with separate metadata and data files

Bundles must declare why the files belong together and which members are required.
\end{lstlisting}

\SFNeedspace{7\baselineskip}

\subsubsection{Artifact retrieval}

\label{app:prompt-DS02}


\begin{lstlisting}[style=sfprompt]
# Resource Research Agent Prompt v0.3

You are the Resource Research Agent for a data synthesis pipeline.

## Core Task

A Requirements Planner Agent has analyzed a complete source _skill_, which is composed of the `SKILL.md` and its companions (`reference/`, `scripts/`, `agents/`, etc.). Finally, produced `resource_requirements.v0.3` JSON that lists the input resources the skill needs to operate. You receive the requirements JSON (inlined below) plus the same complete skill, which has been copied into your working directory at `./skill/` (the `SKILL.md` and its companions). **Before you search for anything, read the skill directly from `./skill/`** so your resource choices match the skill's real operational semantics — do not skip it, and never substitute a web-hosted copy of the skill for the local files.

Your role is to turn these abstract requirements into concrete, downloadable URLs pointing to real, public, stable resources. You do **not** download anything yourself. You search the web, evaluate candidates, and output a structured manifest that a deterministic downloader will later use.

The downstream pipeline depends on your choices: if you pick an unstable, private, unavailable, or semantically weak resource, data synthesis will fail or become misleading. Use the complete original skill to avoid format-only substitutions.

### Your Job

- **First, read the complete source skill from `./skill/` in your working directory** — `./skill/SKILL.md` plus every companion file listed in the source-skill section below — before doing any web search. Read it from disk; it is not inlined in this prompt.
- Read the `resource_requirements.v0.3` JSON (inlined below) and cross-check each requirement against what the skill at `./skill/` actually operates on.
- Use web search to identify real, public, stable resources that satisfy each requirement.
- Return only JSON matching `resource_manifest.schema.json`.
- Provide one selected resource and at least one alternative for every requirement. Alternatives must satisfy the same requirement `resource_type`; do not use an HTML landing page as the alternative for a PDF requirement or similar type substitution.
- Decide when a requirement is best represented by a linked bundle of files/pages instead of a single file.

## General Rules

### Hard Rules

- Do not download files or resource bytes.
- Do not run `curl`, `wget`, `requests`, browser-save operations, package managers, `git clone`, or commands whose purpose is fetching resource bytes.
- Do not create local resource files.
- Do not return search result pages as selected resource URLs.
- Do not invent URLs.
- Do not select resources requiring login, cookies, API keys, CAPTCHA, private credentials, or temporary signed URLs.
- The deterministic downloader will download only URLs listed in your manifest. It will not search, crawl, or infer additional URLs.
- Do not set `agent_assertions.read_full_skill_md: true` unless you actually read `./skill/SKILL.md` and its companions from disk first. Reading a web-hosted copy of the skill does not count and must not replace reading the local `./skill/`.

### Preferences

- Prefer small, stable, public resources with clear provenance.
- Prefer a lightweight project that can run on a normal personal laptop. It must not require server-grade GPUs or large amounts of RAM — and, if it is software, easy to set up and run.

## Required manifest behavior

- `schema_version` must be `resource_manifest.v0.3`.
- `generated_by.agent` must be `codex-resource-research-agent`.
- `agent_assertions.did_not_download`, `agent_assertions.search_only`, and `agent_assertions.read_full_skill_md` must all be `true`.
- `source_requirements_sha256` must exactly match the caller-supplied value.
- `source_skill_context_sha256` must exactly match the caller-supplied value.
- Every candidate must include `url`, `landing_url`, `acquisition`, `repository`, `archive_handling`, `bundle`, provenance, and validation notes.
- `acquisition` must be exactly one of: `direct_file`, `api_endpoint`, `landing_page`, `github_archive`, `raw_file`, `archive_extract`. Do not invent values such as `html_page`, `landing_page_set`, or `raw_file_bundle`; represent sets/bundles only via the `bundle` field.
- Use `repository: null` for non-repository resources.
- Use `archive_handling: null` for non-archive resources.
- Use `bundle: null` for a single-file/single-page candidate.

## Resource type, family, and field compatibility

- A `## Resource type compatibility (authoritative)` table is included later in this prompt. It is generated from the same source the deterministic validator uses, so any value it does not list **will be rejected**. Obey it exactly.
- `resource_type` (on every candidate, bundle member, and resource group) MUST be one of the rows in that table. Do **not** invent new types or subtypes. In particular, there is no `github_issue_comments` or `github_review_comments` type — model PR review comments / issue comments as part of a `github_pull_request` or `github_issue` resource (its public API JSON, or as bundle members of that same type).
- `resource_family` is fully determined by `resource_type`: set it to exactly the Family shown for your chosen type in the table. Never choose it independently.
- `expected_mime_type`, `expected_extension`, and `acquisition` MUST each be one of the values listed for that type's row. The table shows `""` when an empty extension is allowed; for types whose allowed extensions are filenames (e.g. `package_manifest` allows `Gemfile`, `package.json`, `.gemspec`), use one of those exact tokens rather than leaving it blank. For Extractable=yes types, set `archive_handling` with `extract: true`.
- If the only realistic resource for a requirement does not fit any row's allowed values (e.g. a spreadsheet only published inside a `.zip`), pick the closest fitting `resource_type` whose values it satisfies, or choose a different resource — do not bend a type's fields to values the table does not list.

## Specific Rules

### GitHub reproducibility rules

- Selected GitHub resources MUST be pinned to a full 40-character commit SHA.
- `github_repository`: use a GitHub archive URL pinned to a commit, e.g. `https://github.com/<owner>/<repo>/archive/<40-char-sha>.zip`; set `landing_url` to a GitHub page for the repo/ref; fill `repository.ref` and `repository.commit_sha` with that same SHA.
- `source_code`, `package_manifest`, `lockfile`, `config`, `markdown`, `openapi_spec`, `notebook`, and other raw GitHub files: use `https://raw.githubusercontent.com/<owner>/<repo>/<40-char-sha>/<path>`, not `main`, `master`, a tag, or a GitHub `blob` page.
- Do not use `/archive/refs/heads/...`, `/archive/refs/tags/...`, `/main`, `/master`, or any branch/tag ref for selected resources. Selected raw GitHub URLs must include the 40-character commit SHA in the URL path and in `repository.ref`/`repository.commit_sha`.
- If you cannot pin a GitHub resource to a commit SHA, do not select it as a candidate resource; choose a different resource or put it only in alternatives with the risk clearly stated.
- If a candidate has multiple GitHub bundle members that are meant to be used together, they must come from the same owner/repo and the same commit SHA.

### Code-review base pinning

- A code review is inherently a **two-commit** thing: a base tree plus a proposed change applied on top. Keep this in mind whenever a requirement provides a `diff_patch` or `github_pull_request` as the **change under review** (something to review, critique, or apply — not merely background context).
- In that case, when you also select that change's repository as a `github_repository` snapshot, pin the repository archive to the change's **base commit** (the pre-change state), **NOT** the PR `merge_commit_sha` or the `head` SHA. If you pin to the merge/head, the change being reviewed is already present in the snapshot, which defeats the review.
  - For a GitHub PR, the base commit is the PR's `base.sha`. Pin the `github_repository` archive URL and `repository.commit_sha` to that base SHA.
  - The `diff_patch` / `github_pull_request` resource then represents the proposed change (base → head) layered on top of that base snapshot.
  - Record the change endpoints on the **diff/PR resource's** `repository` block: set `repository.base_sha` to the PR `base.sha` and `repository.head_sha` to the PR `head.sha` (that resource's own `commit_sha` may stay the merge/landing commit). This lets the deterministic validator confirm the paired `github_repository` is pinned to the base. Do NOT overload `bundle.linkage.commit` for this.
  - `base_sha` and `head_sha` are required fields on every `repository` object. For any resource that is NOT a diff/PR under review (the repository snapshot itself, raw files, configs, etc.), and whenever a value is unknown, set both to `null`.
- Exception: if the diff/PR is included only as **background/thematic context** (e.g. a refactor skill that edits the current code and merely uses the PR/issue to point at a relevant area), the repository may be pinned to any coherent snapshot and base-pinning is not required. State which mode applies in the candidate's `why_it_matches` / `rationale`.

### Linked bundle rules

- Use `bundle: null` when one URL is sufficient.
- Use `bundle.mode: "multi_file"` when the we need coordinated files, e.g. source + manifest + config + tests + docs from one project snapshot.
- Use `bundle.mode: "html_page_set"` when an HTML resource is split across multiple pages and one page alone is incomplete for the requirement.
- Bundle members must be explicit URLs. Do not rely on the downloader to crawl links.
- Keep bundles small and necessary, normally 2-6 members and never more than the schema limit.
- Every bundle needs a clear `rationale` explaining why the files/pages must be used together.
- For `same_github_repo_commit` bundles, fill `bundle.linkage.repository` as `owner/repo` and `bundle.linkage.commit` as the 40-character SHA.
- Every member needs `id`, `role`, `resource_type`, `url`, `acquisition`, `expected_mime_type`, `expected_extension`, `requires_auth: false`, `repository`, `archive_handling`, `required`, and `why_needed`.

### HTML completeness rules

- For HTML requirements, assess whether the selected page contains the complete relevant content.
- Avoid search pages, login pages, thin landing pages, JavaScript-empty shells, and pages that merely link to the actual content.
- If content is split across pages, sections, print pages, canonical pages, or required subpages, return an `html_page_set` bundle listing the needed pages.
- Do not include many loosely related pages. Include only pages needed to make the resource complete for the requirement.

### Resource-specific rules

- `diff_patch`: prefer `.diff` or `.patch` URLs, or GitHub PR `.diff` URLs.
- `github_issue` / `github_pull_request`: prefer public GitHub API JSON or stable HTML pages if API JSON is not practical.
- Archives: include `archive_handling` with expected entries/extensions and conservative extraction limits.
- Office docs (`docx`, `pptx`, `xlsx`): prefer small public templates/examples from authoritative sources only when the requirement semantically calls for them.
- Media: prefer small files; include size risks if unknown.
- Humanities/religion/philosophy/literature/genealogy: prefer canonical texts, translations, academic references, official exam/teaching materials, public-domain sources, library/archive pages, and primary sources. Do not substitute unrelated policy/statistics resources merely because a keyword overlaps.

## Candidate bundle example shape

```json
{
  "url": "https://raw.githubusercontent.com/owner/repo/0123456789abcdef0123456789abcdef01234567/README.md",
  "landing_url": "https://github.com/owner/repo/tree/0123456789abcdef0123456789abcdef01234567",
  "title": "Example linked project files",
  "publisher": "owner/repo",
  "source_type": "github",
  "acquisition": "raw_file",
  "expected_mime_type": "text/plain",
  "expected_extension": ".md",
  "requires_auth": false,
  "license_or_terms": "Repository license applies.",
  "repository": {"host": "github.com", "owner": "owner", "repo": "repo", "ref": "0123456789abcdef0123456789abcdef01234567", "path": "README.md", "commit_sha": "0123456789abcdef0123456789abcdef01234567", "requested_ref": null, "pin_evidence_url": "https://github.com/owner/repo/commit/0123456789abcdef0123456789abcdef01234567"},
  "archive_handling": null,
  "bundle": null,
  "provenance": {"discovered_via_queries": ["..."], "supporting_pages": ["..."], "evidence_snippets": ["..."], "checked_at": "2026-06-10T00:00:00Z"},
  "agent_validation": {"confidence": 0.9, "why_it_matches": "...", "known_risks": [], "expected_size_bytes": null}
}
```
\end{lstlisting}

\SFNeedspace{7\baselineskip}

\subsubsection{Skill intent summarization}

\label{app:prompt-DS03}


\begin{lstlisting}[style=sfprompt]
# Skill Analysis

## Objective

Read the target skill's description and produce a **general, short sentence** — at most **12 words** — that summarizes what the user wants to do by installing the skill for LLM agents. Note that it is not about what the skill does **but about what the user wants to accomplish**.

It must **not** be written in the **user's first-person voice** (for example, _"Help me ..."_) **but** in an imperative sentence (start with a verb). In addition, it should be conclusive enough to cover the skill's overall purpose and functions rather than emphasizing any single feature. **Note that do not include anything like "follow the skill / guidelines".**

Base the summary **only on the skill's description**; do not inspect the skill's implementation, scripts, or bundled files, and do not invent capabilities the description does not state. Write the result to `skill_analysis.json` with the fields `skill_name` and `general_request`.

## Read Context

- `{{SKILL_DIR}}/SKILL.md` — the target skill's description (read the description only).
- `{{PROMPTS_DIR}}` — this prompt directory.

## Owned Writable Paths

- `{{OUTPUT_DIR}}/skill_analysis.json` — the single output artifact.

## Forbidden Paths and Actions

- Do not read or rely on anything beyond the skill's description.
- Do not modify, rename, or delete the skill files or any path outside `{{OUTPUT_DIR}}`.
- Do not exceed 12 words in `general_request`.
- Do not name the skill. Do not quote skill-internal terms inside it. Do not say anything like "follow the skill / guidelines".
- Do not install dependencies, access the network, or run destructive commands.
\end{lstlisting}

\SFNeedspace{7\baselineskip}

\subsubsection{Forum request retrieval}

\label{app:prompt-DS04}


\begin{lstlisting}[style=sfprompt]
# Request Form

Read `skill_analysis.json`, `SKILL.md`, and skill's companions. The contents of `"general_request"` from `skill_analysis.json` is a very general summary of a user's request. Amplify the request with real human's requests.

To achieve this, search on the internet mainly using the contents of `"general_request"` from `skill_analysis.json`. Note that the contents of `"general_request"` from `skill_analysis.json` **is always much more important** than the skill itself here. Do not use the skill to search. The skill just provides some contexts if needed.

The sources must be the following 20 popular forums / discussion communities:
1. **Reddit** — massive topic-based communities called subreddits
2. **Quora** — Q&A and opinion discussions
3. **Stack Exchange** — expert Q&A network; includes Stack Overflow
4. **Stack Overflow** — programming Q&A
5. **Tripadvisor Forums** — travel discussions
6. **XDA Developers** — Android, mobile, ROMs, device modding
7. **Hacker News** — tech/startup/programming discussions
8. **Digital Point** — SEO, web marketing, webmaster forum
9. **Warrior Forum** — internet marketing and online business
10. **MacRumors Forums** — Apple products and rumors
11. **Mumsnet Talk** — parenting, family life, relationships, and lifestyle discussions.
12. **Linus Tech Tips Forum** — PC building, hardware, tech
13. **Ubuntu Forums** — Ubuntu/Linux support
14. **BleepingComputer Forums** — malware removal and computer help
15. **AVS Forum** — home theater, TVs, audio/video gear
16. **Bodybuilding.com Forums** — fitness and bodybuilding
17. **The Student Room** — student life, universities, exams
18. **Wrong Planet** — autism, neurodiversity, life advice
19. **ResetEra** — gaming, entertainment, culture
20. **Mumsnet Talk** — parenting and lifestyle discussions

Specifically, spawn 20 sub-agents, each of which is responsible for one forum and returns one user request candidate. Then, write one user request based on these 20 candidates. This user request can be composed by selecting, merging, and polishing any of these candidates.

Make sure every bit of information comes from these candidates. LLM's internal knowledge and skill information are **forbidden** to be written into the user request. Skill knowledge must **not** be incorporated into the user request.

Also, write down the url of 20 candidates. Specify which of these 20 candidates are used to construct the final user request.

Finally, quickly check if the skill can be applied to help solve this user request. Again, the contents of `"general_request"` from `skill_analysis.json` **is always much more important** than the skill itself here. Do not include the skill into the composition of the user request.

## Read Context

- `{{SKILL_DIR}}/` — the target skill and its companions.
- `{{SKILL_ANALYSIS_OUTPUT}}/skill_analysis.json` — Stage 0's outcome; its `"general_request"` field is the general summary to amplify.

## Owned Writable Paths

- `{{OUTPUT_DIR}}/request_form.json` — the single output artifact.

```json
{
  "general_request": "string — copied verbatim from skill_analysis.json",
  "user_request": "string — the single final user request composed from the used candidates",
  "used_candidate_ids": [1, 3, 4, 7],
  "skill_applicability": {
    "can_apply": true,
    "reason": "string — one sentence on whether the skill can help with user_request"
  }
}
```

- `{{OUTPUT_DIR}}/source_url.json` — the list of 20 source url.

```json
{
  "candidates": [
    {
      "id": 1,
      "forum": "Reddit",
      "source_url": "https://www.reddit.com/...",
      "candidate_request": "string — the request candidate distilled from this forum",
      "used_in_final_request": true
    },
    {
      "id": 2,
      "forum": "Quora",
      "source_url": null,
      "candidate_request": "string — or 'No qualifying candidate found.'",
      "used_in_final_request": false
    }
  ]
}
```

## Forbidden Paths and Actions

- Do not read or rely on anything, except files from Read Context section.
- Do not modify, rename, or delete the skill files or any path outside `{{OUTPUT_DIR}}`.
- Do not name the skill inside the request. Do not quote skill-internal terms inside it. Do not say anything like "follow the skill / guidelines".
- Do not install dependencies or run destructive commands.
\end{lstlisting}

\SFNeedspace{12\baselineskip}

\subsection{Skill-driven task synthesis}

\SFNeedspace{7\baselineskip}

\subsubsection{Artifact roles and task scenarios}

\label{app:prompt-DS05}


\begin{lstlisting}[style=sfprompt]
# Prompt template: produce a `files_to_change` change-surface analysis (hybrid)

> How to use: send the whole **PROMPT** section below to an agent, after filling in three placeholders:
> - `{SKILL}` — the skill name (e.g. `37signals-rails`)
> - `{RUN_DIR}` — absolute path to that skill's run directory (e.g.
>   `/Users/.../env_search_v0.3/runs/six-skill-e2e/{SKILL}`)
> - `{ENV_SEARCH_ROOT}` — absolute path to the `env_search_v0.3` root (holds `scripts/` and `schemas/`)
>
> The output must conform to the schema at:
> `/Users/.../env_search_v0.3/schemas/search_env/file_change_summary/files_to_change.schema.json`
>
> This is a **hybrid** workflow. A deterministic script computes the *factual* layer
> (provenance, caveats, duplicate/orphan/failure detection, codebase identity). You
> only do the *judgment* layer (which code to change, how, tied to the skill's rules).

---

## PROMPT

You're analyzing a task environment that was synthesized for a particular skill. Your job is to produce a **`files_to_change` change-surface summary** as JSON: a record of **which files the final task should have the agent edit, and how**, along with every data-quality problem the environment carries. This is an **intermediate step** in data synthesis — not the final task itself.

### How the synthesis pipeline works
1. Start from a skill, then plan a set of **resource requirements** (what files to go find).
2. Search the web → a **manifest**, where each requirement has one `selected` source plus a few `alternatives`.
3. Download → a **download_report** recording what actually landed: `selected` vs some `alternative`, success/failure, per-member results for bundles.
4. The files end up under `resources/` — that directory *is* the agent's working tree.
5. The final task = this environment + a user request, engineered so the **skill is genuinely needed**.

⚠️ **Ignore the user request for now** — it gets composed later. You may glance at the request candidates (`request_form.json` / `source_url.json`) as a *directional hint*, but don't bake them in. Your call is driven by **what's in the environment + what the skill is built to do**.

### Step 0 — Start from the deterministic prepass (already run for you)
A deterministic prepass has **already been run** and written to
`{RUN_DIR}/file_change_summary/files_to_change.prefill.json`. (To refresh it yourself:
`python3 {ENV_SEARCH_ROOT}/scripts/search_env/file_change_summary/file_change_prepass.py --run-dir {RUN_DIR}`.)
It has **already reconciled** `manifest ↔ download_report ↔ resources/` for you and produced the **factual layer**:
- `codebase` (repo, commit, `extracted_root`, plus raw `repo_facts` — file counts, `test/` vs `spec/` counts, fixtures present, etc.)
- `resource_roles[]` with `provenance` (selected vs alternative-N, download status, source) and rule-based `caveats`
- `context_inputs_read_only` and a top-level `provenance_and_caveats` summary
- `machine_hints` — `scenario_seeds.review` (files touched by patches/PRs), `patch_vs_snapshot` (SHA mismatches), and structured `duplicates` / `orphans` / `failures` / `misleading_names`
- provisional `task_role`s (some may be `"undetermined"`)

**Trust this output. Do NOT re-derive provenance/caveats by eyeballing the JSON.** The script is exact; you are not. Carry `resource_roles`, `provenance`, `caveats`, `context_inputs_read_only`, and `provenance_and_caveats` through into your final output **verbatim** (you may append, but don't silently drop or contradict them).

### Step 1 — Read the skill's rule system
- `{RUN_DIR}/skill_context.v0.3.json` embeds the full `SKILL.md`. **Internalize the rule system** and note the rule IDs (e.g. `ctrl-crud-only`, `model-concerns`). Every change you propose must cite the specific rules driving it.
- `{RUN_DIR}/resource_requirements.v0.3.json` — for each requirement read `purpose`, `detailed_description`, `expected_content`, `disallowed_content`, `priority`. This tells you what each resource is *for* — which is exactly what you need to resolve the `"undetermined"` roles.

### Step 2 — Read the real content to ground your conclusions (no guessing)
Use `codebase.extracted_root` + `machine_hints` as your map, then open the actual files:
- Repo shape: controllers/models/concerns, the test setup (Minitest or RSpec? fixtures?), migrations, routes, schema, dependency manifest. `repo_facts` already hints at this — confirm by reading.
- Review artifacts (the `scenario_seeds.review` files; the patch / PR json / issue): what they touch and what they're about. Heed `patch_vs_snapshot` — if a patch is already folded into the snapshot, it's review context, not something you re-apply.
> When there are many files, dispatch a read-only explorer subagent to surface relevant paths and keep only the conclusions on the main thread.

### Step 3 — Resolve every `undetermined` task_role
For each resource the script left `"undetermined"`, decide its role from the requirement's `purpose` + the skill's intent: is it a thing the agent **edits** (e.g. a source file in the working tree), or **reads as reference/context** (e.g. a sample doc, a competitor page, a spec the agent writes *against*)? A code-refactoring skill edits the repo; a document-writing skill mostly reads references and **creates a new output file**. Set a concrete `task_role` and drop the `task_role_provisional` flag.

### Step 4 — Build `files_to_change` (the core judgment)
For every candidate change write a concrete `how` grounded in the **actual file content** and attach `skill_rules` (rule IDs) + a `confidence`. Target the spots that land squarely on what the skill changes — e.g. custom member actions → CRUD resources; service objects → rich models; RSpec → Minitest/fixtures; scattered logic → a concern; for a writing skill, the new document to author and the references that constrain it.

Since the request isn't fixed, split the surface into a few **self-consistent scenarios**, each backed by a distinct cluster of resources (e.g. one feature thread stitched from issue + PR + patch; one refactor thread pointed to by the requirements). List each scenario's `targets`. Every `path` is relative to `codebase.extracted_root`. **Edits belong on the repo copy** — never on a standalone resource the script flagged as a duplicate; never on an orphan-mismatch resource; never on a generated artifact (`db/schema.rb`, lockfiles) — those are context only.

### Step 5 — Write the remaining judgment fields
- `codebase.architecture_notes` (what's architecturally notable *for this skill*) and a top-level `key_finding`.
- `recommended_change_set`: a focused subset at the intersection of **grounded in the environment × where the skill has real leverage × verifiable**, with `excluded_with_reason`.
- If the most self-consistent thread in the environment points in a *different direction* than the skill's core strength (or than the request candidates), say so and leave the scenario choice to the final request.

### Output
- Write to: `{RUN_DIR}/file_change_summary/files_to_change.v0.3.json` (merge your judgment layer into the prefill's factual layer).
- **Follow the schema strictly**: `{ENV_SEARCH_ROOT}/schemas/search_env/file_change_summary/files_to_change.schema.json`
  (top-level required: `schema_version`/`skill`/`codebase`/`resource_roles`/`files_to_change`; `change_type` and `confidence` use the controlled enums; every `path` is relative to `codebase.extracted_root`).
- Then validate (deliver only once it passes, exit code 0):
  ```
  python3 {ENV_SEARCH_ROOT}/scripts/search_env/file_change_summary/validate_files_to_change.py --file {RUN_DIR}/file_change_summary/files_to_change.v0.3.json --run-dir {RUN_DIR}
  ```

### Definition of done
1. The prepass was run; its `resource_roles`/provenance/caveats/`context_inputs_read_only`/`provenance_and_caveats` are carried through faithfully (every requirement still has an entry).
2. Every `"undetermined"` task_role has been resolved; no `task_role_provisional` flags remain.
3. Every change_target has a real path relative to `extracted_root`, cites skill rules, and has a concrete `how`.
4. There's a request-agnostic scenario grouping plus one focused recommended set.
5. The output JSON validates against the schema (validator exits 0).
\end{lstlisting}

\SFNeedspace{7\baselineskip}

\subsubsection{Instruction composition}

\label{app:prompt-DS06}


\begin{lstlisting}[style=sfprompt]
# Compose a User Request (Query) for a Downloaded Environment

Write **one** realistic message a real engineer would send about this specific downloaded
codebase — the kind of ask where doing it *well* genuinely needs a particular skill, yet the
message reads like a normal practitioner's request, never like a skill manual or a solution writeup.

## Who is writing
A real engineer who **knew what they wanted, found a skill that fits, and read it once.** They
remember, roughly, its steps/techniques and reach for those **names** in the field's own vocabulary.
They do **not** recall the skill's (cryptic) name, and they never paste definitions, procedures, or
rule numbers. They're describing a goal — not reciting a handbook, and not handing over the answer.

---

## The two rules that make this work

### 1 — Speak in the skill's code words; hide that a manual exists
The remembered phrase-names are the skill's **code words**. An agent that also read the skill
recognizes them and does the task its way; one that didn't reads them as ordinary, slightly-ambiguous
language and drifts. That gap is the whole point.

- **DO** use the skill's own short step/technique names in their natural-phrase form ("keep the
  controllers CRUD-only", "give me a verdict on whether it's ready to merge"). Use only the ones this
  task needs.
- **DON'T** explain or define them. Use the term and move on.
- **DON'T** name the skill or hint one exists ("use the … skill", "per the guide/checklist").
- **DON'T** use identifier/locator forms: rule-ID slugs, config tokens (`safe_auto`), scenario keys,
  `task_role` labels, "Section 4 / Step 3 / Rule R6.12". Say them in plain words.
- **Test:** would a practitioner who never read *this* skill find the phrase ordinary (maybe a touch
  ambiguous) rather than gibberish or a cipher? If yes, it's a good code word.

### 2 — State the GOAL, never the SOLUTION
The user describes the **problem and what they want** — not how to fix it. They are *asking for* the
change; they do not already hold its answer.

- **DO** give the symptom / goal in the user's words: what's wrong or wanted, in which area, grounded
  in real files (e.g. "bad URLs leak a raw urllib3 error out of `requests.get` instead of a
  requests-native one — I want it to behave like the other invalid-URL cases").
- **DON'T** prescribe the implementation: no exact edit location, no "match the neighboring branch /
  existing style", no exact test name / parametrization / input construct, no step-by-step of the
  change. The `how` of the fix is the agent's job to discover.
- **NEVER reference anything under "Synthesis-only inputs"** in the requirement list (the upstream
  PR/patch). Don't name it, cite its number, link it, or hint it exists — it is the answer, and it is
  withheld from the agent's environment.

---

## What to produce
- **Grounded in this environment.** Refer to real files, paths, components, an issue number — taken
  from the requirement list — so the task can only be done in *this* codebase.
- **Cover every test-time requirement — no loss.** Touch **every** resource under **"Available
  resources (usability)"** whose health is `usable`/`degraded`: name the edit targets, reference the
  context resources (issue, docs, config, tests) concretely. A few related asks in one message are
  fine; weave them into one real person's message — not a bulleted spec. **Never** ask to edit an
  `unusable`/orphan resource, and **never** mention a synthesis-only one.
- **Genuinely skill-requiring**, signaled through the code words — expressed as the user's goal.
- **A real user's voice.** First person, concrete. As long as it needs to be, never a checklist.
- **One-shot ask — no reviewer in the loop.** The user states the task and expects it carried out;
  there is no human to check in with mid-task. Don't make the request hinge on the agent sharing a
  plan, pausing for approval, or confirming before it proceeds (no "show me your plan first", "let me
  review before you implement", "run it by me before you change X") — ask directly for the outcome.

## Inputs — read from disk (read ONLY these)
- **THE SKILL** — mine it for code-word vocabulary; learn its name/tokens so you can keep them out:
  {{SKILL_READING_INSTRUCTIONS}}
- **`./requirement_list.md`** — the substance: the task, and which resources are test-time vs
  **synthesis-only** (withheld). Your source of truth for what to ask for.
- **`./fewshot_examples.md`** — real forum requests; model their **voice and shape** only, not topics.
  If empty, just write in a natural human voice.

Do **not** open `../resources/`, and do not read other bookkeeping files in this directory.

## Output
Write JSON to `./query.json` (this turn, nothing else):

```json
{
  "general_request": "<one short line: the gist of what the user wants>",
  "user_request": "<THE FINAL QUERY — human-voiced, grounded, covers every test-time requirement, speaks in code words, states the goal not the solution. This is the artifact.>",
  "grounded_in": ["<every repo path / component / behavior the request targets>"],
  "coverage": {
    "test_time_requirements": ["<every requirement id under Available resources (usable/degraded)>"],
    "covered": ["<the ids actually reflected in user_request — aim to equal test_time_requirements>"],
    "omitted": [{"id": "<id>", "reason": "<why — should be rare>"}]
  },
  "skill_applicability": {"can_apply": true, "reason": "<one sentence: why doing this well needs the skill>"},
  "skill_signal_check": {"names_skill": false, "code_words_used": ["<natural-phrase skill terms used>"], "tokens_avoided": ["<rule ids / locators kept out>"]},
  "answer_leak_check": {"states_goal_not_solution": true, "prescribes_no_fix_steps": true, "names_synthesis_only_resource": false}
}
```

Before writing, re-read `user_request` once and confirm: (a) every test-time requirement is covered;
(b) the skill's name and all tokens/locators are absent; (c) code words appear **without explanation**;
(d) it states the **goal**, prescribes **no** fix steps, and references **no** synthesis-only resource;
(e) it reads like a real practitioner, not a manual; (f) it is a single-shot ask with **no** mid-task
review/approval gate (no "show me your plan / let me review before you proceed").
\end{lstlisting}

\SFNeedspace{12\baselineskip}

\subsection{Packaging and execution validation}

\SFNeedspace{7\baselineskip}

\subsubsection{Environment classification}

\label{app:prompt-setup-classify}


\begin{lstlisting}[style=sfprompt]
You are classifying the Docker BUILD difficulty of one benchmark task into A / B / C / D.
Output ONLY the JSON object required by the output schema — no prose.

## Base image already provides (do NOT count these as extra installs)
{{BASE_IMAGE_INVENTORY}}

## Classes
- **A** — base image suffices. Pure read / write / analysis, or runs code only on base
  runtimes. Source files present but only *read* ⇒ still A. setup.sh stays a no-op.
- **B** — base + build-time package installs ONLY: `pip install ...` and/or `npm install`
  in an extracted repo. No new language/system toolchain.
- **C** — needs a language toolchain NOT in base, installed at build time: Go, Rust,
  JDK+Maven/Gradle, .NET, Dart, Node 22 / pnpm, azure-cli, or a non-3.12 Python. Usually
  also needs offline dependency pre-warming (go mod download / cargo fetch / dotnet restore
  / mvn verify / pnpm install).
- **D** — infeasible on Linux: imports Apple/Xcode frameworks (UIKit, SwiftUI, MapKit,
  Kingfisher, etc.), OR requires a persistent service at *verify* time (a live Postgres /
  MySQL / Redis server — `docker-compose.yml` with a DB service the task actually uses is
  the classic signal). A build-time `RUN` layer cannot keep a service running.

## How to decide
Classify by what the task in `instruction.md` must actually **run or build**, NOT by which
files merely exist. Read the manifests in the env summary (`go.mod` `toolchain`, `package.json`
`engines`/`packageManager`, `Cargo.toml`, `pom.xml`, `*.csproj`, `pubspec.yaml`) — they reveal
the real toolchain. When class==D, fill `d_reason`.

`needs_runtime_internet`: set true if the agent will run a build/make target or script that
FETCHES at runtime and cannot be fully pre-warmed at build — inspect `Makefile`/`mvnw`/`package.json`
scripts for commands like `go get ...@latest`, `npm install`/`npx` of un-pinned deps, `mvn ... -U`,
or `*-install @latest`. If everything can be baked into the image at build (go mod download / cargo
fetch / dotnet restore / pinned pnpm install), keep it false.

## Task instruction (what the agent is asked to do)
{{INSTRUCTION}}

## Environment summary (file tree + manifests)
{{ENV_SUMMARY}}

## Skill (context only — NOT present in the build environment)
{{SKILL_EXCERPT}}

Now emit the classification JSON.
\end{lstlisting}

\SFNeedspace{7\baselineskip}

\subsubsection{Setup generation}

\label{app:prompt-setup-generate}


\begin{lstlisting}[style=sfprompt]
Write `environment/setup.sh` for this Harbor task so its Docker image builds with everything
baked in for an OFFLINE runtime. Your cwd is the task directory; edit files directly.

`setup.sh` runs at `docker build` time: root, cwd `/app`, network UP. `/app` already contains
the task's env files (`COPY files /app/`). At agent runtime the container may be OFFLINE.

## Rules
- Start with `#!/bin/bash`, then `set -euo pipefail`, then `cd /app`.
- Install ONLY what is missing. The base image already has (do NOT reinstall):
  {{BASE_IMAGE_INVENTORY}}
- Install EVERYTHING the agent needs at runtime HERE (runtime may be offline): toolchains AND
  their dependency caches. Pre-warm: `go mod download` / `cargo fetch` / `dotnet restore` /
  `mvn -B clean verify -DskipTests` / `pnpm install --frozen-lockfile` / `dart pub get`.
- Be architecture-adaptive: detect `$(dpkg --print-architecture)` or `$(uname -m)` (arm64 vs amd64).
- Go: also `export GOTOOLCHAIN=local` and `go env -w GOTOOLCHAIN=local` so offline runs never
  auto-download a toolchain for the `go.mod` `toolchain` directive.
- `cd /app/extracted/<repo>-*/` globs must match exactly one directory.
- No long-running services (build layers don't keep processes). Skill-agnostic: NO skill files
  exist at build time. Make optional/best-effort deps tolerant: `... || true`.
- If (and only if) the classification says `needs_runtime_internet: true`, set
  `allow_internet = true` in `task.toml` `[environment]`.
- For heavy toolchains, raise `task.toml` `[environment]` `build_timeout_sec` and `storage_mb`
  if needed (Rust target ~2-3 GB, Maven ~/.m2 ~2 GB, pnpm node_modules ~1.5 GB).
- The script must `exit 0` on success.

## Classification (the target class + detected toolchain/packages)
{{CLASSIFICATION}}

## Task instruction
{{INSTRUCTION}}

## Environment summary (file tree + manifests)
{{ENV_SUMMARY}}

## Gold examples (battle-tested setup.sh from shipped tasks — adapt, don't copy blindly)
{{EXAMPLES}}

Now write `environment/setup.sh` (overwrite the no-op), and edit `task.toml` if the rules above
require it. Keep the script tight and well-commented.
\end{lstlisting}

\SFNeedspace{12\baselineskip}

\subsection{Contrastive rubric construction}

\SFNeedspace{7\baselineskip}

\subsubsection{Rubric generation}

\label{app:prompt-rubric-generation-base}


\begin{lstlisting}[style=sfprompt]
# Generate a contrastive skill rubric (schema v2)

Produce **one** schema-valid contrastive skill rubric: an LLM-as-judge rubric set distilled from a
frontier **oracle**'s no-skill vs with-skill trajectory on the *same* task. The behavioral **DELTA**
between the two arms is *what the skill adds* and becomes the criteria. A judge later applies it to a
**weaker** model's two arms; the with-minus-no score delta estimates skill uptake.

**Ground truth is the behavioral DELTA, never the oracle trajectory.** Oracle-specific files/lines/
commands are *illustrative only*. A weaker no-skill run that scores **0 on every `skill_signal`
criterion is correct by design** — internalize this; it is the #1 failure mode.

This run's facts (substituted by the launcher):
- set_id `{{SET_ID}}` · task_id `{{TASK_ID}}` · output file `{{OUTPUT_FILENAME}}`
- oracle (the contrasted trajectories): `{{ORACLE_MODEL}}`
- generator (you, for `generation_provenance.generator_model`): `{{GENERATOR_MODEL}}`
- single-arm mode: `{{SINGLE_ARM}}` (when `true`, `./with.compact.json` is MISSING)
{{DIMENSION_FOCUS}}
## Inputs in your working directory (read from disk)
- `./no.compact.json`, `./with.compact.json` — oracle no-skill / with-skill trajectories, compacted.
  Events are `{i, t, [s], x, [o]}`: `i`=stable index (cite it; align arms by it), `t`∈
  `user|msg|read|search|exec|edit|list|other`, `s`=`fail|canc`, `x`=text/command, `o`=output.
  `./with.compact.json` is absent in single-arm mode — check.
- `./workspace/no/`, `./workspace/with/` — each arm's FINAL deliverable files (+ `MANIFEST.txt`, the
  full workspace tree). **As authoritative as the trajectory**: edit-bodies in the trajectory are
  capped at 900 chars, so verify what was actually produced — completeness, structure, correctness —
  from these files, not from trajectory narration.
- `./skill/SKILL.md` (+ `references/`) — mine for `named_phases`, `mandated_outputs`, vocabulary,
  hard-rules vs principles.
- `./inputs_meta.json` — `set_id`, `task`, `skill` shell, `oracle`, `source_run_pair`,
  `generated_at`, `single_arm` to copy verbatim (fill `task.one_line_goal`/`domain`/
  `answer_leak_risk` and `skill.name`/`named_phases`/`mandated_outputs` from SKILL.md + trajectories).
- `./contrastive_rubric.schema.json` — the **v2** schema you must validate against (authoritative
  field spec; read its one-line descriptions).
- `./fp-check.rubric.json` — a **v2 worked example**. Study its shape & rigor; do NOT copy its
  security-domain criteria unless this task genuinely is fp-check.
- `./validate_rubric.py` — run it in a loop until VALID.

## Method (in order)
1. **Mine the skill.** Extract `named_phases` (ordered steps), `mandated_outputs`, vocabulary, and
   which rules are hard vs principles.
2. **Read both arms — trajectory AND workspace.** Align by event index `i`; describe each arm's
   *behavior* (not files/lines): phases run, what was verified empirically, output format,
   over/under-reach. Then OPEN each arm's `./workspace/` deliverables to judge the actual artifact
   (existence, completeness, structure) — the file is the truth where the trajectory was capped.
3. **Enumerate candidate behaviors and classify each HONESTLY** (the schema enforces this):
   - `skill_signal` — with-skill does it (or better), no-skill doesn't/weaker ⇒
     `discrimination.with_skill_rank > no_skill_rank`, `arm_separation:"separated"`. **Only these feed
     the skill delta.**
   - `shared_baseline` — **both** arms do it ⇒ ranks **equal**. Scored for calibration, excluded from
     the delta. **Do not inflate a shared baseline into a skill_signal** — the cardinal sin.
   - `anti_pattern` — a no-skill failure mode the skill suppresses ⇒ `no_skill_rank ≥ with_skill_rank`,
     `anti_pattern_polarity:"penalize_when_met"`, and encode it higher=better (`unmet_value:1`).
   - Discard no-signal behaviors. At the frontier tier expect MANY `shared_baseline`; the measurable
     skill delta is usually a small set (durable artifact, empirical fail→fix, output discipline,
     phase adherence). The worked example honestly has ~4 skill_signal of ~9.
4. **Write each criterion** (per the schema `Criterion`). Hard constraints:
   - `statement`: starts `The agent`, ONE behavior, ≤30 words / ≤400 chars, **no oracle specifics**
     (no backticks, file paths, source extensions, `line N`/`:N`, `Response-N` — a regex rejects them).
   - Concreteness lives ONLY in `contrast_provenance` (`with_skill_behavior`/`no_skill_behavior`
     behavioral prose + `anchors[]` with `binding:"illustrative_only"`); set `hypothesized` (true in
     single-arm).
   - `judge_guidance`: `what_to_look_for` (generalizable), `positive_evidence[]`, `negative_anchor`
     (the no-skill/anti-pattern picture in **candidate** terms), `common_false_positives[]`.
   - `scoring`: always a `binary` block; add a `graded` ladder (`mode:"both"`, contiguous `0..max`,
     descriptors in candidate terms, one `is_negative_anchor`) for high-value `skill_signal`.
   - `discrimination`: integer ranks consistent with `criterion_type`; `arm_separation`.
   - `category_id` ∈ a `category_taxonomy` entry you declare; `weight` (0..10);
     `skill_attribution_confidence` (0..1); optional `category_tag`.
5. **Pick exactly ONE `gatekeeper`** — the core correctness check (often the verdict/answer); usually
   a `shared_baseline` (a frontier no-skill arm also gets it right), which is fine.
6. **Fill the set-level metadata — the v2 keys ONLY:** `schema_version` (`contrastive_skill_rubric.v2`),
   `set_id`/`task`/`skill`/`oracle`/`source_run_pair` from `inputs_meta.json`, `contrast_summary`
   (1–2 behavioral sentences ≤600 chars), `category_taxonomy` (only categories you used, with
   `category_weight`), `generation_provenance` (`generator_model:"{{GENERATOR_MODEL}}"`,
   `generator_prompt_id:"generate_rubric.md"`, `generated_at` from inputs_meta, `method`,
   `mandatory_abstraction_applied:true`), `set_confidence`, `criteria`.
   > **Do NOT emit** `judge_protocol`, `judge_output_schema`, `aggregation`, or
   > `trajectory_truth_guard` — in v2 those are harness-owned (`judge_policy.json`). The schema
   > rejects them as `additionalProperties`.

## Honesty rails
- **Single-arm** (`{{SINGLE_ARM}}==true`): `set_confidence.single_arm=true`,
  `source_run_pair.with_skill_run=null`, `method:"single_arm_inferred"`, every criterion
  `hypothesized=true` + `discrimination.arm_separation:"not_measured"`, `confidence≤0.5`. Infer from
  no-skill + skill spec; do not fabricate measured separation.
- **Empty delta** (arms behaviorally identical): `empty_delta=true`, only `shared_baseline`, `≤0.5`.
- **Harness reward is a placeholder** — `task.harness_reward_is_trustworthy=false`; judge by inspecting
  trajectories.
- **Answer leakage** — if the query telegraphs the answer or a withheld patch could leak, set
  `task.answer_leak_risk` and lower confidence (heavy ⇒ `≤0.6`); record in `set_confidence.caveats`.

## Output contract + self-validate loop (mandatory)
Prefer writing a small Python builder (construct the dict → `json.dump`) over hand-typing JSON. Write
`{{OUTPUT_FILENAME}}`, then run `python3 validate_rubric.py {{OUTPUT_FILENAME}}`, **fix every error and
re-run until `VALID … (0 errors)`.** (If `jsonschema` is missing, `pip install jsonschema>=4.0`.)
Write no other file except `{{OUTPUT_FILENAME}}` + optional builder/scratch; do not modify the schema,
worked example, compacted inputs, or validator. Final message (read programmatically): the output path,
criteria counts by `criterion_type`, and the gatekeeper id.
\end{lstlisting}

\par\clearpage
\endgroup

%% file: iclr2027_conference.bib
@article{yang2026make,
  title={What Makes Interaction Trajectories Effective for Training Terminal Agents?},
  author={Sidi Yang and Chaofan Tao and Jierun Chen and Tiezheng Yu and Ruoyu Wang and Yuxin Jiang and Yiming Du and Wendong Xu and Jing Xiong and Taiqiang Wu and Lifeng Shang and Xiaohui Li and Ngai Wong and Haoli Bai},
  journal={ArXiv},
  year={2026},
  volume={abs/2606.03461}
}

@article{Wu2026LargeScaleTA,
  title={Large-Scale Terminal Agentic Trajectory Generation from {Dockerized} Environments},
  author={Siwei Wu and Yizhi Li and Yuyang Song and Wei Zhang and Yang Wang and Riza Batista-Navarro and Xian Yang and Mingjie Tang and Bryan Dai and Jian Yang and Chenghua Lin},
  journal={ArXiv},
  year={2026},
  volume={abs/2602.01244}
}

@article{Hua2026CLIUniverseTV,
  title={{CLI-Universe}: Towards Verifiable Task Synthesis Engine for Terminal Agents},
  author={Zhanbo Hua and Yifan Yao and Weihao Xie and Yongchi Zhao and Minghao Liu and Ruizhi Qiu and Zhewei Huang and Zun Wang and Yiyan Ji and Yunhai Ye and Letian Zhu and Xinping Lei and Han Li and Zhiyuan Ma and Zili Wang and Zhaoxiang Zhang and Jiaheng Liu},
  journal={ArXiv},
  year={2026},
  volume={abs/2606.22883}
}

@article{Han2026RideWayBE,
  title={{RideWay}: Benchmarking Efficient Task Completion for Tool-Using Language Agents},
  author={Qingnuan Han and Boli Fang and Mingzhi Hou and Claire Liu},
  journal={ArXiv},
  year={2026},
  volume={abs/2609.17985}
}

@inproceedings{jimenez2024swebench,
title={{SWE}-bench: Can Language Models Resolve Real-world {GitHub} Issues?},
author={Carlos E Jimenez and John Yang and Alexander Wettig and Shunyu Yao and Kexin Pei and Ofir Press and Karthik R Narasimhan},
booktitle={International Conference on Learning Representations},
year={2024},
url={https://openreview.net/forum?id=VTF8yNQM66}
}

@inproceedings{zhou2024webarena,
      title={{WebArena}: A Realistic Web Environment for Building Autonomous Agents}, 
      author={Shuyan Zhou and Frank F. Xu and Hao Zhu and Xuhui Zhou and Robert Lo and Abishek Sridhar and Xianyi Cheng and Tianyue Ou and Yonatan Bisk and Daniel Fried and Uri Alon and Graham Neubig},
      year={2024},
      eprint={2307.13854},
      booktitle={International Conference on Learning Representations},
}

@inproceedings{zhang2026clawbench,
      title={{ClawBench}: Can {AI} Agents Complete Everyday Online Tasks?}, 
      author={Yuxuan Zhang and Yubo Wang and Yipeng Zhu and Penghui Du and Junwen Miao and Xuan Lu and Zhuofeng Li and Xingwei Qu and Zhengkang Guo and Yuanzhe Shen and Dingjie Song and Han Zhou and Tuney Zheng and Xian Wu and Hao Yu and Songcheng Cai and Yi Lu and Yunzhuo Hao and Minyi Lei and Liang Chen and Kai Zou and Huifeng Yin and Wendong Xu and Dongfu Jiang and Ping Nie and Jiaheng Liu and Wenhu Chen and Kelsey R. Allen},
      year={2026},
      eprint={2604.08523},
      booktitle={arXiv},
}

@inproceedings{ye2026claweval,
      title={{Claw-Eval}: Towards Trustworthy Evaluation of Autonomous Agents}, 
      author={Bowen Ye and Rang Li and Qibin Yang and Yuanxin Liu and Linli Yao and Hanglong Lv and Zhihui Xie and Chenxin An and Lei Li and Lingpeng Kong and Qi Liu and Zhifang Sui and Tong Yang},
      year={2026},
      eprint={2604.06132},
      booktitle={arXiv},
}

@inproceedings{lv2026realclawbench,
      title={{RealClawBench}: Live {OpenClaw} Benchmarks from Real Developer-Agent Sessions}, 
      author={Zongwei Lv and Yaoming Li and Zhewen Tan and Yilun Yao and Yuxuan Tian and Lin Sun and Xiangzheng Zhang and Weihong Lin and Tong Yang and Guangxiang Zhao},
      year={2026},
      eprint={2606.03889},
      booktitle={arXiv},
}

@inproceedings{ding2026wildclawbench,
      title={{WildClawBench}: A Benchmark for Real-World, Long-Horizon Agent Evaluation}, 
      author={Shuangrui Ding and Xuanlang Dai and Long Xing and Shengyuan Ding and Ziyu Liu and Yang JingYi and Penghui Yang and Zhixiong Zhang and Xilin Wei and Xinyu Fang and Yubo Ma and Haodong Duan and Jing Shao and Jiaqi Wang and Dahua Lin and Kai Chen and Yuhang Zang},
      year={2026},
      eprint={2605.10912},
      booktitle={arXiv},
}

@inproceedings{long2026liveclawbench,
      title={{LiveClawBench}: Benchmarking {LLM} Agents on Complex, Real-World Assistant Tasks}, 
      author={Xiang Long and Li Du and Yilong Xu and RongJian Xu and Qiyanhui Lu and Ying Gao and Qinhua Xie and Fangcheng Liu and Ning Ding and Haoqing Wang and Ziheng Li and Changjiang Zhou and Jianyuan Guo and Yehui Tang},
      year={2026},
      eprint={2604.13072},
      booktitle={arXiv},
}

@inproceedings{li2026clawenvkit,
      title={{ClawEnvKit}: Automatic Environment Generation for Claw-Like Agents}, 
      author={Xirui Li and Ming Li and Ion Stoica and Cho-Jui Hsieh and Tianyi Zhou},
      year={2026},
      eprint={2604.18543},
      booktitle={arXiv},
}

@inproceedings{lai2026clawforge,
      title={{ClawForge}: Generating Executable Interactive Benchmarks for Command-Line Agents}, 
      author={Yuxiang Lai and Peng Xia and Haonian Ji and Kaiwen Xiong and Kaide Zeng and Jiaqi Liu and Fang Wu and Jike Zhong and Zeyu Zheng and Cihang Xie and Huaxiu Yao},
      year={2026},
      eprint={2605.14133},
      booktitle={arXiv},
}

@inproceedings{
yao2025taubench,
title={{$\tau$-bench}: A Benchmark for {{\underline{T}}ool}-{{\underline{A}}gent}-{{\underline{U}}ser} Interaction in Real-World Domains},
author={Shunyu Yao and Noah Shinn and Pedram Razavi and Karthik R Narasimhan},
booktitle={International Conference on Learning Representations},
year={2025},
url={https://openreview.net/forum?id=roNSXZpUDN}
}

@inproceedings{trivedi2024appworld,
    title = {{A}pp{W}orld: A Controllable World of Apps and People for Benchmarking Interactive Coding Agents},
    author = {Trivedi, Harsh  and
      Khot, Tushar  and
      Hartmann, Mareike  and
      Manku, Ruskin  and
      Dong, Vinty  and
      Li, Edward  and
      Gupta, Shashank  and
      Sabharwal, Ashish  and
      Balasubramanian, Niranjan},
    booktitle = {Annual Meeting of the Association for Computational Linguistics},
    year = {2024},
    pages = {16022--16076},
}

@inproceedings{xie2024osworld,
 author = {Xie, Tianbao and Zhang, Danyang and Chen, Jixuan and Li, Xiaochuan and Zhao, Siheng and Cao, Ruisheng and Hua, Toh Jing and Cheng, Zhoujun and Shin, Dongchan and Lei, Fangyu and Liu, Yitao and Xu, Yiheng and Zhou, Shuyan and Savarese, Silvio and Xiong, Caiming and Zhong, Victor and Yu, Tao},
 booktitle = {Advances in Neural Information Processing Systems},
 pages = {52040--52094},
 title = {{OSWorld}: Benchmarking Multimodal Agents for Open-Ended Tasks in Real Computer Environments},
 volume = {37},
 year = {2024}
}

@inproceedings{
merrill2026terminalbench,
title={{Terminal-Bench}: Benchmarking Agents on Hard, Realistic Tasks in Command Line Interfaces},
author={Mike A. Merrill and Alexander Glenn Shaw and Nicholas Carlini and Boxuan Li and Harsh Raj and Ivan Bercovich and Lin Shi and Jeong Yeon Shin and Thomas Walshe and E. Kelly Buchanan and Junhong Shen and Guanghao Ye and Haowei Lin and Jason Poulos and Maoyu Wang and Marianna Nezhurina and Di Lu and Orfeas Menis Mastromichalakis and Zhiwei Xu and Zizhao Chen and Yue Liu and Robert Zhang and Leon Liangyu Chen and Anurag Kashyap and Jan-Lucas Uslu and Jeffrey Li and Jianbo Wu and Minghao Yan and Song Bian and Vedang Sharma and Ke Sun and Steven Dillmann and Akshay Anand and Andrew Lanpouthakoun and Bardia Koopah and Changran Hu and Etash Kumar Guha and Gabriel H. S. Dreiman and Jiacheng Zhu and Karl Krauth and Li Zhong and Niklas Muennighoff and Robert Kwesi Amanfu and Shangyin Tan and Shreyas Pimpalgaonkar and Tushar Aggarwal and Xiangning Lin and Xin Lan and Xuandong Zhao and Yiqing Liang and Yuanli Wang and Zilong Wang and Changzhi Zhou and David Heineman and Hange Liu and Harsh Trivedi and John Yang and Junhong Lin and Manish Shetty and Michael Yang and Nabil Omi and Negin Raoof and Shanda Li and Terry Yue Zhuo and Wuwei Lin and Yiwei Dai and Yuxin Wang and Wenhao Chai and Shang Zhou and Dariush Wahdany and Ziyu She and Jiaming Hu and Zhikang Dong and Yuxuan Zhu and Sasha Cui and Ahson Saiyed and Arinbj{\"o}rn Kolbeinsson and Jesse Hu and Christopher Michael Rytting and Ryan Marten and Yixin Wang and Jenia Jitsev and Alex Dimakis and Andy Konwinski and Ludwig Schmidt},
booktitle={International Conference on Learning Representations},
year={2026},
url={https://openreview.net/forum?id=a7Qa4CcHak}
}

@misc{pinchbench2026,
  author = {{Kilo AI}},
  title = {{PinchBench}: Benchmarking System for {OpenClaw} Coding Agents},
  year = {2026},
  howpublished = {\url{https://pinchbench.com}},
}

@inproceedings{liu2024apigen,
 author = {Liu, Zuxin and Hoang, Thai and Zhang, Jianguo and Zhu, Ming and Lan, Tian and Kokane, Shirley and Tan, Juntao and Yao, Weiran and Liu, Zhiwei and Feng, Yihao and Murthy, Rithesh and Yang, Liangwei and Savarese, Silvio and Niebles, Juan Carlos and Wang, Huan and Heinecke, Shelby and Xiong, Caiming},
 booktitle = {Advances in Neural Information Processing Systems},
 pages = {54463--54482},
 title = {{APIGen}: Automated {PI}peline for Generating Verifiable and Diverse Function-Calling Datasets},
 url = {https://proceedings.neurips.cc/paper_files/paper/2024/file/61cce86d180b1184949e58939c4f983d-Paper-Datasets_and_Benchmarks_Track.pdf},
 volume = {37},
 year = {2024}
}

@inproceedings{wang2026synthagent,
    title = {{S}ynth{A}gent: Adapting Web Agents with Synthetic Supervision},
    author = {Wang, Zhaoyang  and
      Liang, Yiming  and
      Zhang, Xuchao  and
      Wu, Qianhui  and
      Han, Siwei  and
      Bastos, Anson  and
      Wang, Rujia  and
      Bansal, Chetan  and
      Peng, Baolin  and
      Gao, Jianfeng  and
      Rajmohan, Saravan  and
      Yao, Huaxiu},
    booktitle = {Annual Meeting of the {A}ssociation for {C}omputational {L}inguistics},
    year = {2026},
    pages = {15730--15752},
}

@inproceedings{
xu2025agenttrek,
title={{AgentTrek}: Agent Trajectory Synthesis via Guiding Replay with Web Tutorials},
author={Yiheng Xu and Dunjie Lu and Zhennan Shen and Junli Wang and Zekun Wang and Yuchen Mao and Caiming Xiong and Tao Yu},
booktitle={International Conference on Learning Representations},
year={2025},
url={https://openreview.net/forum?id=EEgYUccwsV}
}

@inproceedings{wang2024executable,
author = {Wang, Xingyao and Chen, Yangyi and Yuan, Lifan and Zhang, Yizhe and Li, Yunzhu and Peng, Hao and Ji, Heng},
title = {Executable Code Actions Elicit Better {LLM} Agents},
year = {2024},
booktitle = {Proceedings of the 41st International Conference on Machine Learning},
series = {Proceedings of Machine Learning Research},
volume = {235},
pages = {50208--50232},
publisher = {PMLR},
}

@inproceedings{zhu2026termigen,
      title={{TermiGen}: High-Fidelity Environment and Robust Trajectory Synthesis for Terminal Agents}, 
      author={Kaijie Zhu and Yuzhou Nie and Yijiang Li and Yiming Huang and Jialian Wu and Jiang Liu and Ximeng Sun and Zhenfei Yin and Lun Wang and Zicheng Liu and Emad Barsoum and William Yang Wang and Wenbo Guo},
      year={2026},
      eprint={2602.07274},
      booktitle={arXiv},
}

@inproceedings{dong2026agentworld,
      title={{Agent-World}: Scaling Real-World Environment Synthesis for Evolving General Agent Intelligence}, 
      author={Guanting Dong and Junting Lu and Junjie Huang and Wanjun Zhong and Longxiang Liu and Shijue Huang and Zhenyu Li and Yang Zhao and Xiaoshuai Song and Xiaoxi Li and Jiajie Jin and Yutao Zhu and Hanbin Wang and Fangyu Lei and Qinyu Luo and Mingyang Chen and Zehui Chen and Jiazhan Feng and Ji-Rong Wen and Zhicheng Dou},
      year={2026},
      eprint={2604.18292},
      booktitle={arXiv},
}

@inproceedings{cheng2026terminalworld,
      title={{Terminal-World}: Scaling Terminal-Agent Environments via Agent Skills}, 
      author={Zihao Cheng and Hongru Wang and Zeming Liu and Xinyi Wang and Xiangrong Zhu and Yuhang Guo and Wei Lin and Jeff Z. Pan and Yunhong Wang},
      year={2026},
      eprint={2605.20876},
      booktitle={arXiv},
}

@misc{li2026skillsbenchbenchmarkingagentskills,
      title={{SkillsBench}: Benchmarking How Well Agent Skills Work Across Diverse Tasks}, 
      author={Xiangyi Li and Yimin Liu and Wenbo Chen and Bingran You and Zonglin Di and Yifeng He and Shenghan Zheng and Kyoung Whan Choe and Jiankai Sun and Shuyi Wang and Chujun Tao and Binxu Li and Xuandong Zhao and Hejia Geng and Xiaojun Wu and Junwei Zhou and Xiaokun Chen and Hanwen Xing and Yubo Li and Qunhong Zeng and Di Wang and Yuanli Wang and Roey Ben Chaim and Penghao Jiang and Haotian Shen and Luyang Kong and Xinyi Liu and Runhui Wang and Xuanqing Liu and Jiachen Li and Xin Lan and Yueqian Lin and Wengao Ye and Junwei He and Songlin Li and Yue Zhang and Yipeng Gao and Yijiang Li and Ze Ma and Liqiang Jing and Tianyu Wang and Kaixin Li and Yiqi Xue and Haoran Lyu and Yizhuo He and Yuchen Tian and Shutong Wu and Bowei Wang and Yixuan Gao and Bo Chen and Litong Liu and Sikai Cheng and Jiajun Bao and Shuaicheng Tong and Shuwen Xu and Terry Yue Zhuo and Tinghan Ye and Qi Qi and Miao Li and Longtai Liao and Zelin Tan and Chang Shi and Xilin Tang and Srinath Tankasala and Boqin Yuan and Yaoyao Qian and Jianhong Tu and Chenguang Wang and Yizhou Sun and Wei Wang and Aaron Taylor and Ziyue Yang and Changkun Guan and Zhikang Dong and Xinyu Zhang and Steven Dillmann and Han-chung Lee and Dawn Song},
      year={2026},
      eprint={2602.12670},
      archivePrefix={arXiv},
      primaryClass={cs.AI},
      url={https://arxiv.org/abs/2602.12670}, 
}

@misc{shaposhnikov2026frameworkevaluatingagenticskills,
      title={A Framework for Evaluating Agentic Skills at Scale}, 
      author={Maksim Shaposhnikov and Nicolas Fortuin and Simon Stipcich and Maria I. Gorinova and Amy Heineike and Rob Willoughby},
      year={2026},
      eprint={2606.17819},
      archivePrefix={arXiv},
      primaryClass={cs.SE},
      url={https://arxiv.org/abs/2606.17819}, 
}

@article{kwiatkowski-etal-2019-natural,
    title = "{Natural Questions}: A Benchmark for Question Answering Research",
    author = "Kwiatkowski, Tom  and
      Palomaki, Jennimaria  and
      Redfield, Olivia  and
      Collins, Michael  and
      Parikh, Ankur  and
      Alberti, Chris  and
      Epstein, Danielle  and
      Polosukhin, Illia  and
      Devlin, Jacob  and
      Lee, Kenton  and
      Toutanova, Kristina  and
      Jones, Llion  and
      Kelcey, Matthew  and
      Chang, Ming-Wei  and
      Dai, Andrew M.  and
      Uszkoreit, Jakob  and
      Le, Quoc  and
      Petrov, Slav",
    editor = "Lee, Lillian  and
      Johnson, Mark  and
      Roark, Brian  and
      Nenkova, Ani",
    journal = "Transactions of the Association for Computational Linguistics",
    volume = "7",
    year = "2019",
    address = "Cambridge, MA",
    publisher = "MIT Press",
    url = "https://aclanthology.org/Q19-1026/",
    doi = "10.1162/tacl_a_00276",
    pages = "452--466"
}

@inproceedings{zhou-etal-2026-rubricbench,
    title = "{R}ubric{B}ench: Aligning Model-Generated Rubrics with Human Standards",
    author = "Zhou, Junyi  and
      Zhang, Qiyuan  and
      Wang, Yufei  and
      Lyu, Fuyuan  and
      Ming, Yidong  and
      Xu, Can  and
      Sun, Qingfeng  and
      Zheng, Kai  and
      Kang, Peng  and
      Liu, Xue  and
      Ma, Chen",
    editor = "Liakata, Maria  and
      Moreira, Viviane P.  and
      Zhang, Jiajun  and
      Jurgens, David",
    booktitle = "Proceedings of the 64th Annual Meeting of the {A}ssociation for {C}omputational {L}inguistics (Volume 1: Long Papers)",
    month = jul,
    year = "2026",
    address = "San Diego, California, United States",
    publisher = "Association for Computational Linguistics",
    url = "https://aclanthology.org/2026.acl-long.1439/",
    doi = "10.18653/v1/2026.acl-long.1439",
    pages = "31179--31200",
    ISBN = "979-8-89176-390-6"
}

@article{
li2026dynamic,
title={Dynamic Agent Skills: A Lifecycle Survey and Taxonomy of Evolving Skill Libraries},
author={Yubo Li},
journal={Transactions on Machine Learning Research},
issn={2835-8856},
year={2026},
url={https://openreview.net/forum?id=cjU3YbcRr8},
note={Survey Certification}
}
